\pdfoutput=1
\documentclass[10pt,twocolumn,letterpaper]{article}
\usepackage{preprint}

\newcommand{\authorone}{Zhiqi Li}
\newcommand{\authortwo}{Bo Zhu}
\newcommand{\affiliationA}{Georgia Institute of Technology}
\newcommand{\contactemail}{zli3167@gatech.edu}

\usepackage[T1]{fontenc}
\usepackage[utf8]{inputenc}
\usepackage{amsmath,amssymb,amsfonts,amsthm,mathtools,bm}
\usepackage{booktabs,array,tabularx,multirow}
\usepackage{algorithm,algpseudocode}
\usepackage{graphicx,xcolor,etoolbox}
\usepackage{capt-of}
\usepackage{placeins}
\definecolor{linkblue}{rgb}{0.21,0.49,0.74}
\usepackage[pagebackref,breaklinks,colorlinks,allcolors=linkblue]{hyperref}
\usepackage{url}
\usepackage{fvextra}
\DefineVerbatimEnvironment{pycode}{Verbatim}{fontsize=\fontsize{6.5pt}{7.4pt}\selectfont,breaklines=true,breakanywhere=true,baselinestretch=0.92,xleftmargin=2pt}
\usepackage{tikz}
\usetikzlibrary{decorations.pathreplacing}
\hypersetup{pdftitle={dKV Forcing: Separating Generation and Memory Representations through a Differentiable KV Cache},pdfauthor={Zhiqi Li, Bo Zhu}}

\newcommand{\KV}{\mathrm{KV}}
\newcommand{\sg}{\operatorname{sg}}

\newcommand{\Kwriter}{\mathcal{K}}
\newcommand{\Hist}{\mathcal{H}}
\newcommand{\transpose}{\mathsf{T}}
\newcommand{\ctx}{\mathrm{ctx}}
\newcommand{\gen}{\mathrm{gen}}
\newcommand{\mem}{\mathrm{mem}}

\newcommand{\DMD}{\mathrm{DMD}}
\newcommand{\diff}{\mathop{}\!\mathrm{d}}

\newcommand{\rev}[1]{{#1}}
\newtheorem{proposition}{Proposition}
\newtheorem*{restatedproposition}{Proposition~\ref{prop:exact}}
\newtheorem{lemma}{Lemma}

\newcolumntype{Y}{>{\raggedright\arraybackslash}X}
\title{dKV Forcing: Separating Generation and Memory Representations\\through a Differentiable KV Cache}
\author{\authorone\qquad \authortwo\\[2pt]
\affiliationA\\[2pt]
{\tt\small \contactemail}\\[3pt]
{\small Project page: \url{https://zhiqili-cg.github.io/dKV-Forcing-Project}}\\
{\small Code: \url{https://github.com/ZhiqiLi-CG/dKV-Forcing}}}
\begin{document}
\addtocontents{toc}{\protect\setcounter{tocdepth}{-10}}   

\definecolor{kvGray}{HTML}{778595}
\twocolumn[{%
\renewcommand\twocolumn[1][]{#1}%
\maketitle
\vspace{-16pt}
\begin{center}
\begingroup
\setlength{\tabcolsep}{0pt}\renewcommand{\arraystretch}{0}
\begin{tabular}{@{}c@{}}
\includegraphics[width=\textwidth]{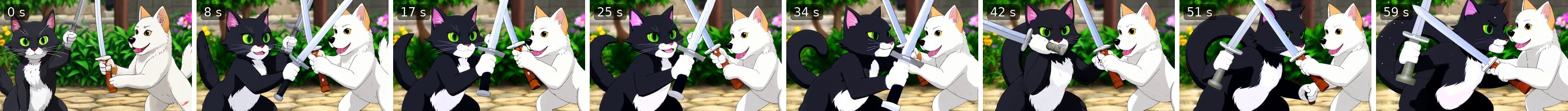}\\[1.5pt]
\includegraphics[width=\textwidth]{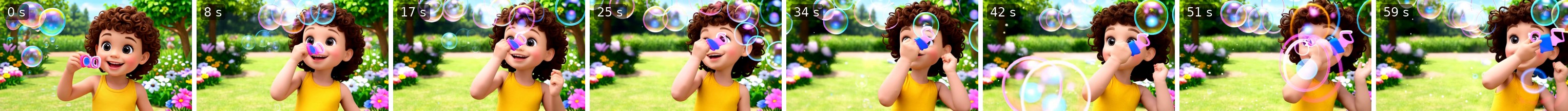}\\[1.5pt]
\includegraphics[width=\textwidth]{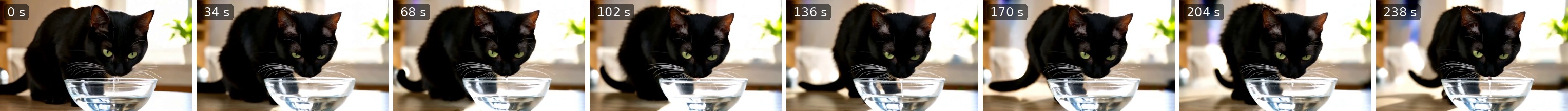}\\[1.5pt]
\includegraphics[width=\textwidth]{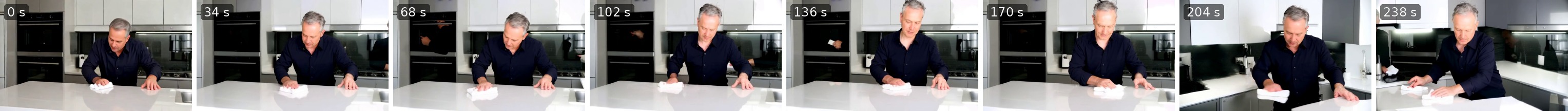}
\end{tabular}
\endgroup

\vspace{-2pt}
\captionof{figure}{\textbf{Trained on 5\,s clips, dKV Forcing generates up to four minutes.} The top two rows are 60\,s generations and the bottom two are 240\,s generations, $11.5$ and $46$ times the latent length of a training clip, with the time stamp drawn in each frame. The prompts are a cat and a dog in a sword fight, a child blowing bubbles, a cat drinking water, and a man wiping a kitchen counter. Subjects stay recognizable and in motion over the whole video; the same generations are compared with Causal Forcing++ (CF++) and Self Gradient Forcing (SGF) in Figure~\ref{fig:qual60} and Appendix~\ref{app:qualitative}.}
\label{fig:teaser}
\end{center}
\vspace{4pt}
}]

\input{figures/memory_learning}  
\begin{abstract}
A causal video generator writes every segment it produces into a key--value (KV) cache, and later segments know the past only through it. Self-rollout training stops the gradient at this cache, and Self Gradient Forcing (SGF) restores it, but applies it to the same weights that generate the current segment. We find that writing and generating are different roles: in a pretrained generator the written KV already differs from the KV used for denoising, and on shared weights the gradient from future reads is negatively aligned with the gradient of generation. We propose dKV Forcing (dKV denotes differentiable KV), which gives the write its own parameters, a low-rank residual used only when a segment enters the cache, and trains them with the gradient of the later segments that read what they wrote. Opening this gradient through the cache in common use exhausts GPU memory. We trace this to checkpoint replay re-writing the cache, not to the gradient path, which removes the need for the second reconstruction pass of SGF. dKV Forcing keeps the cache as persistent differentiable pages and computes the exact gradient in one temporal traversal, with an $11\%$ lower memory peak and $28\%$ less resident memory. With pages that store hidden states, it trains 15\,s clips on which the two-pass reconstruction runs out of memory. Trained on 5\,s clips, dKV Forcing reaches the highest VBench-Long quality score at 60\,s and 240\,s and outperforms SGF by $3.6$ and $6.1$ points of aesthetic quality, with a higher dynamic degree.
\end{abstract}


\section{Introduction}
Causal video diffusion models generate a video segment by segment (a latent frame, or a short chunk of frames), each new segment attending to a KV cache of the segments already produced, and reach streaming speed on a single GPU~\citep{chen2024diffusionforcing,yin2025causvid,huang2025self}. In such a model the cache is the memory: what a later segment knows about the past is exactly what earlier segments wrote into the cache.

Self Forcing made these models usable at length by training them on the histories they generate themselves, under a video-level distribution-matching objective~\citep{huang2025self,yin2024dmd,yin2024dmd2}, and Causal Forcing and Causal Forcing++ improve its teacher and its initialization~\citep{zhu2026causalforcing,zhao2026causalforcingpp}. Their gradient, however, stops at the cache. Each segment is trained to generate well given the history it is handed, while no gradient reaches the operations that produced that history. As a result, the model is not taught how to remember.

Self Gradient Forcing (SGF) restores this gradient by reconstructing the whole clip in a second differentiable forward, so that every write is trained by the segments that read it~\citep{zhuang2026sgf}. But the gradient lands on the projections that also denoise the current segment, and there the two roles conflict: in a trained model the gradient from future reads is as large as the gradient of generation and negatively aligned with it. The two roles also produce different KV: the KV a segment writes cannot be replaced by the KV it used for denoising without degrading the continuation (Sec.~\ref{sec:representation}). Writing memory and generating a frame are thus different roles, and they need different representations.

We propose \emph{dKV Forcing} (dKV denotes differentiable KV), which gives the write its own parameters and trains them only with the gradient of the segments that later read what they wrote (Figure~\ref{fig:memorylearning}). The writer can be a second network that shares nothing with the generator. We use a lightweight form: a low-rank residual~\citep{hu2022lora} on the K and V projections of the shared backbone, active only when a segment is written into the cache (Sec.~\ref{sec:representation}). Denoising uses the backbone unchanged, so the residual changes a video only through what later segments read.
\begin{figure*}[t]
\centering
\begingroup
\definecolor{kvInk}{HTML}{24374A}
\definecolor{kvGray}{HTML}{778595}
\definecolor{kvGold}{HTML}{5B7083}
\definecolor{kvTeal}{HTML}{E0602B}
\definecolor{kvPurple}{HTML}{8661A4}
\definecolor{kvRed}{HTML}{C0392B}
\definecolor{kvBlue}{HTML}{2456C6}
\resizebox{\textwidth}{!}{\begin{tikzpicture}[x=1cm,y=1cm,>=stealth,font=\sffamily\footnotesize,line cap=round,line join=round,
 dit/.style={draw=kvInk!45,fill=kvInk!5,rounded corners=1.5pt,minimum width=1.0cm,minimum height=0.62cm,inner sep=1pt,text=kvInk},
 ditoff/.style={dit,draw=kvInk!25,fill=white,text=kvGray},
 slot/.style={draw=kvInk!55,fill=kvInk!12,minimum width=1.7cm,minimum height=0.36cm,inner sep=0pt,font=\sffamily\scriptsize,text=kvInk},
 page/.style={draw=kvTeal!85,fill=kvTeal!22,minimum width=1.0cm,minimum height=0.36cm,inner sep=0pt,font=\sffamily\scriptsize,text=kvInk},
 handle/.style={draw=kvPurple,fill=white,circle,minimum size=0.4cm,inner sep=0pt,font=\sffamily\scriptsize,text=kvPurple},
 pass/.style={->,draw=kvGray!45,line width=0.8pt},
 stop/.style={->,draw=kvInk!45,line width=1.3pt},
 rew/.style={->,draw=kvRed,line width=1.3pt},
 hgrad/.style={->,draw=kvPurple,line width=0.9pt,dash pattern=on 2pt off 1.4pt},
 alive/.style={draw=kvInk,line width=1.0pt,dash pattern=on 1.2pt off 1.6pt,rounded corners=2.5pt},
 heading/.style={font=\sffamily\small\bfseries,text=kvInk,anchor=west},
 rowlab/.style={font=\sffamily\scriptsize,text=kvGray,anchor=east,align=right},
 note/.style={font=\sffamily\scriptsize,text=kvInk,anchor=north west,align=left},
 seg/.style={font=\sffamily\scriptsize,text=kvGray}]
\newcommand{\kvpanel}[2]{%
 \begin{scope}[shift={(#1,0)}]
 \foreach \j/\lab in {0/i,1/i{+}1,2/i{+}2}{
   \node[seg] at (1.7*\j,-0.9) {$\lab$};
   \node[dit] (D\j) at (1.7*\j,-0.35) {denoise};
   \node[draw=kvInk!45,fill=white,rounded corners=1pt,minimum width=0.6cm,minimum height=0.3cm,inner sep=0pt,font=\sffamily\scriptsize,text=kvInk] (X\j) at (1.7*\j,0.45) {$\tilde x$};
   \node[dit] (W\j) at (1.7*\j,1.25) {write};
 }
 \ifx#2c
   \draw[kvInk!45,dash pattern=on 2pt off 1.5pt,rounded corners=3pt] (-0.75,2.2) rectangle (4.15,2.86);
   \node[font=\sffamily\scriptsize,text=kvGray,anchor=south east,inner sep=1pt] at (4.15,2.92) {detached values};
   \foreach \j/\lab in {0/i,1/i{+}1,2/i{+}2}{
     \node[page] (K\j) at (1.7*\j-0.12,2.53) {page};
     \draw[pass,draw=kvTeal] (W\j.north-|K\j) -- (K\j.south);
     \node[handle] (H\j) at (1.7*\j+0.72,1.9) {$h$};
     \draw[hgrad] (H\j) -- (W\j.north east);
   }
   \foreach \a/\b in {0/1,1/2}{
     \draw[stop,line width=0.9pt] ([xshift=6pt]K\a.south) -- ([xshift=-3pt]W\b.north);
     \draw[stop,line width=0.9pt] ([xshift=1pt]K\a.south) -- ([xshift=-3pt]D\b.north);
     \draw[hgrad] (W\b.north west) -- (H\a.east);
     \draw[hgrad] (D\b.north west) -- (H\a.south east);
   }
 \else
   \foreach \j/\lab in {0/i,1/i{+}1,2/i{+}2}{
     \node[slot] (K\j) at (1.7*\j,2.53) {slot};
   }
   \node[font=\sffamily\scriptsize,text=kvGray,anchor=south east] at (4.25,2.72) {one tensor, written in place};
   \ifx#2a
     \foreach \j in {0,1,2}{ \draw[stop] (W\j.north) -- (K\j.south); }
     \foreach \a/\b in {0/1,1/2}{
       \draw[stop] ([xshift=10pt]K\a.south) -- (W\b.north west);
       \draw[stop] ([xshift=4pt]K\a.south) -- (D\b.north west);
     }
   \else
     \draw[pass] (W0.north) -- (K0.south);
     \foreach \a/\b in {0/1,1/2}{
       \draw[pass] ([xshift=10pt]K\a.south) -- (W\b.north west);
       \draw[pass] ([xshift=4pt]K\a.south) -- (D\b.north west);
     }
     \draw[rew] (W1.north) -- (K1.south);
     \draw[rew] (W2.north) -- (K2.south);
     \draw[rew] (D2.east) -- ++(0.62,0) |- (K2.east);
   \fi
 \fi
 \foreach \j in {0,1,2}{ \draw[pass] (D\j.north) -- (X\j.south); \draw[stop] (X\j.north) -- (W\j.south); }
 \end{scope}}
\node[rowlab] at (-0.85,2.53) {KV cache};
\node[rowlab] at (-0.85,1.25) {$\Kwriter_{\theta,\phi}$};
\node[rowlab] at (-0.85,-0.35) {$G_\theta$};
\node[rowlab] at (-0.85,-0.9) {segment};
\node[heading] at (-0.7,3.52) {(a) Self Forcing: detached cache};
\kvpanel{0}{a}
\draw[alive] (1.7-0.62,-0.75) rectangle (1.7+0.62,0.05);
\node[note] at (-0.7,-1.2) {Replay is released.\\No gradient reaches the writes.};
\node[heading] at (5.95,3.52) {(b) Gradient through the in-place cache};
\kvpanel{6.7}{b}
\draw[alive] (6.7+1.7-0.62,1.25-0.4) rectangle (6.7+1.7+0.62,1.25+0.4);
\draw[alive] (6.7+3.4-0.62,1.25-0.4) rectangle (6.7+3.4+0.62,1.25+0.4);
\draw[alive] (6.7+3.4-0.62,-0.75) rectangle (6.7+3.4+0.62,0.05);
\node[note] at (5.95,-1.2) {Replay writes the cache again and\\is not released: \textcolor{kvRed}{\textbf{out of memory}}.};
\node[heading] at (12.65,3.52) {(c) dKV Forcing: pages and handles};
\kvpanel{13.4}{c}
\draw[alive] (13.4+1.7-0.62,1.25-0.4) rectangle (13.4+1.7+0.62,1.25+0.4);
\node[note] at (12.65,-1.2) {Replay only reads and is released.\\The gradient reaches the writes through $h$.};
\draw[->,kvInk,line width=1.4pt] (17.6,3.95) -- (-0.7,3.95) node[midway,above,font=\sffamily\footnotesize\bfseries,text=kvInk] {backward pass, at the replay of segment $i{+}1$};
\node[font=\sffamily\scriptsize,text=kvInk,anchor=west] at (-0.7,-2.2) {%
 \tikz{\draw[pass] (0,0)--(0.5,0);}\ forward, gradient passes\quad
 \tikz{\draw[stop] (0,0)--(0.5,0);}\ forward, stop-gradient\quad
 \tikz{\draw[rew] (0,0)--(0.5,0);}\ write repeated by replay\quad
 \tikz{\draw[hgrad] (0,0)--(0.5,0);}\ gradient via handle\quad
 \tikz{\draw[alive] (0,0) rectangle (0.5,0.22);}\ replayed activations currently alive};
\end{tikzpicture}}
\endgroup
\caption{\textbf{Only pages and handles carry the memory gradient through replay.} \textbf{(a)} Self Forcing detaches the cache: each replay is released, and the writes receive no gradient. \textbf{(b)} With the gradient passing through the in-place cache, every replay writes the cache again and its activations stay attached until backward runs out of memory. \textbf{(c)} dKV Forcing stores the values in a page table, written once outside replay, and a zero-valued handle $h$ carries the gradient to the write, so replay only reads and is released.}
\label{fig:pagecontract}
\end{figure*}
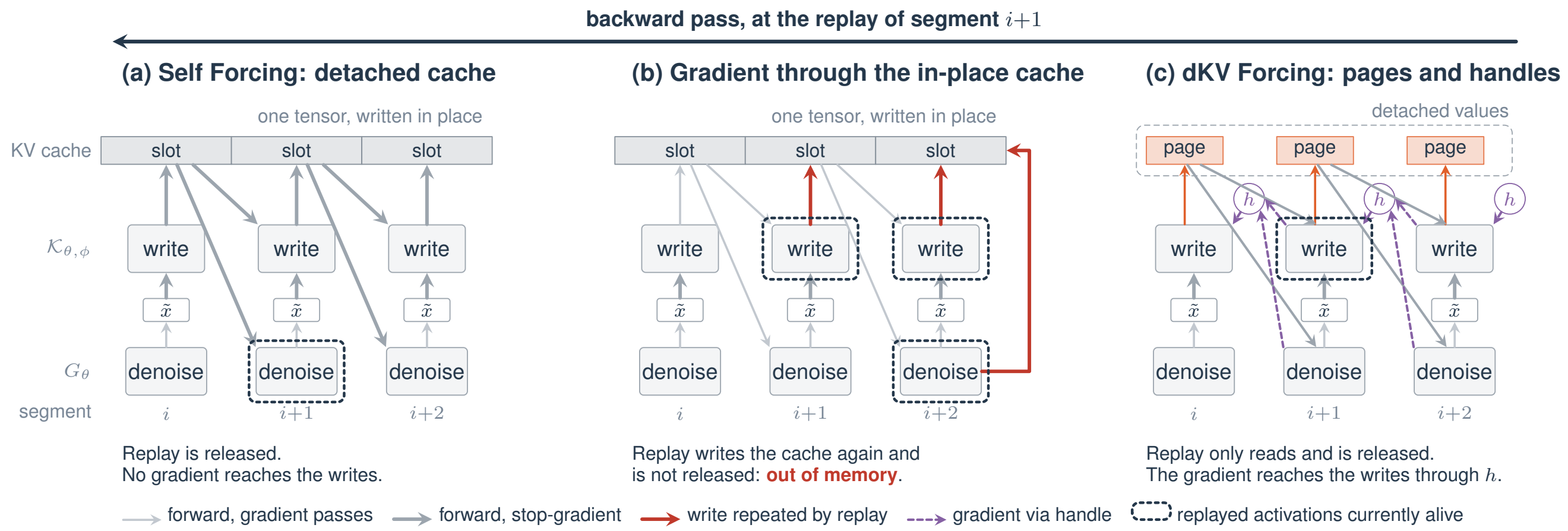   

Training such a writer needs the gradient from future reads back to past writes, and both known ways to compute it are costly. Opening the gradient through the ordinary KV cache runs out of memory. SGF attributes this to the autograd graph that every cache write must retain, which grows with the rollout~\citep{zhuang2026sgf}, and avoids it with a second forward that pays for a second pass and holds the graph of the whole clip at once. We find a different cause. In video training a rollout fits in GPU memory only when its forwards are checkpointed and replayed in backward, and every replay then leaves another copy of its activations attached to the non-detached cache, until training runs out of memory even on 5\,s clips (Figure~\ref{fig:pagecontract}). The growth comes from the replay, not from the gradient path. dKV Forcing computes the same gradient in a single temporal traversal by separating the two roles of the cache (Sec.~\ref{sec:algorithm}): a page table stores the values, detached and written once, while a zero-valued handle per page carries the gradient to its writer, so that replay only reads the cache and the traversal stays exact (Proposition~\ref{prop:exact}).

The single traversal is cheaper than the two-pass reconstruction of SGF and trains longer clips. Its pages need $28\%$ less resident memory and make every rollout faster, which shortens the training cycle by $20.8\%$, and with pages that store the attention inputs in place of K and V (hidden-state pages) it trains 15\,s clips on which the two-pass reconstruction runs out of memory (Sec.~\ref{sec:exp-algo}). Trained on 5\,s clips of the 1.3B Causal Forcing++ generator, dKV Forcing reaches the highest VBench-Long quality score at 60\,s and 240\,s (Figure~\ref{fig:teaser}). It outperforms SGF by $3.6$ and $6.1$ points of aesthetic quality with a higher dynamic degree, and it loses less than half as much aesthetic quality over a four-minute video (Sec.~\ref{sec:exp-main}).

In summary, we contribute (1)~a writer separated from the generator, as a dedicated network or as a low-rank residual, that learns the memory representation from its future reads; (2)~persistent differentiable pages with local rematerialization, which compute the exact memory gradient in one temporal traversal; and (3)~the highest long-horizon quality at 60\,s and 240\,s from 5\,s training clips, at a lower training cost than the two-pass reconstruction.

\section{Related work}
\label{sec:related}
\paragraph{Self-rollout training and context gradients.}
Self Forcing trains autoregressive video diffusion on generated histories with a video-level objective and detached writes~\citep{huang2025self}, and Causal Forcing~\citep{zhu2026causalforcing} and Causal Forcing++~\citep{zhao2026causalforcingpp} keep the detached write. Self Gradient Forcing~\citep{zhuang2026sgf} restores context gradients through a detached rollout followed by a second, whole-sequence differentiable forward. We compute the same gradient in a single temporal traversal and use its two-pass implementation as the correctness and cost reference. Concurrent Connected Self Forcing~\citep{zhang2026connected} reconnects gradients across chunks through the generated latents with shortcut gradient replay. Concurrent SGF+~\citep{su2026sgfplus} also separates the parameters of context writing and denoising, and both works observe the conflict between their gradients. SGF+ studies this conflict, while we contribute the single-traversal algorithm and the attribution of the open-cache memory failure. Our replay builds on activation checkpointing~\citep{griewank2000revolve,williams1990efficient,gruslys2016memory,chen2016checkpoint,jain2020checkmate,kirisame2021dtr,korthikanti2023reducing,pytorch2026checkpoint}, with the committed KV pages kept outside the checkpointed computations.

\begin{table*}[t]
\caption{\textbf{dKV Forcing has the highest quality score at 60\,s and at 240\,s, in both generators.} VBench-Long standard prompts, mean $\pm$ standard deviation (SD) over training seeds $\{1,10,20\}$, step 1500; 240\,s is $963$ latent frames, $46$ times a training clip. Quality is the official VBench quality score over the six dimensions scored on every video (aesthetic, dynamic, imaging, subject, background, smoothness). Per block, best value per column in bold, second best underlined.}
\label{tab:main}
\centering\small
\setlength{\tabcolsep}{3.5pt}
\resizebox{\linewidth}{!}{\begin{tabular}{@{}lcccccccccc@{}}
\toprule
& \multicolumn{5}{c}{VBench-Long 60\,s} & \multicolumn{5}{c}{VBench-Long 240\,s}\\
\cmidrule(lr){2-6}\cmidrule(lr){7-11}
configuration & Quality & aesthetic & dynamic & imaging & subject & Quality & aesthetic & dynamic & imaging & subject\\
\midrule
\multicolumn{11}{@{}l}{\emph{frame-wise (one latent frame per segment)}}\\
CF++ & $\underline{78.7}\pm0.8$ & $52.7\pm1.4$ & $\mathbf{76.5}\pm19.0$ & $63.3\pm1.6$ & $94.7\pm1.3$ & $\underline{78.0}\pm1.0$ & $49.0\pm2.5$ & $\mathbf{81.7}\pm14.9$ & $61.7\pm2.7$ & $93.8\pm1.0$\\
SGF & $78.0\pm0.7$ & $\underline{56.2}\pm1.4$ & $38.5\pm6.9$ & $\underline{67.4}\pm2.2$ & $\underline{96.5}\pm0.4$ & $76.4\pm1.1$ & $\underline{51.3}\pm1.7$ & $38.9\pm7.4$ & $\underline{64.6}\pm3.5$ & $\underline{95.9}\pm0.3$\\
dKV Forcing (ours) & $\mathbf{81.2}\pm1.8$ & $\mathbf{59.8}\pm1.3$ & $\underline{71.4}\pm34.5$ & $\mathbf{68.8}\pm1.3$ & $\mathbf{96.6}\pm1.2$ & $\mathbf{80.6}\pm2.2$ & $\mathbf{57.4}\pm0.4$ & $\underline{72.0}\pm33.8$ & $\mathbf{68.2}\pm0.5$ & $\mathbf{96.2}\pm1.0$\\
\midrule
\multicolumn{11}{@{}l}{\emph{chunk-wise (three latent frames per segment)}}\\
CF++ & $78.9\pm1.4$ & $57.8\pm1.6$ & $\underline{50.6}\pm15.8$ & $65.6\pm1.2$ & $\underline{96.8}\pm0.7$ & $76.9\pm2.4$ & $52.5\pm4.3$ & $\underline{50.4}\pm16.1$ & $62.3\pm2.9$ & $95.9\pm0.5$\\
SGF & $\underline{79.4}\pm1.5$ & $\underline{58.1}\pm0.1$ & $48.4\pm19.8$ & $\underline{67.8}\pm1.4$ & $\mathbf{96.9}\pm0.5$ & $\underline{77.8}\pm2.0$ & $\underline{53.2}\pm1.0$ & $46.8\pm20.4$ & $\underline{65.4}\pm2.7$ & $\underline{96.4}\pm0.5$\\
dKV Forcing (ours) & $\mathbf{81.1}\pm1.8$ & $\mathbf{59.6}\pm0.8$ & $\mathbf{61.9}\pm19.7$ & $\mathbf{68.7}\pm1.7$ & $\underline{96.8}\pm0.8$ & $\mathbf{79.5}\pm2.7$ & $\mathbf{55.7}\pm2.2$ & $\mathbf{57.1}\pm24.3$ & $\mathbf{66.2}\pm2.3$ & $\mathbf{96.6}\pm0.6$\\
\bottomrule
\end{tabular}}
\end{table*}
\paragraph{Long video generation.}
Methods built for long video generation extend the context a segment sees: Rolling Forcing denoises a rolling window~\citep{liu2025rollingforcing}, LongLive tunes on streams and re-caches the KV~\citep{yang2025longlive}, FramePack packs the frame context~\citep{zhang2025framepack}, and retrieval brings back stored memories~\citep{yu2025contextasmemory,xiao2025worldmem}. Cache compression and paging manage which KV entries stay available at inference~\citep{xiao2024streamingllm,zhang2023h2o,li2024snapkv,kwon2023pagedattention}. dKV Forcing is orthogonal to them: it trains what each write puts into that context, with a low-rank residual~\citep{hu2022lora,zhang2023adalora,liu2024dora} that only the write uses, and leaves the inference cache and its schedule unchanged, so it applies under any of these context designs.

\section{Background}
\label{sec:formulation}
\paragraph{Notation.}
A video is a sequence of $T=n+1$ segments $i=0,\dots,n$, each one latent frame (frame-wise) or three (chunk-wise), with clean latent $x_i$. Segment $i$'s denoising state is $z_i^{(t)}$ with $z_i^{(0)}=\epsilon_i$ and $z_i^{(1)}=x_i$ on the grid $0=t_0<\cdots<t_k=1$, and $\tilde x_i^{t_j}$ is the clean prediction made at step $t_j$, the last one being the output $x_i$. Text conditioning is omitted, $\epsilon$ denotes fresh Gaussian noise and $\KV_{<i}$ the cached KV of the earlier segments. Each denoising step predicts the clean latent and re-noises it for the next step:
\begin{equation}
\begin{aligned}
\tilde x_i^{t_j}&=G_\theta\bigl(z_i^{(t_j)},t_j,\KV_{<i}\bigr),\\
z_i^{(t_{j+1})}&=t_{j+1}\tilde x_i^{t_j}+(1-t_{j+1})\epsilon_{i,j+1},\qquad x_i=\tilde x_i^{t_{k-1}}.
\end{aligned}
\label{eq:generate}
\end{equation}

\paragraph{The memory write.}
After segment $i$ is generated, one more forward of the same transformer writes it into the cache: run on the segment's prediction $\tilde x_i^{t_s}$, it returns the K and V that later segments attend to. Training takes the prediction at a sampled exit step $s$ (Sec.~\ref{sec:memoryobjective}), inference at the last step. As in previous methods~\citep{huang2025self,zhao2026causalforcingpp} the write input is the clean prediction, $\tau=1$ in
\begin{equation}
x_i^\ctx=\tau\tilde x_i^{t_s}+(1-\tau)\xi_i,
\qquad \KV_i=\Kwriter_{\theta,\phi}(x_i^\ctx,\tau,\KV_{<i}),
\label{eq:context}
\end{equation}
where $\xi_i$ is Gaussian noise, $\Kwriter_{\theta,\phi}$ returns the K and V of every layer and $\phi$ is the writer residual of Sec.~\ref{sec:representation}, added to those projections. With $\phi=0$ the write uses the generation projections alone, as in Self Forcing, Causal Forcing++ and SGF. Unlike the K and V computed while denoising, the written K and V are read by every later segment, so the quality of a write is decided by the generations that follow.

\begin{figure}[t]
\centering
\includegraphics[width=\linewidth]{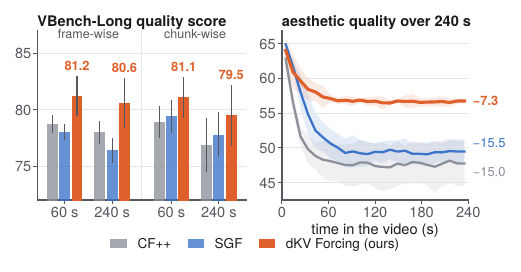}
\vspace{-14pt}
\caption{\textbf{dKV Forcing leads at every long horizon and keeps its quality over time.} Left: VBench-Long quality score. Right: aesthetic quality within 240\,s videos, with its loss from start to end. Mean $\pm$ SD over seeds.}
\label{fig:decay}
\end{figure}
\begin{table}[t]
\caption{\textbf{Writing and generating are different roles.} (a)~Memory against direct gradient on the shared K and V projections (Appendix~\ref{app:controls}). (b)~Memory of frames 0--2 replaced by their generation K/V: latent change of the next frame, an independent seed as reference, and change of the DINO~\citep{caron2021dino} consistency of later frames with the opening (Appendix~\ref{app:repr}).}
\label{tab:roleevidence}
\centering\footnotesize\setlength{\tabcolsep}{4pt}
\begin{tabular}{@{}lccc@{}}
\toprule
(a) checkpoint & norm ratio & cosine & cosine, Adam\\
\midrule
SGF & $1.04$--$1.06$ & $-0.08$ to $-0.14$ & $-0.22$\\
\midrule
(b) checkpoint & next frame & other seed & $\Delta$ consistency\\
\midrule
CF++ initialization & $0.55$ & $0.97$ & $-0.231\pm0.024$\\
SGF & $0.57$ & $1.08$ & $-0.059\pm0.010$\\
\bottomrule
\end{tabular}
\end{table}
\section{Separating generation and memory representations}
\label{sec:representation}
In the standard design the write and the denoising share the K and V projections of every layer. A shared projection then receives two gradients: the \emph{direct gradient} of the segment it denoises and the \emph{memory gradient} of the later segments that read what it wrote. In a trained model the memory gradient is as large as the direct gradient and negatively aligned with it (Table~\ref{tab:roleevidence}a). The KV a segment writes is also not interchangeable with the KV it denoised with: replacing the memory of the first three frames by the K/V of their last denoising evaluation, computed from a noisy input, moves the next frame by about half the distance between two independent seeds and lowers the consistency of later frames with the opening (Table~\ref{tab:roleevidence}b). We therefore add parameters that only the write uses, trained by the memory gradient alone, in two forms.

\begin{figure*}[t]
\centering
\begingroup
\setlength{\tabcolsep}{0.75pt}\renewcommand{\arraystretch}{1.0}
\sffamily\footnotesize
\begin{tabular}{@{}>{\raggedleft\arraybackslash}m{0.105\textwidth}@{\hspace{4pt}}*{6}{>{\centering\arraybackslash}m{0.145\textwidth}}@{}}
 & 0\,s & 12\,s & 24\,s & 36\,s & 48\,s & 59\,s\\[1pt]
CF++ & \multicolumn{6}{@{\hspace{4pt}}>{\centering\arraybackslash}m{\dimexpr0.87\textwidth+7.5pt\relax}@{}}{\includegraphics[width=\linewidth]{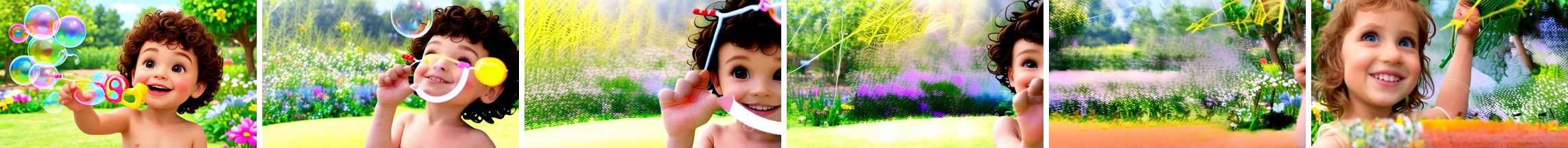}}\\
SGF & \multicolumn{6}{@{\hspace{4pt}}>{\centering\arraybackslash}m{\dimexpr0.87\textwidth+7.5pt\relax}@{}}{\includegraphics[width=\linewidth]{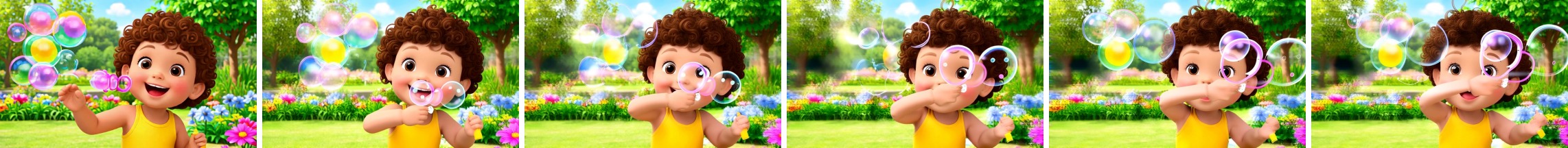}}\\
\textbf{dKV Forcing (ours)} & \multicolumn{6}{@{\hspace{4pt}}>{\centering\arraybackslash}m{\dimexpr0.87\textwidth+7.5pt\relax}@{}}{\includegraphics[width=\linewidth]{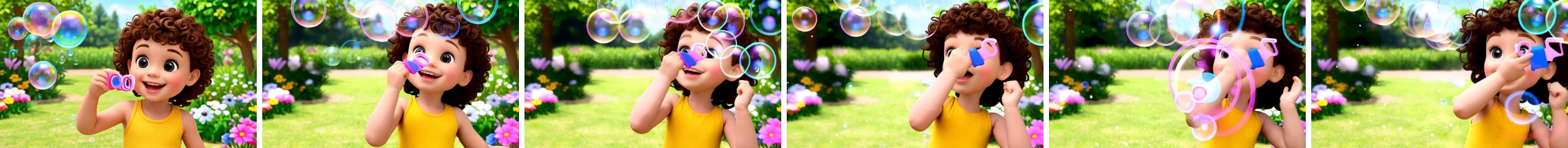}}\\
\end{tabular}
\endgroup

\caption{\textbf{Comparison on 60\,s video generation.} Frames at $0$--$59$\,s for CF++, SGF and dKV Forcing on the MovieGen~\citep{polyak2024moviegen} prompt ``child blowing bubbles''. CF++ loses the subject within the minute, while dKV Forcing keeps subject, layout and motion. More prompts are in Appendix~\ref{app:qualitative}.}
\label{fig:qual60}
\end{figure*}
\paragraph{Dedicated writer.}
In its full form the writer is a separate network. A second copy of the generator computes every write and is trained only by the later segments that read it, while the generator is trained only by its own predictions. The two roles then share no parameters, which isolates what the separation alone is worth (Sec.~\ref{sec:exp-roles}).

\paragraph{Writer residual.}
To avoid the cost of a second network, the form we use throughout keeps one backbone and adds a low-rank residual on its K and V projections, the two tensors that a write stores, active only when a segment is written into the cache. At layer $\ell$ the write uses
\begin{equation}
\begin{aligned}
K_{i,\ell}^\mem&=(W_{K,\ell}+\Delta W_{K,\ell})\,h_{i,\ell}^\mem,\\
V_{i,\ell}^\mem&=(W_{V,\ell}+\Delta W_{V,\ell})\,h_{i,\ell}^\mem,
\end{aligned}
\label{eq:memkv}
\end{equation}
with $h_{i,\ell}^\mem$ the hidden state of the write forward and $a\in\{K,V\}$ below, followed by the same key normalization and positional rotation as generation, which uses $W_{K,\ell}$ and $W_{V,\ell}$ alone. $\Delta W_{a,\ell}=U_{a,\ell}Z_{a,\ell}$ is the low-rank adaptation (LoRA) of \citet{hu2022lora} (rank $64$, $U$ zero-initialized, so training starts from the pretrained write), and only these residual parameters form $\phi$. The residual is switched on only in the forward that writes a segment into the cache and is off in every denoising step. It therefore cannot change how a segment is denoised, only what is stored for later segments to read. The write passes through both the backbone projections and the residual, so the memory gradient reaches both: the residual is the part of the write that generation never uses, and the dedicated writer is its fully separated limit. It adds no forward pass, since the base model already runs this write forward after each segment, and inference only switches the residual on during that write.

\paragraph{Rollout dropout.}
With the residual active in every rollout, fidelity rises but motion falls (Sec.~\ref{sec:exp-roles}): the backbone only ever reads memory written with the residual and can come to rely on it. Standard dropout~\citep{srivastava2014dropout} removes units at random so that a network cannot rely on any one of them, and rollout dropout removes the residual in the same way: it is disabled for an entire training rollout with probability $p$, in both the critic and the generator step, so the backbone must also produce the video, motion included, from writes made without it. This alleviates the loss of motion while fidelity stays above SGF; we use $p=0.5$ (Sec.~\ref{sec:exp-tradeoff}). Training $\phi$ requires the gradient from the future reads of a write back to its writer, which the next section computes.

\section{Single-traversal learning through persistent causal states}
\label{sec:algorithm}
\begin{figure*}[t]
\centering
\begingroup
\setlength{\tabcolsep}{0.75pt}\renewcommand{\arraystretch}{1.0}
\sffamily\footnotesize
\begin{tabular}{@{}>{\raggedleft\arraybackslash}m{0.105\textwidth}@{\hspace{4pt}}*{6}{>{\centering\arraybackslash}m{0.145\textwidth}}@{}}
 & 0\,s & 48\,s & 96\,s & 144\,s & 192\,s & 238\,s\\[1pt]
CF++ & \multicolumn{6}{@{\hspace{4pt}}>{\centering\arraybackslash}m{\dimexpr0.87\textwidth+7.5pt\relax}@{}}{\includegraphics[width=\linewidth]{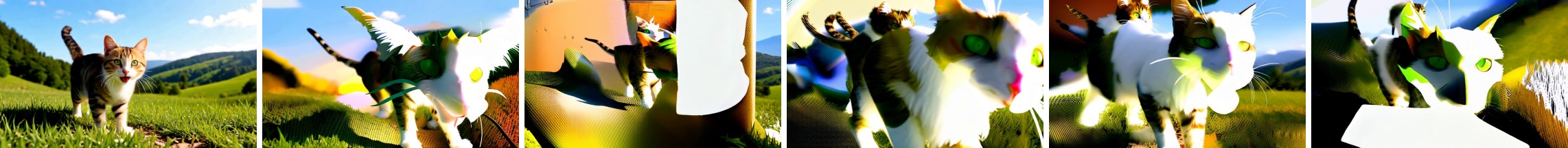}}\\
SGF & \multicolumn{6}{@{\hspace{4pt}}>{\centering\arraybackslash}m{\dimexpr0.87\textwidth+7.5pt\relax}@{}}{\includegraphics[width=\linewidth]{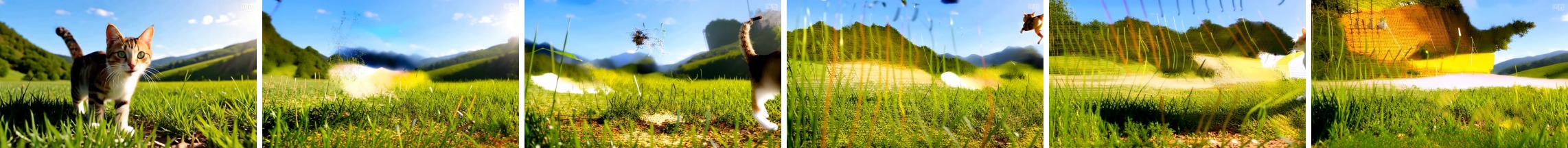}}\\
\textbf{dKV Forcing (ours)} & \multicolumn{6}{@{\hspace{4pt}}>{\centering\arraybackslash}m{\dimexpr0.87\textwidth+7.5pt\relax}@{}}{\includegraphics[width=\linewidth]{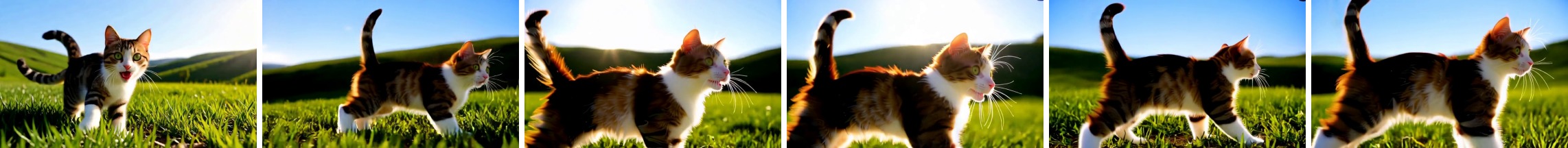}}\\
\end{tabular}
\endgroup

\caption{\textbf{Comparison on 240\,s video generation.} Frames at $0$--$238$\,s, $46$ times the training clip, for CF++, SGF and dKV Forcing on the VBench-Long standard prompt ``a cat running happily''. CF++ has lost the scene and SGF loses the cat, while dKV Forcing keeps the cat, the layout and the lighting across the full four minutes with the cat in motion throughout. More prompts are in Appendix~\ref{app:qualitative}.}
\label{fig:qualmain}
\end{figure*}
\subsection{The objective and its memory gradient}
\label{sec:memoryobjective}
Following~\citep{huang2025self,zhao2026causalforcingpp}, we sample an exit index $s\in\{0,\ldots,k-1\}$, and only the exit step and the write are differentiated. The other sampling steps run without gradient, and the write input is detached from the exit prediction, with $\sg$ the stop-gradient:
\begin{equation}
\bar z_i=\sg\!\left(z_i^{(t_s)}\right),\qquad
\bar x_i^\ctx=\sg(x_i^\ctx).
\label{eq:stop}
\end{equation}
dKV Forcing differentiates the recurrence of Eqs.~\eqref{eq:generate} and~\eqref{eq:context} at the exit step under the distribution matching distillation (DMD) objective $L_\DMD(\tilde X)$~\citep{yin2024dmd,yin2024dmd2,huang2025self,zhao2026causalforcingpp} on the exit predictions $\tilde X=(\tilde x_0^{t_s},\ldots,\tilde x_n^{t_s})$, with $\tilde x_i^{t_s}=G_\theta(\bar z_i,t_s,\KV_{<i})$. Its gradient $v_i=\partial L_\DMD/\partial\tilde x_i^{t_s}$ is the difference between the score of a critic trained alongside on the generator's own samples and the score of the frozen teacher. The differential of the exit prediction is
\begin{equation}
\diff\tilde x_i^{t_s}=D_\theta \tilde x_i^{t_s}\,\diff\theta+
\sum_{j<i}D_{\KV_j}\tilde x_i^{t_s}\,\diff\KV_j,
\label{eq:differential}
\end{equation}
where $D_a$ denotes the Jacobian with respect to $a$. Self Forcing and Causal Forcing++ treat the history as $\sg(\KV_{<i})$~\citep{huang2025self,zhao2026causalforcingpp}, which removes the second term. The model then learns to generate given its history, but the operations that built that history receive no gradient, and a residual used only for writing would receive none at all. dKV Forcing preserves this term, and its gradient is the exact gradient of this objective (Proposition~\ref{prop:exact}).

\label{sec:causaladjoint}
To compute the gradient that this term sends to each write, let $\lambda_i$ be the gradient that reaches $\KV_i$ from all later segments. Reversing the recurrence gives
\begin{equation}
\lambda_i
=\sum_{r=i+1}^{n}(D_{\KV_i}\tilde x_r^{t_s})^\transpose v_r
+\sum_{r=i+1}^{n}(D_{\KV_i}\KV_r)^\transpose\lambda_r.
\label{eq:adjoint}
\end{equation}
The first sum comes from the segments that read $\KV_i$, the second from the later writes that depend on it. Because $\phi$ is used only by the writer,
\begin{align}
\nabla_\theta L_\DMD&=\sum_{i=0}^{n}(D_\theta \tilde x_i^{t_s})^\transpose v_i+
\sum_{i=0}^{n}(D_\theta\KV_i)^\transpose\lambda_i,
\label{eq:theta}\\
\nabla_\phi L_\DMD&=\sum_{i=0}^{n}(D_\phi\KV_i)^\transpose\lambda_i,
\label{eq:phi}
\end{align}
so $\phi$ is optimized for what future consumers need to read, including effects mediated by later writes. Unrolling Eq.~\eqref{eq:adjoint} gives a finite sum over chains of writes, which reverse mode evaluates through vector--Jacobian products (Appendix~\ref{app:causalfactorization}).

\subsection{Persistent pages and handles}
\label{sec:pages}
\paragraph{Failure of the mutable cache.}
\label{sec:naiveopen}
Even with a detached cache, every exit forward keeps about $5$\,GiB of activations per frame, so Self Forcing, Causal Forcing and SGF all checkpoint every block and replay it in backward. With a detached cache a replayed forward is released as soon as its backward is done (Figure~\ref{fig:pagecontract}(a)). Enabling the gradient through the cache breaks this. The cache is a mutable buffer that each forward writes in place, so replay writes the same buffer again, and this write attaches the replayed graph, with its recomputed activations, to the cache. The cache outlives the replay, so the activations of every replayed forward accumulate and exhaust GPU memory even on 5\,s clips (Figure~\ref{fig:pagecontract}(b)). The memory profile confirms it: for the generator alone, on one 5\,s clip, the forward ends at $15.4$\,GiB, the peak of $57.2$\,GiB is reached in backward, and $42.3$\,GiB stay allocated after backward has finished, whereas a graph retained for the gradient path would peak at the end of forward and be released by backward (Appendix~\ref{app:replay}).

\begin{figure}[t]
\centering
\includegraphics[width=\linewidth]{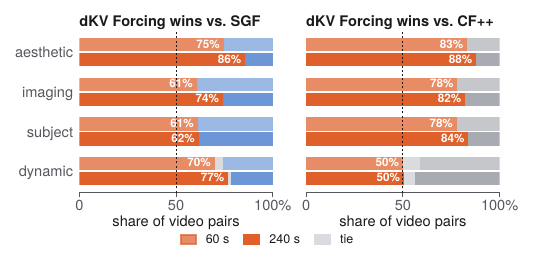}
\caption{\textbf{Head to head, dKV Forcing wins most video pairs.} Share of video pairs of the same prompt and training seed in which dKV Forcing has the higher VBench-Long score, against SGF (left) and CF++ (right).}
\label{fig:h2h}
\end{figure}
\begin{table*}[t]
\caption{\textbf{One traversal carries the memory gradient at the lowest cycle time and resident memory.} Training cost on one node (4 H100, batch 2 per GPU, bf16, 21-frame clips; median over steady cycles; max over ranks). Cycle = 5 critic + 1 generator update. ``Gen.\ update'' = rollout, gradient-bearing forwards, teacher/critic scoring and backward of one generator update. Bold and underline: best and second best among the methods that carry the memory gradient; CF++ carries none and is listed as the reference.}
\label{tab:perf}
\centering\small
\resizebox{\linewidth}{!}{\begin{tabular}{@{}lcccccc@{}}
\toprule
& \multicolumn{3}{c}{time (s)} & \multicolumn{3}{c}{GPU memory (GiB)}\\
\cmidrule(lr){2-4}\cmidrule(lr){5-7}
method & cycle & gen.\ update & vs.\ two-pass & gen.\ update peak & whole-run peak & resident\\
\midrule
two-pass reconstruction & 122.8 & \underline{35.7} & -- & 69.9 & 75.4 & 40.7\\
CF++ (writes detached) & 104.7 & 31.8 & $-14.7\%$ & 55.9 & 55.9 & 40.7\\
mutable cache (write differentiated) & \multicolumn{6}{c}{out of memory in the first generator backward}\\
mutable cache (replay writes repaired) & 115.5 & 41.3 & $-5.9\%$ & 73.1 & 73.1 & 40.7\\
dKV Forcing (default) & \textbf{97.3} & \textbf{33.9} & $-20.8\%$ & \underline{62.1} & \underline{62.1} & \textbf{29.2}\\
dKV Forcing (hidden-state pages) & \underline{105.9} & 43.1 & $-13.8\%$ & \textbf{56.1} & \textbf{56.1} & \underline{29.3}\\
\bottomrule
\end{tabular}}
\end{table*}
\begin{table*}[t]
\caption{\textbf{The memory gradient raises fidelity, the separate writer raises it again, and rollout dropout raises the motion.} VBench-Long 60\,s, frame-wise, mean $\pm$ SD over training seeds $\{1,10,20\}$, as in Table~\ref{tab:main}. ``Memory gradient'' says which parameters receive gradient from future reads. $^{*}$The rank-64 writer residual with rollout dropout $0.5$. Among the four rows above the rule, best in bold and second best underlined.}
\label{tab:roles}
\centering\small
\setlength{\tabcolsep}{4.3pt}
\begin{tabular}{@{}llccccc@{}}
\toprule
configuration & memory gradient & Quality & aesthetic & dynamic & imaging & subject\\
\midrule
CF++ (detached writes) & none & $78.7\pm0.8$ & $52.7\pm1.4$ & $\mathbf{76.5}\pm19.0$ & $63.3\pm1.6$ & $94.7\pm1.3$\\
SGF & backbone & $78.0\pm0.7$ & $56.2\pm1.4$ & $38.5\pm6.9$ & $67.4\pm2.2$ & $96.5\pm0.4$\\
dKV Forcing (ours)$^{*}$ & backbone + residual & $\mathbf{81.2}\pm1.8$ & $\underline{59.8}\pm1.3$ & $\underline{71.4}\pm34.5$ & $\underline{68.8}\pm1.3$ & $\underline{96.6}\pm1.2$\\
dKV Forcing without rollout dropout & backbone + residual & $\underline{79.1}\pm1.4$ & $\mathbf{62.8}\pm0.4$ & $21.8\pm17.5$ & $\mathbf{69.7}\pm0.6$ & $\mathbf{98.2}\pm0.2$\\
\midrule
separate writer network (second 1.3B copy) & writer network & $80.2\pm1.1$ & $64.9\pm0.6$ & $26.0\pm18.2$ & $71.6\pm0.7$ & $98.6\pm0.4$\\
\bottomrule
\end{tabular}
\end{table*}
\paragraph{Pages and handles.}
One tensor thus does both jobs of the cache: it stores the values that later segments read, and it carries the gradient back to their writer. dKV Forcing separates the two (Figure~\ref{fig:pagecontract}(c)). \emph{(1)~The values stay detached.} Each layer has a page table $P$ with one slot per segment. The writer output $\KV_j$ is copied into its slot once, without gradient, and is only read afterwards, so the history $\Hist_i=(\KV_0,\ldots,\KV_{i-1})$ is an ordered tuple of immutable pages and a write appends $\KV_i$ without changing any earlier entry (Appendix~\ref{app:execution}). \emph{(2)~A zero-valued handle carries the gradient.} Each writer output is also connected, in the autograd graph, to a new tensor $h_j$: an expanded zero scalar with the page's shape whose backward passes the incoming gradient unchanged to $\KV_j$. A handle holds no page data and is never written, so nothing accumulates on it during replay. A segment reads the stored pages of its prefix in forward, and in backward the gradient of each page is routed through $h_j$ to the writer that produced it (code in Appendix~\ref{app:code}).

\subsection{Local rematerialization in one temporal traversal}
\label{sec:lifetime}
Replay must also recompute exactly what the forward computed. We checkpoint in groups: every graph-carrying forward, $G_\theta$ at the exit step or $\Kwriter_{\theta,\phi}$, is split into groups of five blocks, and each group keeps only the hidden state $h_{i,g}$ that enters it and recomputes its activations in backward. By then the page table holds the final history $\Hist_n$, although segment $i$ saw only its first $i$ pages in forward. A committed page is never modified, and the snapshot of segment $i$ exposes only those $i$ pages, so the recomputation reads what the forward read. With $\mathcal T_g(h;\Hist)$ the computation of group $g$ on the hidden state $h$ and the history $\Hist$, and $\Pi_i$ the projection that keeps the first $i$ writes,
\begin{equation}
\mathcal T_g(h_{i,g};\Pi_i\Hist_n)=\mathcal T_g(h_{i,g};\Hist_i).
\label{eq:rematerializationidentity}
\end{equation}
Each group therefore recovers its original outputs and local Jacobians, and replacing the stored activations by recomputed ones leaves the gradient unchanged.

\begin{proposition}[Exact single-traversal differentiation]
\label{prop:exact}
Fix the recorded inputs, model parameters, visibility, positions and random choices across forward execution and backward replay. Assume that each role computation is a pure function of its explicit inputs, state extension preserves the historical prefix and its differential (Eqs.~\eqref{eq:prefixinvariant} and~\eqref{eq:extensionjacobian} in Appendix~\ref{app:execution}), and replay satisfies Eq.~\eqref{eq:rematerializationidentity}. Then the traversal (Algorithm~\ref{alg:kvforcing} in Appendix~\ref{app:execution}) computes the same predictions and parameter gradients as the uncheckpointed recurrence of Sec.~\ref{sec:memoryobjective}, in exact arithmetic.
\end{proposition}
Appendix~\ref{app:proofs} gives the proof.

\section{Experiments}
\label{sec:experiments}
\subsection{Setup}
\label{sec:exp-setup}
All runs fine-tune the 1.3B causal generator of Causal Forcing++~\citep{zhao2026causalforcingpp} (Wan2.1-T2V-1.3B) with its DMD objective and frozen 14B teacher, from its released frame-wise and chunk-wise initializations: $1500$ generator updates at a global batch of $8$ clips of $21$ latent frames ($\approx5$\,s). The rank-64 writer residual adds $11.8$M parameters ($0.9\%$ of the generator); the backbone learning rate is $2\times10^{-6}$ and that of the residual $10^{-4}$.

\emph{CF++}, the state-of-the-art few-step causal generator we start from, continues its original training with detached writes for the same $1500$ updates. \emph{SGF} adds the memory gradient of Self Gradient Forcing~\citep{zhuang2026sgf} and is our primary baseline; we compute its gradient with our single traversal, which matches its public two-pass implementation (Sec.~\ref{sec:exp-algo}). \emph{dKV Forcing} further adds the writer residual. Frame-wise it is trained with rollout dropout $0.5$; chunk-wise the residual does not lower motion and is used without dropout. All configurations share the initialization, data order, schedule, seeds and inference window, and differ only in how the write is trained.

We evaluate with VBench-Long~\citep{huang2024vbenchpp} at 60\,s and 240\,s ($963$ latent frames, $46$ times a training clip) on its $326$ standard prompts, none of which appears among the training prompts, and report a mean $\pm$ SD over training seeds. Frame-wise inference keeps the first $4$ frames and a sliding window of $16$. Appendix~\ref{app:setupdetails} gives the remaining details and Table~\ref{tab:vbench5s} the VBench scores at 5\,s~\citep{huang2024vbench}.

\begin{table}[t]
\caption{\textbf{Gains at matched motion.} Paired differences of dKV Forcing against CF++ and SGF over video pairs of the same prompt and training seed whose dynamic fractions differ by at most $10$ points, with $95\%$ bootstrap intervals over prompts. Text alignment is the VBench overall-consistency score. Every difference is significant after Holm correction ($77$--$113$ pairs against CF++, $44$--$107$ against SGF).}
\label{tab:matched}
\centering\footnotesize\setlength{\tabcolsep}{2.5pt}
\resizebox{\linewidth}{!}{\begin{tabular}{@{}lcccc@{}}
\toprule
& \multicolumn{2}{c}{60\,s} & \multicolumn{2}{c}{240\,s}\\
\cmidrule(lr){2-3}\cmidrule(lr){4-5}
metric & vs.\ CF++ & vs.\ SGF & vs.\ CF++ & vs.\ SGF\\
\midrule
aesthetic & $+6.5$ [$5.1$, $7.9$] & $+4.1$ [$2.9$, $5.2$] & $+9.5$ [$8.1$, $11.0$] & $+5.8$ [$4.5$, $7.1$]\\
imaging & $+5.4$ [$3.8$, $7.1$] & $+3.4$ [$2.0$, $4.8$] & $+7.2$ [$5.6$, $8.8$] & $+5.4$ [$4.0$, $6.9$]\\
subject & $+1.1$ [$0.7$, $1.5$] & $+1.0$ [$0.6$, $1.4$] & $+2.2$ [$1.7$, $2.7$] & $+1.1$ [$0.7$, $1.4$]\\
text alignment & $+2.1$ [$1.1$, $3.2$] & $+1.4$ [$0.6$, $2.2$] & $+4.1$ [$2.9$, $5.5$] & $+2.6$ [$1.8$, $3.5$]\\
\bottomrule
\end{tabular}}
\end{table}
\subsection{Benchmark results}
\label{sec:exp-main}
dKV Forcing has the highest VBench quality score at both horizons and in both generators (Table~\ref{tab:main}; Figures~\ref{fig:qual60} and~\ref{fig:qualmain}). Frame-wise it leads SGF by $3.2$ points at 60\,s and $4.2$ at 240\,s, and chunk-wise by $1.7$ at both horizons. The lead comes from fidelity: aesthetic quality is $3.6$ points above SGF at 60\,s and $6.1$ at 240\,s, at a mean dynamic degree of $71$ against $39$. The lead widens with the horizon because quality decays less over time: over a 240\,s video dKV Forcing loses $7.3$ points of aesthetic quality, less than half the loss of CF++ and SGF (Figure~\ref{fig:decay}). Head to head on the same prompt and seed, dKV Forcing has the higher aesthetic quality than SGF in $75\%$ of the 60\,s videos and $86\%$ of the 240\,s videos (Figure~\ref{fig:h2h}). Nor do the gains come at the cost of motion: on video pairs of matched motion, aesthetic quality is $4.1$ points above SGF at 60\,s and $5.8$ at 240\,s, and imaging quality, subject consistency and text alignment improve as well (Table~\ref{tab:matched}).

\begin{figure}[t]
\centering
\includegraphics[width=\linewidth]{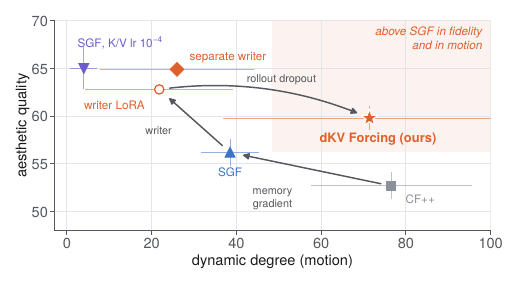}
\caption{\textbf{Only dKV Forcing is above SGF in both fidelity and motion.} The rows of Table~\ref{tab:roles}, and SGF trained at the writer's learning rate, on the fidelity--motion plane (60\,s, mean $\pm$ SD over seeds).}
\label{fig:plane}
\end{figure}
\subsection{Effects of memory gradients and writer adaptation}
\label{sec:exp-roles}
The memory gradient and the separate writer contribute two distinct gains at 60\,s (Table~\ref{tab:roles}, Figure~\ref{fig:plane}). First, the memory gradient alone (CF++$\to$SGF) is worth $3.5$ points of aesthetic quality while dynamic degree halves, so the VBench quality score does not rise. Second, separating the writer adds a gain on top of it. With a dedicated writer network the two roles share no parameters: the writer receives the memory gradient alone and the generator the direct gradient alone, both at the backbone learning rate. Aesthetic quality then rises over SGF by $8.7$ points, and the gain holds at 240\,s ($63.4$ against $51.3$). Most of this gain survives in the lightweight form: the rank-64 residual recovers $6.6$ of the $8.7$ points with $0.9\%$ of the parameters. In contrast, a faster learning rate is no substitute for the separation: training the shared K and V projections of SGF at the writer's rate of $10^{-4}$ freezes the videos, with a dynamic degree of $4.0\pm3.2$ at 60\,s and $5.2\pm4.4$ at 240\,s. Finally, rollout dropout raises the motion: the mean dynamic degree returns to $71$ with aesthetic quality still above SGF, and the quality score becomes the highest of the table.\label{sec:exp-tradeoff}

\subsection{Single traversal versus two-pass reconstruction}
\label{sec:exp-algo}
\paragraph{Same gradient, less time and memory.}
In fp64 the relative gradient error of every dKV Forcing execution against the two-pass reference is at most $4.1\times10^{-12}$ (Appendix~\ref{app:implementation}). Table~\ref{tab:perf} compares implementations of this gradient. The full memory gradient adds only $7\%$ to the generator update of detached-write training. Against the two-pass reconstruction, the default execution lowers the generator-update peak from $69.9$ to $62.1$\,GiB and, because pages live only for one traversal, resident memory by $28\%$. Pages also speed up every rollout, critic updates included, so the cycle is $20.8\%$ shorter. For comparison, once every replayed block writes a private copy that is released after its backward, the mutable cache also trains, but at a $19\%$ longer cycle and an $11$\,GiB higher peak, which locates its failure in the replay (Appendix~\ref{app:replay}). Finally, hidden-state pages lower the peak further, to $56.1$\,GiB (Appendix~\ref{app:pageimpl}).

\begin{figure}[t]
\centering
\includegraphics[width=\linewidth]{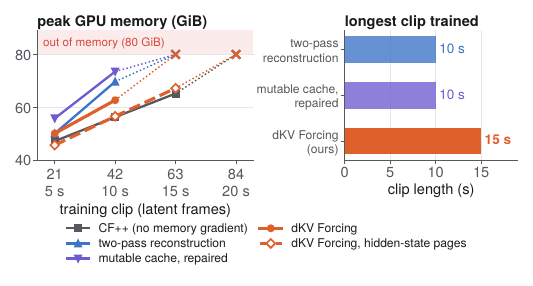}
\vspace{-14pt}
\caption{\textbf{dKV Forcing trains 15\,s clips; the two-pass reconstruction stops at 10\,s} (full trainer, batch 1 per GPU). At 63 frames dKV Forcing trains with hidden-state pages.}
\label{fig:tscale}
\end{figure}
\paragraph{Longer training clips.}
Tuning on long videos is reported to be critical for long generation~\citep{yang2025longlive}, so Figure~\ref{fig:tscale} repeats the measurement for clips of $21$ to $84$ latent frames, with whole-clip attention. At $63$ frames ($15$\,s) the two-pass reconstruction runs out of memory in the generator backward, while dKV Forcing with hidden-state pages trains at $67.2$\,GiB, $2$\,GiB above training without the memory gradient.

\section{Conclusion}
dKV Forcing learns what to write into memory from the gradients of its future reads. A writer residual separates memory writing from generation, and persistent KV pages compute these gradients in one traversal. Trained on 5\,s clips, it holds subject, layout and motion for four minutes. \textbf{Limitations.} We study one base generator, Causal Forcing++ on Wan2.1-1.3B, and leave other generators and the combination with orthogonal long-video techniques such as LongLive~\citep{yang2025longlive} to future work.

{\small
\bibliographystyle{ieeenat_fullname}
\bibliography{refs}
}

\clearpage
\appendix
\appendixtitle
\addtocontents{toc}{\protect\setcounter{tocdepth}{2}}
{\renewcommand{\contentsname}{Contents of the appendix}\makeatletter\renewcommand*\l@subsection{\@dottedtocline{2}{1.5em}{3.1em}}\renewcommand*\numberline[1]{\hb@xt@\@tempdima{\hfil#1\hskip 1em}}\makeatother\tableofcontents}
\section{Derivations and correctness}
\label{app:proofs}
\subsection{\rev{Persistent state and local rematerialization}}
\label{app:execution}
\begin{figure}[t]
\centering
\begingroup
\definecolor{kvInk}{HTML}{24374A}
\definecolor{kvGray}{HTML}{778595}
\definecolor{kvGold}{HTML}{5B7083}
\definecolor{kvTeal}{HTML}{E0602B}
\definecolor{kvPurple}{HTML}{8661A4}
\resizebox{\linewidth}{!}{\begin{tikzpicture}[x=1cm,y=1cm,>=stealth,font=\sffamily\footnotesize,line cap=round,line join=round,
 box/.style={rounded corners=2pt,draw=kvInk!30,fill=white,align=center,inner sep=3pt,minimum height=0.5cm},
 gen/.style={box,fill=kvGold!10,draw=kvGold!50},
 mem/.style={box,fill=kvTeal!9,draw=kvTeal!50},
 slot/.style={draw=kvTeal!80,fill=kvTeal!55,minimum width=0.8cm,minimum height=0.42cm,inner sep=0pt,font=\sffamily\scriptsize,text=white},
 free/.style={draw=kvInk!25,fill=kvInk!5,minimum width=0.8cm,minimum height=0.42cm,inner sep=0pt,font=\sffamily\scriptsize,text=kvGray},
 handle/.style={draw=kvPurple,fill=white,circle,minimum size=0.36cm,inner sep=0pt,font=\sffamily\scriptsize},
 gbuf/.style={draw=kvPurple,fill=kvPurple!22,minimum width=0.5cm,minimum height=0.24cm,inner sep=0pt},
 graphbox/.style={draw=kvPurple!80,dash pattern=on 2.2pt off 1.6pt,rounded corners=3pt,line width=0.8pt},
 lab/.style={font=\sffamily\scriptsize,text=kvGray,align=center,inner sep=1pt},
 commit/.style={->,draw=kvTeal,line width=0.9pt},
 readonly/.style={->,draw=kvTeal,line width=0.6pt,dash pattern=on 1.8pt off 1.2pt},
 represents/.style={draw=kvInk!35,dash pattern=on 0.8pt off 1.2pt,line width=0.5pt},
 grad/.style={->,draw=kvPurple,line width=0.85pt,dash pattern=on 3pt off 2pt},
 flow/.style={->,draw=kvInk!75,line width=0.7pt}]
\path[use as bounding box] (0,-0.2) rectangle (8.4,5.6);
\draw[graphbox] (0.05,1.85) rectangle (8.35,5.55);
\node[lab,text=kvPurple,anchor=south east] at (8.35,5.57) {autograd graph};
\node[gen,minimum width=1.9cm] (g) at (2.3,5.05) {$G_\theta$, segment $i$};
\node[box,minimum width=4.7cm,draw=kvTeal!50] (cat) at (2.85,4.25) {};
\node[inner sep=0pt] at (0.75,4.25) {$[$};
\node[inner sep=0pt] (c0) at (1.0,4.25) {$\rev{p}_0$};
\node[inner sep=0pt] (c1) at (2.1,4.25) {$\rev{p}_1$};
\node[inner sep=0pt] at (2.8,4.25) {$\cdots$};
\node[inner sep=0pt] (ci) at (3.5,4.25) {$\rev{p}_{i-1}$};
\node[inner sep=0pt] at (4.0,4.25) {$|$};
\node[inner sep=0pt] at (4.55,4.25) {$K_i,V_i$};
\node[inner sep=0pt] at (5.0,4.25) {$]$};
\node[lab,anchor=south west,align=left] at (3.45,4.52) {attention K/V\\(rematerialized)};
\draw[flow] (2.3,4.5) -- (g);
\node[mem,minimum width=1.9cm] (w) at (6.7,5.05) {$\Kwriter_{\theta,\phi}$, segment $i{-}1$};
\node[box,minimum width=1.0cm] (kv) at (6.7,4.25) {$\KV_{i-1}$};
\draw[flow] (w) -- (kv);
\foreach \x/\n in {1.0/0,2.1/1,4.3/{i-1}}{
 \node[handle] (h\n) at (\x,2.45) {$h_{\n}$};
 \node[gbuf] (gb\n) at (\x,3.2) {};
 \draw[grad] (gb\n.south) -- (h\n.north);}
\node[lab] at (3.2,2.45) {$\cdots$};
\node[lab,text=kvPurple,align=left,anchor=west] at (2.45,3.2) {gradient\\buffers};
\node[lab,text=kvPurple] at (3.2,2.08) {handles};
\draw[grad] ([xshift=3pt]c0.south) -- (gb0.north);
\draw[grad] ([xshift=3pt]c1.south) -- (gb1.north);
\draw[grad] ([xshift=3pt]ci.south) -- (gbi-1.north);
\draw[grad] (hi-1.east) .. controls (5.4,2.6) and (6.0,3.6) .. (kv.south west) node[pos=0.78,above left=-2pt,lab,text=kvPurple] {gradient};
\draw[kvInk!35] (0.3,0.72) rectangle (8.1,1.18);
\foreach \x/\n in {1.0/0,2.1/1,4.3/{i-1}}{ \node[slot] (p\n) at (\x,0.95) {$\rev{p}_{\n}$}; \draw[represents] (h\n.south) -- (p\n.north); }
\node[lab] at (3.2,0.95) {$\cdots$};
\node[free] at (5.4,0.95) {free};
\node[free] at (6.5,0.95) {free};
\node[lab] at (7.5,0.95) {$\cdots$};
\node[lab,anchor=east] at (8.1,1.5) {page table};
\draw[commit] (kv.south) .. controls (6.7,3.0) and (4.3,2.3) .. (pi-1.north);
\node[lab,text=kvTeal,anchor=west] at (4.75,1.5) {commit once};
\draw[readonly] (0.7,1.17) -- ([xshift=-3pt]c0.south);
\draw[readonly] (1.8,1.17) -- ([xshift=-3pt]c1.south);
\draw[readonly] (4.0,1.17) -- ([xshift=-3pt]ci.south);
\node[lab,text=kvTeal] at (1.25,1.5) {read-only};
\draw[decorate,decoration={brace,amplitude=3pt},kvInk!60] (4.72,0.6) -- (0.58,0.6);
\node[lab,anchor=north] at (2.65,0.48) {snapshot of segment $i$};
\end{tikzpicture}}
\endgroup
\vspace{-8mm}
\caption{\textbf{Page storage and gradient flow in one transformer layer.} The page table (bottom row) stores each write in a separate, write-once slot outside the autograd graph (dashed frame), committed once (orange solid arrow). A \rev{zero-valued} handle (purple circle) preserves the gradient connection to the writer without retaining the page data. Each forward reads a snapshot of the visible pages (orange dashed arrows), while backward returns an independent, contiguous gradient buffer (purple rectangles) to each page's writer (purple dashed arrows).}
\label{fig:pagetable}
\end{figure}
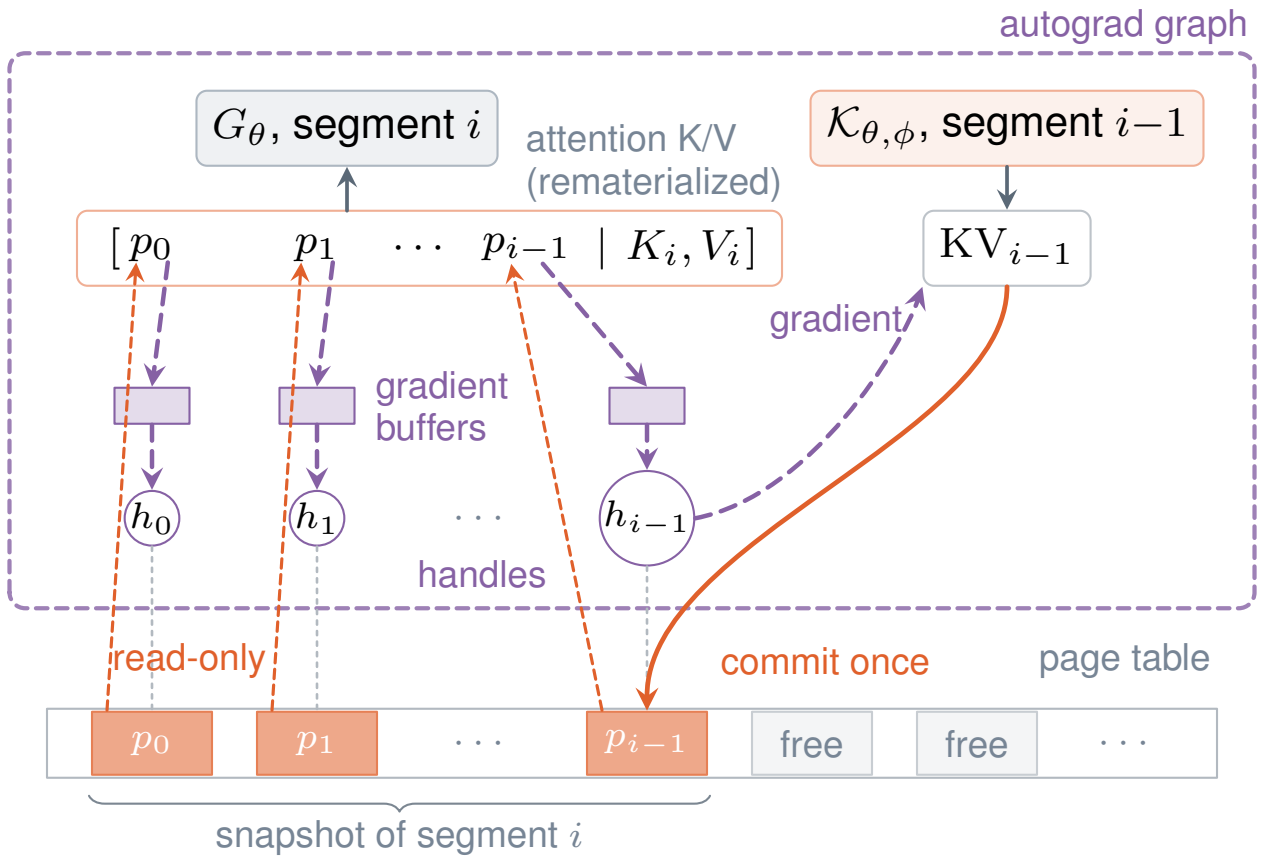

\begin{algorithm*}[t]
\caption{dKV Forcing: single-traversal learning of specialized memory}
\label{alg:kvforcing}
\begin{algorithmic}[1]
\Require Parameters $\theta,\phi$; schedule $\{t_j\}$; context time $\tau$; $T=n+1$
\State Sample exit index $s$ and rollout random variables; allocate a page table $P$ with $T-1$ write-once slots
\For{$i=0,\ldots,n$}
  \State $\KV_{<i}\gets$ snapshot of $P$: pages $0,\ldots,i-1$ and handles $h_0,\ldots,h_{i-1}$ \Comment{Eq.~\eqref{eq:prefixinvariant}}
  \State $z_i^{(t_0)}\gets\epsilon_i$
  \For{$j=0,\ldots,s-1$} \Comment{no gradient, Eq.~\eqref{eq:generate}}
    \State $z_i^{(t_{j+1})}\gets t_{j+1}\,G_\theta(z_i^{(t_j)},t_j,\KV_{<i})+(1-t_{j+1})\,\epsilon_{i,j+1}$
  \EndFor
  \State $\bar z_i\gets\sg(z_i^{(t_s)})$ \Comment{Eq.~\eqref{eq:stop}}
  \State $\tilde x_i^{t_s}\gets G_\theta(\bar z_i,t_s,\KV_{<i})$ \Comment{checkpointed, Eq.~\eqref{eq:groupmap}}
  \State $\bar x_i^\ctx\gets\sg(\tau\tilde x_i^{t_s}+(1-\tau)\xi_i)$ \Comment{Eqs.~\eqref{eq:context}, \eqref{eq:stop}}
  \If{$i<n$} \Comment{no segment reads the last write}
    \State $\KV_i\gets\Kwriter_{\theta,\phi}(\bar x_i^\ctx,\tau,\KV_{<i})$ \Comment{checkpointed}
    \State Commit $\KV_i$ to slot $i$ of $P$; keep handle $h_i$ in the graph
  \EndIf
\EndFor
\State $g_\theta,g_\phi\gets\nabla_{\theta,\phi}L_\DMD(\tilde x_0^{t_s},\ldots,\tilde x_n^{t_s})$ by reverse mode \Comment{Eqs.~\eqref{eq:adjoint}--\eqref{eq:phi}, \eqref{eq:rematerializationidentity}}
\State Update $\theta,\phi$; release $P$ and the graph
\end{algorithmic}
\end{algorithm*}
\paragraph{State extension.}
For layer $\ell\in\{1,\ldots,B\}$ the writer emits a page $\rev{p}_{i,\ell}=(K_{i,\ell}^{\mem},V_{i,\ell}^{\mem})$, \rev{the K and V of that layer, with K after the key normalization and positional rotation,} and $\KV_i=(\rev{p}_{i,1},\ldots,\rev{p}_{i,B})$. \rev{Its values are stored in a detached write-once slot, and a handle keeps its gradient connection (Sec.~\ref{sec:pages}).} \rev{Only the writer outputs are retained in the cache, and generation reads} the historical pages together with the transient KV of its own tokens. Writing $w=(\theta,\phi)$, the history before segment $i$ and its extension by the write of segment $i$ are
\begin{equation}
\begin{aligned}
\Hist_i&=(\KV_0,\ldots,\KV_{i-1}),\\
\Hist_{i+1}&=\Hist_i\mathbin{\Vert}\KV_i,\qquad \KV_i=\Kwriter_{\theta,\phi}(\bar x_i^\ctx,\tau,\Hist_i).
\end{aligned}
\label{eq:stateextension}
\end{equation}
The transition never changes earlier entries:
\begin{equation}
\Pi_i\Hist_r=\Hist_i\quad(r\ge i),
\label{eq:prefixinvariant}
\end{equation}
where $\Pi_i$ is the projection that keeps the first $i$ writes of a history. After vectorizing the state tuples, its differential is
\begin{equation}
\diff\Hist_{i+1}
=\begin{bmatrix}I\\D_{\Hist_i}\KV_i\end{bmatrix}\diff\Hist_i
+\begin{bmatrix}0\\D_w\KV_i\end{bmatrix}\diff w.
\label{eq:extensionjacobian}
\end{equation}
\rev{The identity block preserves the differentials of the previous writes, and the lower block captures the dependence of the new write on the history and the parameters.} 
In backward the page slices of the history gradient are copied into separate contiguous buffers, so that each page gradient has its own lifetime, independent of the whole history-gradient buffer. Later writes extend the table and leave the visible prefix of earlier snapshots unchanged (Figure~\ref{fig:pagetable}).

\paragraph{Group map.}
\rev{We partition the blocks of the transformer into groups and apply activation checkpointing to each group. As in Sec.~\ref{sec:lifetime}, $\mathcal T_g$ denotes the computation of group $g$, in the exit forward $G_\theta$ or in the write $\Kwriter_{\theta,\phi}$:
\begin{equation}
h_{i,g+1}=\mathcal T_g(h_{i,g};\Hist_i),
\label{eq:groupmap}
\end{equation}
where $h_{i,g}$ is the hidden state entering group $g$; in the write, the group also outputs the pages of its blocks.} Only the entry states are kept for backward, and the internal attention and feed-forward activations are recomputed from them. By the time backward runs, the page table holds the final history $\Hist_n$, and each snapshot still exposes only the prefix that segment $i$ saw in forward, which gives Eq.~\eqref{eq:rematerializationidentity}. \rev{Recomputation requires the same inputs as the original computation~\citep{pytorch2026checkpoint}. A forward that reads a mutable buffer which is later overwritten or shifted would not recover its original history in replay. Pages are committed outside the checkpointed regions and never modified, so the recomputation reads the history of the forward, and replay attaches no graph to the cache.}

\subsection{Finite causal factorization of the memory gradient}
\label{app:causalfactorization}
Vectorize and stack all memory outputs as $\rev{\mathbf m}=\operatorname{col}(\operatorname{vec}\KV_0,\ldots,\operatorname{vec}\KV_n)$, where the terminal write $\KV_n$ is a formal node with zero cotangent that Algorithm~\ref{alg:kvforcing} omits. With the recorded inputs fixed, each write $\KV_i$ depends on the parameters $w$ and on the earlier writes $\KV_{<i}$. Stacking the writes gives the causal recurrence $\rev{\mathbf m}=\rev{\mathbf R}_w(\rev{\mathbf m})$, whose component $i$ uses only components $j<i$. Let $\mathsf A=D_{\rev{\mathbf m}}\rev{\mathbf R}_w$ and $\rev{\mathsf J}_w=D_w\rev{\mathbf R}_w$, with historical inputs held independent when taking the local parameter derivative. Linearizing gives
\begin{equation}
\diff\rev{\mathbf m}=\mathsf A\,\diff\rev{\mathbf m}+\rev{\mathsf J}_w\,\diff w,
\qquad
\diff\rev{\mathbf m}=(I-\mathsf A)^{-1}\rev{\mathsf J}_w\,\diff w.
\label{eq:linearizedmemory}
\end{equation}
Causality makes $\mathsf A$ strictly block lower triangular. Every product of $q$ blocks corresponds to a chain of $q$ edges between increasing segment indices. No such chain has $n+1$ edges, so $\mathsf A^{n+1}=0$ and
\begin{equation}
(I-\mathsf A)^{-1}=\sum_{q=0}^{n}\mathsf A^q.
\label{eq:nilpotentseries}
\end{equation}
For the stacked generation map $\mathbf F_\theta(\rev{\mathbf m})$ and output cotangent $\mathbf v$, define $\boldsymbol\eta=(D_{\rev{\mathbf m}}\mathbf F_\theta)^\transpose\mathbf v$. Then
\begin{align}
\diff L\rev{{}_\DMD}
&=\mathbf v^\transpose D_w\mathbf F_\theta\,\diff w
 +\boldsymbol\eta^\transpose\diff\rev{\mathbf m},\\
\nabla_w L\rev{{}_\DMD}
&=(D_w\mathbf F_\theta)^\transpose\mathbf v
 +\rev{\mathsf J}_w^\transpose(I-\mathsf A^\transpose)^{-1}\boldsymbol\eta.
\label{eq:factorizedgradient}
\end{align}
The direct-generation term has no $\phi$ component. \rev{Expanding blockwise recovers} Eqs.~\eqref{eq:adjoint}--\eqref{eq:phi}. The inverse is a finite sum because the recurrence is causal, and reverse-mode differentiation evaluates these paths without constructing $\mathsf A$.

The cotangent $\lambda_i$ is the gradient with respect to the complete KV output of writer $i$, across all layers. Within one writer, an early-layer page may also influence a later-layer page. That intra-writer dependency belongs inside the local Jacobian and is not a separate temporal edge. The accumulated \texttt{.grad} of a raw intermediate tensor can include such internal use and is not the external $\lambda_i$.

\paragraph{\rev{Blockwise form.}}
\rev{In blocks, the recursion of Eq.~\eqref{eq:adjoint} reads}
\begin{equation}
\mathsf A_{ri}=\begin{cases}
D_{\KV_i}\KV_r,&i<r,\\
0,&i\ge r,
\end{cases}\qquad
\eta_i=\sum_{r=i+1}^{n}(D_{\KV_i}\tilde x_r^{t_s})^\transpose v_r,
\label{eq:causaljacobian}
\end{equation}
\rev{and} the stacked cotangents $\boldsymbol\lambda=(\lambda_0,\ldots,\lambda_n)$ and $\boldsymbol\eta=(\eta_0,\ldots,\eta_n)$ satisfy
\begin{equation}
\boldsymbol\lambda
=(I-\mathsf A^\transpose)^{-1}\boldsymbol\eta
=\underbrace{\boldsymbol\eta}_{\rev{\text{direct reads}}}
+\underbrace{\sum_{q=1}^{n}(\mathsf A^\transpose)^q\boldsymbol\eta}_{\text{via later writes}}.
\label{eq:creditfactorization}
\end{equation}
The expansion is finite, and reverse mode realizes it through vector--Jacobian products without forming $\mathsf A$. \rev{Detaching the dependence of a write on earlier writes} removes $\mathsf A$ and keeps only \rev{the gradient of the direct reads}.

\paragraph{\rev{Temporal gradient routes.}}
Expanding the writer differential gives
\begin{equation}
\diff\KV_i=D_\theta\KV_i\,\diff\theta+
D_\phi\KV_i\,\diff\phi+
\sum_{j<i}D_{\KV_j}\KV_i\,\diff\KV_j.
\end{equation}
The first two terms describe how parameters directly change a write with its current history fixed. The last term propagates a change in earlier memory into later memory. Combining this writer differential with Eq.~\eqref{eq:differential} gives two temporal routes from a past write to the loss: \rev{paths through a direct read}
\[
\KV_j\longrightarrow\tilde x_i^{t_s}\longrightarrow L_\DMD
\]
and recurrent-write paths \rev{through $q\ge1$ later writes, the term $(\mathsf A^\transpose)^q\boldsymbol\eta$ of Eq.~\eqref{eq:creditfactorization},}
\[
\KV_j\to\rev{\KV_{r_1}\to\cdots\to\KV_{r_q}}\to\tilde x_i^{t_s}\to L_\DMD,
\]
\rev{with $j<r_1<\cdots<r_q<i$.}
The sampled context $\bar x_r^\ctx$ is detached (\rev{Eq.~\eqref{eq:stop}}), so neither route passes through \rev{the prediction that an intermediate writer takes as its input: temporal gradients travel through the written KV alone}.

Since the loss has no direct dependence on the written KV, detaching the historical pages from every generation call removes all gradient to writer-only parameters, even when the dependencies between successive writes are retained (Eqs.~\eqref{eq:adjoint} and~\eqref{eq:phi}).

\paragraph{\rev{Role-specific gradients.}}
At fixed $\theta$ and a history given as an independent tensor,
\begin{equation}
D_\phi G_\theta(\bar z_i,t_s,\KV_{<i})=0.
\end{equation}
\rev{Under the stop-gradient of Eq.~\eqref{eq:stop}, which holds the exit input $\bar z_i$ fixed,} the total derivative through a writer-produced history is
\begin{equation}
\frac{\diff\tilde x_i^{t_s}}{\diff\phi}
=\sum_{j<i}D_{\KV_j}\tilde x_i^{t_s}\frac{\diff\KV_j}{\diff\phi}.
\end{equation}
Generation therefore depends on the residual only through the historical KV it produces.

\subsection{Proof of Proposition~\ref{prop:exact}}
\begingroup
\begin{restatedproposition}[Exact single-traversal differentiation]
Fix the recorded inputs, model parameters, visibility, positions and random choices across forward execution and backward replay. Assume that each role computation is a pure function of its explicit inputs, state extension preserves the historical prefix and its differential (Eqs.~\eqref{eq:prefixinvariant} and~\eqref{eq:extensionjacobian}), and replay satisfies Eq.~\eqref{eq:rematerializationidentity}. Then the traversal (Algorithm~\ref{alg:kvforcing}) computes the same predictions and parameter gradients as the uncheckpointed recurrence of Sec.~\ref{sec:memoryobjective}, in exact arithmetic.
\end{restatedproposition}
\begin{lemma}[Page reads preserve the gradient]
\label{lem:pageread}
In one layer, let $\KV_0,\ldots,\KV_{i-1}$ be the writer outputs, $\overline{\KV}_j=\sg(\KV_j)$ their committed pages and $h_j$ their handles, whose forward value is zero and whose backward passes the incoming gradient to $\KV_j$ unchanged. Let $u$ be the K/V of the reading segment. The page read returns $y=[\overline{\KV}_0,\ldots,\overline{\KV}_{i-1},u]$ and, for an incoming gradient $(g_0,\ldots,g_{i-1},g_u)$ split at the page boundaries, sends $g_j$ to $h_j$ and $g_u$ to $u$. Then the page read has the value and the vector--Jacobian product of the differentiable concatenation $[\KV_0,\ldots,\KV_{i-1},u]$.
\end{lemma}
\begin{proof}
A committed page is copied from $\KV_j$ once and never modified, so $\overline{\KV}_j$ equals $\KV_j$ in value and the two concatenations agree. The vector--Jacobian product of the concatenation returns $g_j$ to $\KV_j$ and $g_u$ to $u$. The page read hands $g_j$ to $h_j$, whose backward passes it to $\KV_j$ unchanged, and $g_u$ to $u$, so every input receives the same gradient. A page read by several segments accumulates their contributions by summation in both cases.
\end{proof}
The custom backward rule of the handle is not the derivative of its constant forward function. Combined with the page read, it recovers the first-order vector--Jacobian product of the differentiable concatenation.
\endgroup
\begin{proof}
\rev{Fix the recorded inputs $\bar z_i$ and $\bar x_i^\ctx$, $i=0,\ldots,n$, of one update, so that $L_\DMD(\tilde X)$ of Sec.~\ref{sec:memoryobjective}, with Eqs.~\eqref{eq:generate} and~\eqref{eq:context} evaluated on them, is a function of $(\theta,\phi)$. We show that Algorithm~\ref{alg:kvforcing} computes the same $\nabla_{\theta,\phi}L_\DMD$ as the uncheckpointed recurrence of Sec.~\ref{sec:memoryobjective}}: the recorded latents are constants, and every dependency through the historical writes $\KV_j$ is retained.

At the segment level the differentiable computation consists of the exit prediction $\tilde x_i^{t_s}$ of Eq.~\eqref{eq:generate} and the write $\KV_i$ of Eq.~\eqref{eq:context}, and both depend on earlier segments only through their writes $\KV_j$, $j<i$. The segment computations therefore form a directed acyclic graph ordered by time, with the terminal write $\KV_n$ a formal node that Algorithm~\ref{alg:kvforcing} omits.

Both executions start from an empty history. Suppose their writes agree through segment $i-1$. Segment $i$ then receives the same recorded inputs, historical KV, visibility and positions in both executions, so its exit prediction and its write agree. By induction all segment outputs are identical. \rev{By Lemma~\ref{lem:pageread}, the page reads of Algorithm~\ref{alg:kvforcing} have the value and the vector--Jacobian product of reading the writer outputs directly, so the traversal differentiates the same graph as the recurrence.}

Differentiating the state extension of \rev{Eq.~\eqref{eq:stateextension}} gives Eq.~\eqref{eq:extensionjacobian}. Its identity block preserves the derivatives of earlier writes, and its lower block carries the dependence of the current write on the parameters and on the historical KV. Reversing these dependencies accumulates gradient from later generation and writer nodes, giving Eq.~\eqref{eq:adjoint} with the finite causal expansion of Appendix~\ref{app:causalfactorization}. Summing the local parameter contributions gives Eqs.~\eqref{eq:theta}--\eqref{eq:phi}.

Committed pages preserve the historical prefix seen by each segment (Eq.~\eqref{eq:prefixinvariant}). In backward, checkpoint replay uses the same prefix, group entry state, parameters and randomness as the original forward (Eq.~\eqref{eq:rematerializationidentity}), so it reconstructs the same local computation and the same local Jacobian. Replacing the retained internal activations by this reconstruction leaves every vector--Jacobian product unchanged, and the state extension runs outside the reconstructed map, so replay cannot alter the history. Reverse-mode differentiation therefore produces the same parameter gradients as the uncheckpointed recurrence, in exact arithmetic.
\end{proof}

\rev{The proposition holds in exact arithmetic. Floating-point results can differ through numerical precision, accumulation order and nondeterministic kernels.} Appendix~\ref{app:implementation} reports the numerical agreement of the implementation. \rev{The gradient is the first-order gradient of the recurrence of Sec.~\ref{sec:memoryobjective} with its stop-gradients (Eq.~\eqref{eq:stop}).}

\subsection{\rev{Execution and memory cost}}
\label{app:pageexecution}
\paragraph{Execution and storage.}
A single traversal is one autoregressive forward sweep with no second whole-sequence reconstruction (Algorithm~\ref{alg:kvforcing}). Each nonterminal segment computes a \rev{graph-carrying} exit prediction and a memory write, the final write is omitted, and the other sampling steps run without gradient. \rev{Algorithm~\ref{alg:kvforcing} omits the sampling steps after the exit step: the implementation runs them to keep the rollout schedule, and they enter neither the loss nor the write.} In backward, internal activations are rematerialized in groups of five transformer blocks. Let $B$ be the number of transformer blocks and $b$ the group size, and let $S$ be the all-layer KV storage of one segment, $\rev{N}$ the entry state kept per group, $Q$ the retained prediction and $\rev{M_b}$ the transient attention workspace per segment of history when one group of $b$ blocks is recomputed and back-propagated, live for one group at a time. The retained tensor memory for a clip of $T$ segments is approximately
\begin{equation}
O\!\left(TS+2T\left\lceil B/b\right\rceil \rev{N}+TQ+T\rev{M_b}\right),
\label{eq:memory}
\end{equation}
excluding parameters, optimizer state, score models and metadata. The four terms are the pages ($TS$, including their cotangents up to a constant), the entry states of the exit and write forwards of every segment ($2T\lceil B/b\rceil \rev{N}$), the retained predictions ($TQ$) and the workspace of the one group under recomputation ($T\rev{M_b}$). Every term is linear in $T$, the group size $b$ trades the entry states against the workspace, and attention over the full history keeps its quadratic cost (Figure~\ref{fig:components}).

\paragraph{Retained state of each scheme.}
\label{sec:pagecontract}
We compare four executions; Figure~\ref{fig:pagecontract} draws the two cache-based alternatives and dKV Forcing. Each rollout \rev{segment} uses five forwards: four denoising forwards, including \rev{the sampled exit step, whose prediction enters the loss,} and one write forward. The schemes differ in which forwards carry a graph, how historical K/V enter it and when activations are released. \emph{Self Forcing} keeps only the exit-forward graph, with the write detached. \emph{Differentiating the mutable cache directly} runs out of memory under checkpoint replay (Sec.~\ref{sec:pages}), and without checkpointing before \rev{the sixth segment}. \emph{The two-pass reconstruction} records exit inputs and clean contexts during a no-grad rollout, then recomputes context and exit forwards in one parallel teacher-forced pass over all $2T$ \rev{segments}. Historical K/V are ordinary outputs, with no in-place writes. It retains the whole second-pass forward's checkpoint boundaries and attention working set, replays it layer by layer in backward, and keeps the $T$-\rev{segment} cache allocated between steps. \emph{dKV Forcing} allocates the page table once and keeps the exit and write graphs within the rollout. The write input is detached, and new K/V are committed to immutable pages outside replay, \rev{and zero-valued handles keep the gradient connection.} Across \rev{segments}, page data, group entry states and the exit predictions persist, and internal activations are rematerialized in groups of five blocks. In backward each page receives its own contiguous gradient buffer \rev{(Appendix~\ref{app:code}), and the page table is released after backward}. The default execution retains the $(T-1)$ committed pages and their cotangents, and the hidden-state variants retain \rev{the per-layer inputs of the K and V projections} instead (Table~\ref{tab:perf}). \rev{Table~\ref{tab:perf} reports the costs and Figure~\ref{fig:components} the memory composition of one generator update.}

\begingroup
\paragraph{Activation storage.}
Backpropagation through the historical KV requires the local Jacobians of the earlier writes, not only their values. Without recomputation, the activations of every graph-carrying forward are kept until backward, a storage that grows with the number of such forwards. Enabling the memory gradient adds the writes to this storage, and checkpointing removes it for them as it does for the exit forwards. The failure of the mutable cache arises in the replay (Appendix~\ref{app:replay}).
\endgroup

\section{Implementation and experimental details}
\label{app:protocol}
\subsection{Experimental setup}
\label{app:setupdetails}
\begingroup
\paragraph{Training data.}
Training uses only text prompts and no video data: the $248{,}221$ filtered and extended VidProM prompts~\citep{wang2024vidprom} of Self Forcing~\citep{huang2025self}. None of the $326$ VBench-Long evaluation prompts appears among them verbatim.

\paragraph{Training configuration.}
Frame-wise runs start from the released frame-wise initialization of CF++, chunk-wise runs from its chunk-wise initialization, and both train for 1500 steps at 4 GPUs $\times$ batch 2. Training uses four denoising steps at timesteps $1000/750/500/250$ with shift $5$, on $16$ latent channels at $60\times104$. Each generator update follows five critic updates. AdamW uses betas $(0,0.999)$ and weight decay $0.01$, the backbone learning rate is $2\times10^{-6}$, the writer-residual learning rate $10^{-4}$ and the EMA decay $0.99$. Training uses bf16 mixed precision under FSDP for the generator and the score networks.

\paragraph{Inference.}
Inference keeps the first $4$ frames, a sliding window of $16$ and the current frame in the frame-wise setting, and a $3+15+3$ layout in the chunk-wise setting, with the writer residual active only at cache refresh.

\paragraph{Evaluation.}
VBench at 5\,s~\citep{huang2024vbench} evaluates 944 prompts with 5 videos each plus 75 flicker prompts with 25 videos each, and reports the Total, Quality and Semantic scores. VBench-Long at 60\,s uses 326 prompts and $241$ latent frames ($240$ chunk-wise), at 240\,s $963$ latent frames, and reports subject and background consistency, motion smoothness, dynamic degree, aesthetic and imaging quality, and flicker on the static-filtered subset.

\paragraph{Training seeds.}\label{app:registry}
Every configuration is trained with seeds 1, 10 and 20, and the main tables report these three seeds so that all rows and horizons are matched. Some configurations have additional seeds, which Tables~\ref{tab:complete-fw60}--\ref{tab:complete-240} include (Appendix~\ref{app:complete}). Repeated runs of a seed are averaged before the mean $\pm$ SD over training seeds is taken.
\endgroup

\subsection{Implementation}
\label{app:pageimpl}
\paragraph{\rev{Page storage.}}
Each layer stores the post-RoPE K and V of the committed \rev{segments} in one contiguous \texttt{\_PageTable} of shape \rev{$[\text{batch},(T-1)L,\text{heads},\text{head dim}]$, with $L$ tokens per segment}, allocated on first use in the attention dtype, \rev{and a slot is never overwritten once it is committed}. The stored values stay outside the autograd graph, and a \texttt{\_PageHandle}, an expanded scalar placeholder, keeps the gradient connection to each writer. A \texttt{PageRead} snapshot records the table, the visible length and the handles, and later appends leave its prefix unchanged. \rev{Forwards without gradient} use the next uncommitted slot as scratch. Differentiable reads concatenate the historical values with the current KV through \texttt{\_PageCat}, which returns one contiguous gradient per page, \rev{so that a retained page gradient does not keep the gradient of the whole concatenation in memory} (code in Appendix~\ref{app:code}).

\paragraph{\rev{Checkpointing.}}
Non-reentrant checkpoint groups hold five blocks, with the patch embedding in the first group and the prediction head in the last, \rev{and the RNG state is not saved because the blocks draw no random numbers.} Differentiable cross-attention recomputes the text KV inside replay, and \rev{sampling without gradient} shares a cached text KV, outside any checkpointed region, \rev{until the rollout ends.} The writer omits its output head and the final block's unused attention output, \rev{and the write of the last segment is omitted because no segment reads it.}

\paragraph{Execution variants.} The default execution reuses the \rev{graph-carrying} exit prediction as the detached context, which removes a duplicate exit evaluation. \rev{The fused variant runs the exit forward and the write as one batch, with separate attention masks and the residual applied to the write only, and obtains} the detached write input through \rev{an exit evaluation without gradient}. \rev{The hidden-state variant keeps, as the page that carries the graph, the input $h$ of a layer's K and V projections in the write, which has half as many values as K and V together, and so does its gradient. A forward with gradient rebuilds the historical K and V from these inputs with the writer's projections inside its own checkpoint region, so they exist only while that region is computed or recomputed, and the gradient is that of the same function. A detached KV table serves the reads without gradient.} \rev{The two can be combined.} \rev{All four compute the gradients of the same objective.}

\subsection{Validation}
\label{app:implementation}
\begingroup
We validate the implementation against the two-pass reconstruction with identical parameters, recorded inputs and sampling randomness, and compare the predictions and the parameter gradients of all four executions.

\paragraph{Gradient equivalence.}
All errors are relative $\ell_2$ errors $\|a-b\|_2/\|b\|_2$ against the two-pass reference. In fp64, on the 30-layer generator at a reduced latent size with dense attention, with and without the residual and at $T=8$ and $21$, the outputs differ from the reference by at most $3.0\times10^{-15}$ and the gradients over all parameters by at most $2.8\times10^{-14}$ ($3.2\times10^{-14}$ with hidden-state pages). The gradients of the residual parameters alone differ by at most $5.7\times10^{-13}$. Over all fp64 tests, including the fused variant and those under FSDP, the relative gradient error of every execution stays below $4.1\times10^{-12}$. At $T=42$ and $T=63$ the gradients agree to at most $2.1\times10^{-14}$. Three-frame chunks agree with the reference to $3\times10^{-15}$ in the outputs and $2.6\times10^{-14}$ in the gradients. In the configuration used for training (full-size model, $T=21$, bf16, FSDP) with deterministic attention, the default execution matches the two-pass reference bitwise in outputs and gradients, and 42-step training runs with either give the same final weights. The fused and hidden-state variants match the reference in their outputs.

\paragraph{Training and inference paths.}
With identical checkpoint, prompt and noise, we compare in fp64 the training rollout with the last exit step against inference, with the plain cache and with the sliding window, on every frame of a full-size 21-frame clip that both caches cover. The three rollouts are bitwise identical.
\endgroup

\subsection{Page execution in PyTorch}
\label{app:code}
The following PyTorch excerpts implement the execution of Figure~\ref{fig:memorylearning}. They keep the operations and identifiers of the training code and omit dtype handling, the execution variants and the distributed bookkeeping. The implementation runs through ordinary module calls under FSDP with \rev{two} small autograd functions and no custom kernel.

The pages exist only in the training rollout. Inference runs the ordinary cache path listed at the end of this appendix, where the same writer residual is active during the cache refresh and denoising uses the shared backbone without it. A page stores one segment's post-RoPE K and V at one layer, written once and connected to its writer through a \rev{zero-valued} handle. When the inference cache covers the full training clip, the training and inference paths attend to the same historical K/V.

\paragraph{\rev{Values and gradient.}}
\rev{The following lines correspond to Figure~\ref{fig:pagecontract}(b) and~(c) and to the two points of Sec.~\ref{sec:pages}.} In the mutable cache one tensor stores the value and carries the graph. In dKV Forcing the page table stores the value and a handle carries the gradient.
\par\noindent\begin{minipage}{\linewidth}
\begin{pycode}
# mutable cache, gradient open: one tensor, both jobs
kv_cache["k"][:, lo:hi] = k     # value AND graph, in place
k_attn = kv_cache["k"][:, :hi]  # both pass through it

# dKV Forcing: the two jobs are split
with torch.no_grad():           # (1) the value
    table.put(slot, k, v)       # detached, written once
handle = _PageHandle.apply(k)   # (2) the gradient:
                                # a zero node wired to k
k_attn = _PageCat.apply(        # read: values from the
    table.k[:, :n * L], k_own,  # table, gradient to the
    *handles)                   # handles
\end{pycode}
\end{minipage}\par

\paragraph{Write-once slots and the commit.}
Each layer allocates one contiguous page table on first use, at its full size. The page of segment $i$ occupies rows $[iL,(i+1)L)$, and the commit writes that slot once, under \texttt{no\_grad} and outside any checkpoint region.
\begin{pycode}
class _PageTable:
    def put(self, slot, k, v):
        L = k.shape[1]
        if self.k is None:  # allocated once, appended to
            shape = (k.shape[0], self.max_frames * L,
                     *k.shape[2:])
            self.k = k.new_empty(shape)
            self.v = v.new_empty(shape)
        rows = slice(slot * L, (slot + 1) * L)
        self.k[:, rows].copy_(k)
        self.v[:, rows].copy_(v)
        return self.k[:, rows], self.v[:, rows]

class PersistentKVStore:
    def append_all_layers(self, pages):
        # pages = (k0, v0, k1, v1, ...), the writer output
        for l in range(self.num_layers):
            k, v = pages[2 * l], pages[2 * l + 1]
            kh = self.k_handles[l]
            vh = self.v_handles[l]
            with torch.no_grad():
                # the commit: outside any graph
                self.tables[l].put(self.length, k, v)
            if k.requires_grad:
                # the page's place in the graph
                kh.append(_PageHandle.apply(k))
                vh.append(_PageHandle.apply(v))
        self.length += 1

    def snapshot_all_layers(self):
        # a later append cannot change a snapshot
        return tuple(
            PageRead(self.tables[l], self.length,
                     tuple(self.k_handles[l]),
                     tuple(self.v_handles[l]))
            for l in range(self.num_layers))
\end{pycode}
A committed slot is never written again, so a checkpointed forward recomputed in backward reads the history it saw in forward and has no write to re-execute.

\paragraph{Handles.}
\rev{The K/V values are stored in the page table, and the handles keep their autograd connections.} \texttt{\_PageHandle} takes the writer's output and returns a zero scalar expanded to the page's shape, \rev{so the graph keeps the dependency on the writer's output without a second K/V-sized tensor.} Its backward passes the incoming gradient unchanged to the \texttt{k}, \texttt{v} tensors the writer returned, and from there into the writer's own graph. A snapshot is a tuple of the table, its visible length and the handles of the pages that carry a graph, and later appends leave it unchanged.
\begin{pycode}
class _PageHandle(torch.autograd.Function):
    @staticmethod
    def forward(ctx, page):  # the page's shape, no data
        return page.new_zeros(()).expand(page.shape)

    @staticmethod
    def backward(ctx, grad):  # the gradient, unchanged
        return grad

class PageRead(NamedTuple):  # one layer's history
    table: _PageTable
    length: int
    k_handles: tuple
    v_handles: tuple
\end{pycode}

\paragraph{Reading through handles.}
A forward with gradient attends to the history slots plus its own K/V. \texttt{\_PageCat} concatenates a view of the visible slots with the segment's own K/V into a tensor \rev{that exists only inside the rematerialized region.} Its backward splits the history's gradient page by page and returns each slice through the corresponding handle. Every returned slice, the segment's own included, is cloned into its own contiguous buffer, so a retained page gradient does not retain the gradient of the whole concatenation. A forward without gradient, every sampling step and every forward of the critic step, writes its own K/V into the next free slot as scratch and attends to a view of the table, \rev{so the attention receives the same values in both cases.} \rev{The last segment, which writes no page and has no slot, concatenates as a forward with gradient does.}
\begin{pycode}
class _PageCat(torch.autograd.Function):
    @staticmethod
    def forward(ctx, history, own, *handles):
        ctx.sizes = [h.shape[1] for h in handles]
        ctx.split = history.shape[1]
        return torch.cat([history, own], dim=1)

    @staticmethod
    def backward(ctx, grad):
        def owned(g):  # one contiguous buffer per page
            fmt = torch.contiguous_format
            return g.clone(memory_format=fmt)
        pages = grad[:, :ctx.split].split(ctx.sizes, 1)
        return (None, owned(grad[:, ctx.split:]),
                *(owned(g) for g in pages))

# the self-attention of one block
k, v = self.k(x), self.v(x)
if self.memory_lora and write_page:  # phi: write only
    k = k + self.k_memory_lora(x)
    v = v + self.v_memory_lora(x)
k = rope(self.norm_k(k))
table, n = history.table, history.length
if not torch.is_grad_enabled() and n < table.max_frames:
    # sampling: in place
    k, v = table.put(n, k, v)  # scratch slot
    k_attn = table.k[:, :(n + 1) * L]
    v_attn = table.v[:, :(n + 1) * L]
else:  # with graph, or last segment: read via handles
    k_attn = _PageCat.apply(table.k[:, :n * L], k,
                            *history.k_handles)
    v_attn = _PageCat.apply(table.v[:, :n * L], v,
                            *history.v_handles)
y = self.o(flash_attention(q, k_attn, v_attn))
return (y, k, v) if write_page else y
\end{pycode}

\paragraph{The writer residual ($\phi$).}
The \texttt{write\_page} flag adds the writer residual to the K and V projections of every block. The exit forward and all sampling steps use the backbone projections alone. The up-projection is zero-initialized, so at step 0 the written KV equals the pretrained one. \rev{The residual is a standard LoRA of rank 64 on each of the two projections.}

\paragraph{Checkpoint groups.}
Both graph-carrying forwards run their blocks in non-reentrant checkpoint regions of $g$ blocks ($g=5$ in our runs, six regions over 30 blocks). \rev{Each region receives only the histories of its own blocks,} so a snapshot's handles enter the graph where they are read. Only the region inputs are kept between forward and backward, and the pages, being region outputs, are always kept. \rev{The blocks draw no random numbers, so no RNG state is saved.}
\begin{pycode}
def _persistent_blocks(self, x, histories, write_page):
    n = len(self.blocks)
    g = self.persistent_checkpoint_group
    pages = []
    for first in range(0, n, g):
        last = min(first + g, n)
        x, *page = torch.utils.checkpoint.checkpoint(
            self._run_blocks, first, last, x,
            histories[first:last], write_page,
            use_reentrant=False, preserve_rng_state=False)
        pages += page
    return x, pages
\end{pycode}
\texttt{\_run\_blocks} runs blocks \texttt{first}\,..\,\texttt{last}$-1$ against their histories and returns the residual stream plus, for a write, those blocks' pages; the first region also runs the patch embedding and the last region of an exit forward runs the output head, so neither the embedded frame nor the head input is retained.

\paragraph{The two forwards.}
\rev{The write forward omits the prediction head, and its last block computes only its page, since its output is not used.}
\begin{pycode}
def forward_noisy_frame(self, latent, histories):  # G
    x, _ = self._persistent_blocks(
        latent, histories, write_page=False)
    return self.unpatchify(x)  # head output, reshaped

def forward_clean_writer(self, latent, histories):  # K
    _, pages = self._persistent_blocks(
        latent, histories, write_page=True)
    return tuple(pages)  # (k0, v0, k1, v1, ...), no head
\end{pycode}

\paragraph{The traversal.}
The rollout follows the sampling of inference. Per segment, two forwards carry a graph: the exit forward $G_\theta$, whose prediction enters the loss, and the write $\Kwriter_{\theta,\phi}$, whose output becomes the segment's pages. Every other sampling step runs without gradient.
\begin{pycode}
store = PersistentKVStore(num_layers, max_frames=T - 1)
outputs = []
for i in range(T):
    hist = store.snapshot_all_layers()  # the prefix
    x_t = noise[:, i:i + 1]
    for s, t in enumerate(steps):  # sampling steps
        if s == exit_step:  # exit forward, with graph
            out = noisy(x_t, t, hist)
            outputs.append(out)
            x0 = x0_exit = out.detach()  # stop-gradient
        else:
            with torch.no_grad():
                x0 = noisy(x_t, t, hist)
        if s < len(steps) - 1:
            x_t = add_noise(x0, steps[s + 1])
    with torch.no_grad():  # the write input
        x_ctx = add_noise(x0_exit, context_t)
    if i < T - 1:  # the last segment writes no page
        pages = write(x_ctx, context_t, hist)
        store.append_all_layers(pages)  # commit once
store.end_rollout()
loss = video_loss(torch.cat(outputs, dim=1))
\end{pycode}
\rev{Detaching the exit prediction implements the stop-gradient of Eq.~\eqref{eq:stop}, so the gradient reaches the writer only through its pages.} \texttt{end\_rollout} drops only the prompt's cross-attention K/V that the forwards without gradient shared, and the pages stay, referenced by the graph, until the backward pass is done.

\paragraph{Backward.}
\rev{No explicit backward schedule is required:} autograd carries the gradient of each read through \texttt{\_PageCat} and the handles to the \texttt{k}, \texttt{v} the writer returned and into the writer's own regions, which are recomputed in turn, and the same route continues from a writer to the earlier pages it read. \rev{Replay does not write, since the committed slots are read-only, and} the tables are released \rev{when the backward pass has finished.}

\paragraph{The mutable cache.}
The ordinary path keeps one K/V buffer per layer and writes each segment into it in place, inside the block's checkpoint region:
\begin{pycode}
# the ordinary path: a block inside its checkpoint region
for i, block in enumerate(self.blocks):
    x = checkpoint(block, x, kv_cache=kv_cache[i], ...)

# inside the block's attention
k, v = self.k(x), self.v(x)
kv_cache["k"][:, lo:hi] = k  # in place, with graph
kv_cache["v"][:, lo:hi] = v
k_attn = kv_cache["k"][:, :hi]
v_attn = kv_cache["v"][:, :hi]
y = self.o(flash_attention(q, k_attn, v_attn))
\end{pycode}
With gradient enabled, the assignment makes the buffer a node of the segment's graph, and every later reader saves that buffer for its backward. When a region is recomputed in backward, the assignment runs again on the same tensor and attaches a second graph to it, while the readers' saved buffers and the recomputed graphs accumulate. On 5\,s clips the first backward pass runs out of memory (Sec.~\ref{sec:pagecontract}). Self Forcing and Causal Forcing++ avoid this by \rev{detaching the cache} after every segment, \rev{which removes the gradient that dKV Forcing restores:}
\begin{pycode}
# Self Forcing and CF++: detach the cache after each segment
for entry in kv_cache:
    entry["k"] = entry["k"].detach()
    entry["v"] = entry["v"].detach()
\end{pycode}

\paragraph{The two-pass reconstruction.}
\rev{SGF keeps the detached mutable cache and recovers the gradient with a second forward.} Pass 1 is the rollout without gradient: it builds the cache and records each segment's noisy input at the exit step and its write input. Pass 2 is one teacher-forced forward over the whole clip with gradient, in which the recorded write inputs provide the context KV and the recorded noisy inputs the predictions:
\begin{pycode}
# pass 1: the rollout without gradient builds the cache
with torch.no_grad():
    for i in range(T):
        for s, t in enumerate(steps):
            if s == exit_step:
                noisy_at_t[:, i] = x_t  # kept for pass 2
            x0 = model(x_t, t, kv_cache)
            ...  # sampling as before
        x_ctx = add_noise(x0_exit, context_t)
        x_ctx_hat[:, i] = x_ctx  # kept for pass 2
        model(x_ctx, context_t, kv_cache)  # the write
# pass 2: one teacher-forced forward over the whole clip
output = model(noisy_at_t, t_exit,
               clean_x=x_ctx_hat.detach(), aug_t=context_t)
loss = video_loss(output)
\end{pycode}
\rev{Pass 2 recomputes every exit forward and every write from the recorded inputs, which adds a second forward over the whole clip.} The gradient equals the single-traversal one (Proposition~\ref{prop:exact}), at the cost of the second pass and of holding the whole-sequence graph (Table~\ref{tab:perf}).

\subsection{Performance measurements}
\label{app:performancearchive}
\paragraph{\rev{Measurement protocol.}}
\rev{We time the full DMD trainer and synchronize CUDA on every rank before each timestamp. A training cycle consists of five critic updates and one generator update and includes text encoding, rollout, scoring, backward, FSDP communication and optimizer updates. We report medians over steady cycles, which exclude startup, the final step and the cycles affected by periodic cache clearing. Phase medians are separate statistics, so their sum approximates the median cycle. Peaks are the maximum allocated memory over ranks and include resident tensors.}

\paragraph{Memory retention under checkpoint replay.}\label{app:replay}
The examined mutable implementation writes projected KV into an external buffer inside checkpointed forward. Non-reentrant backward replay repeats those writes and attaches new graph connections (including \texttt{CopySlices}) to externally retained tensors. \rev{The replayed graphs are then kept after their backward is done.} Disabling checkpointing removes replay but retains all forward activations, producing a different OOM mechanism.

\rev{The generator-only run below uses eight H100 GPUs, 30 layers, 21 frames, 1560 tokens per frame and bf16 FSDP, without teacher, critic and optimizer.}

\paragraph{\rev{Source of the memory growth.}}
Self Gradient Forcing attributes the cost of a differentiable cache to the autograd graph that every cache write must retain until its readers are processed, a graph that grows with the rollout~\citep{zhuang2026sgf}. \rev{We test whether the memory overhead comes from differentiating through the historical KV or from the cache writes that checkpoint replay repeats. In the generator-only run, the mutable cache ends the forward at $15.36$\,GiB, reaches $57.22$\,GiB during backward and keeps $42.29$\,GiB allocated after backward has finished. The memory that remains after backward is not released with the graph of the update, so it is held by references from outside that graph. With the same gradient through the same rollout, pages peak at $28.09$\,GiB, and in the setting of Table~\ref{tab:perf} the full memory gradient raises the generator-update peak over detached-write training by $6.2$\,GiB ($62.1$ against $55.9$\,GiB). The mutable cache also trains once its replay writes are isolated: with the writer input detached, a private cache for every replayed block and the cache references released after each backward, it completes every update at $73.1$\,GiB with no growth between updates (Table~\ref{tab:perf}). The excess memory of the mutable cache therefore comes from replayed graphs that stay referenced by the external cache, which pages avoid.}

\paragraph{\rev{Memory breakdown of one generator update}.}\label{app:components}
\begin{figure*}[t]
\centering
\includegraphics[width=\textwidth]{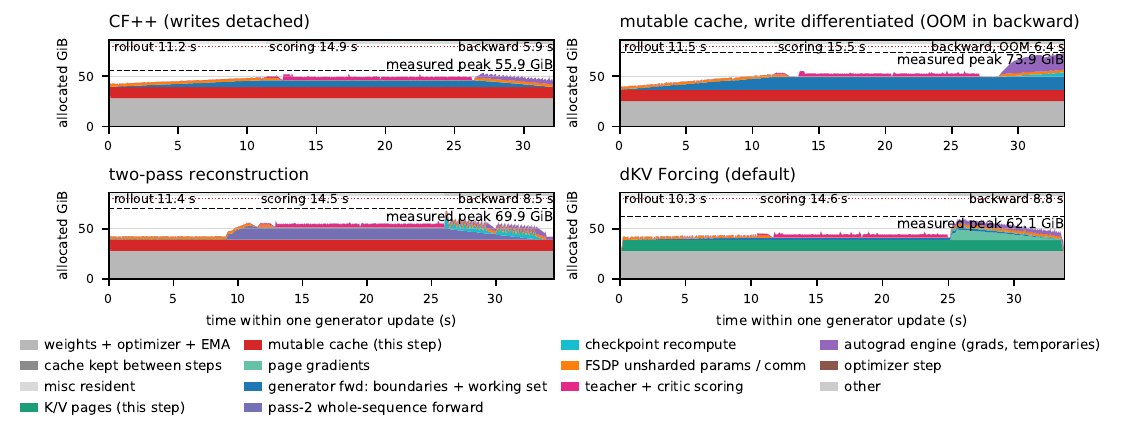}
\caption{\textbf{\rev{GPU memory breakdown of one generator update.}} Measured component accounting for the four training schemes. CF++ (detached writes) keeps only exit-forward boundaries and its backward is short. The \rev{mutable cache with the write differentiated} keeps both forwards' boundaries and the sampled trajectory, and its backward exhausts memory while replay attaches new graph to the external cache (purple: engine-allocated gradients and temporaries). The two-pass reconstruction keeps no graph in the rollout but retains the whole second-pass forward (violet) and replays it layer by layer in backward (cyan), with the cache resident between steps (red). dKV Forcing allocates the page table once (green), keeps only group entry states during the rollout, and in backward adds one contiguous gradient per page (light green) and the local replay of five-block groups.}
\label{fig:components}
\end{figure*}
Figure~\ref{fig:components} gives the measured memory composition of one generator update on rank 0 (4 H100 $\times$ batch 2, bf16, FSDP, 21 frames). \rev{The allocator trace is classified by allocation site, the resident part covers the weights, optimizer states and EMA, and for CF++ and the two-pass reconstruction the cache kept between steps, and the phase durations are the steady-state medians of Table~\ref{tab:perf}. The mutable cache with the write differentiated ends at its out-of-memory failure in the first backward. Dashed lines are the measured peaks.}

\paragraph{Execution variants.}
Table~\ref{tab:perf} gives the training cycles of the execution variants at 4 GPUs $\times$ batch 2. At 8 GPUs $\times$ batch 1, Table~\ref{tab:variants} summarizes them and Table~\ref{tab:cycles8x1} gives the complete statistics, with and without the residual. Table~\ref{tab:variants} summarizes the execution variants at $8$ GPUs $\times$ batch 1 by cycle time and generator peak, and Table~\ref{tab:cycles8x1} breaks that measurement down, with and without the residual. At $8\times1$, the default execution takes 66.9 s per cycle against 78.2 s for the two-pass reconstruction. \rev{The $59.0$\,GiB first-step peak of the two-pass reconstruction comes from the first construction of its attention mask.} \rev{The default execution reuses the exit evaluation, stores the history in contiguous memory and omits the terminal write.} \rev{The fused variant runs the exit forward and the write as one batch.} At $8\times1$, the rollout plus gradient-forward phase falls from 11.09 to 6.91 s and backward from 7.56 to 4.72 s, and the generator update from 25.69 to 19.20 s. Critic time remains approximately 8 s. \rev{The values are unchanged, and batching the two forwards reduces the time. Keeping the activations of both raises the allocated peak slightly} (Table~\ref{tab:cycles8x1}). Hidden-state pages retain attention inputs $h$ instead of both K and V and reconstruct the latter when needed. Relative to the default execution, the generator peak falls (Table~\ref{tab:cycles8x1}) while the generator update rises from 25.69 to 29.19 s and backward from 7.56 to 9.53 s, with the critic time unchanged. Fused+hidden combines this storage reduction with batching, with a generator update of 22.09 s. \rev{Only the forwards with gradient change, and the rollout without gradient is the same.}
\begin{table}[t]
\caption{\textbf{Execution variants of the single traversal} at $8$ GPUs $\times$ batch 1 (no residual, bf16): a full training cycle of five critic updates and one generator update; times are steady-state medians, peaks the maximum allocated memory across ranks during a generator update. The variants are defined in Appendix~\ref{app:pageimpl}, and all rematerialize in groups of five blocks. Bold and underline: best and second best per column.}
\label{tab:variants}
\centering\small
\resizebox{0.8\linewidth}{!}{\begin{tabular}{@{}lrr@{}}
\toprule
method & cycle (s) & peak (GiB)\\
\midrule
Two-pass reconstruction & 78.2 & 36.1\\
default & 66.9 & 36.1\\
fused & $\mathbf{59.5}$ & 36.6\\
hidden & 69.4 & $\mathbf{31.8}$\\
fused+hidden & $\underline{62.6}$ & $\underline{33.3}$\\
\bottomrule
\end{tabular}}
\end{table}
\begin{table*}[t]
\caption{\textbf{Training cycles at 8 GPUs $\times$ batch 1} (bf16). Steady-cycle medians $\pm$ half the min--max range over six cycles. Memory in GiB (no-LoRA maximum over the eight ranks, resident memory at critic-rollout entry). Bold and underline: best and second best per column.}
\label{tab:cycles8x1}
\centering\footnotesize\setlength{\tabcolsep}{2.5pt}\renewcommand{\arraystretch}{0.95}
\begin{tabular}{@{}lcccccccccc@{}}
\toprule
Method & Cycle, no LoRA (s) & vs.\ two-pass & Cycle, LoRA (s) & vs.\ two-pass & Gen.\ peak & Critic peak & Resident & Step-0 peak & All-step peak & Reserved\\
\midrule
Two-pass & 78.2 $\pm$ 1.1 & +0.0\% & 81.0 $\pm$ 1.0 & +0.0\% & 36.1 & 29.3 & 20.4 & 59.0 & 59.0 & 60.8\\
default & 66.9 $\pm$ 0.9 & $-$14.5\% & 67.9 $\pm$ 2.8 & $-$16.2\% & 36.1 & 23.5 & 14.7 & 20.2 & 36.1 & 58.3\\
fused & $\mathbf{59.5}$ $\pm$ 1.2 & $-$23.9\% & $\mathbf{61.8}$ $\pm$ 0.5 & $-$23.7\% & 36.6 & 23.5 & 14.7 & 20.2 & 36.6 & 60.6\\
hidden & 69.4 $\pm$ 1.9 & $-$11.2\% & 75.2 $\pm$ 1.0 & $-$7.2\% & $\mathbf{31.8}$ & 23.6 & 14.7 & 20.2 & $\mathbf{31.8}$ & $\mathbf{56.2}$\\
fused+hidden & $\underline{62.6}$ $\pm$ 0.6 & $-$20.0\% & $\underline{65.7}$ $\pm$ 0.7 & $-$19.0\% & $\underline{33.3}$ & 23.6 & 14.7 & 20.2 & $\underline{33.3}$ & $\underline{57.9}$\\
\bottomrule
\end{tabular}
\end{table*}

\paragraph{\rev{Checkpoint group size.}}
Generator-only measurements of the checkpoint group size (weights excluded) give 20.5\,GiB and 25.3\,s for groups of one block, 17.3\,GiB and 15.9\,s for groups of five and 25.3\,GiB and 13.7\,s for groups of thirty, which sets the default of five.

\paragraph{Clip-length scaling.}
Figure~\ref{fig:tscale} uses the full trainer (teacher, critic, both optimizers, bf16 FSDP) on 4 H100 at batch 1 per GPU, with the clip length fixed to $T$ latent frames, whole-clip attention in every method, and the teacher and critic scoring the last $21$ frames as in the public long-clip configurations. Each point is a single run with steady-state generator updates. The two-pass reconstruction runs its public configuration with per-block checkpointing, and its attention mask is built row by row, since the dense mask of the public builder cannot be allocated at $T=42$. Table~\ref{tab:tscale} lists the measurements. At batch 2 per GPU every method fits $T=21$ and none fits $T=42$.
\begin{table*}[!t]
\caption{\textbf{Only dKV Forcing carries the memory gradient to 63 frames.} Clip-length scaling at batch 1 per GPU: generator-update peak (GiB) / generator-update time (s) / resident memory (GiB). OOM: out of memory. $^\dagger$First update completes at $73.5$\,GiB, the second runs out of memory.}
\label{tab:tscale}
\centering\small
\begin{tabular}{@{}lcccc@{}}
\toprule
method & $T=21$ & $T=42$ & $T=63$ & $T=84$\\
\midrule
CF++ (detached writes) & 47.2 / 20.6 / 35.0 & 56.3 / 38.8 / 40.8 & 65.2 / 58.5 / 46.4 & OOM\\
two-pass reconstruction & 50.1 / 20.8 / 35.0 & 69.7 / 40.5 / 40.8 & OOM & OOM\\
mutable cache, repaired & 55.8 / 27.6 / 35.0 & 73.5 / 58.5 / 40.8 & OOM & OOM\\
dKV Forcing (default) & 50.1 / 23.0 / 29.2 & 62.7 / 48.5 / 29.5 & OOM$^\dagger$ & OOM\\
dKV Forcing (hidden-state pages) & 45.7 / 26.2 / 29.3 & 56.6 / 69.0 / 29.5 & 67.2 / 134.2 / 29.6 & OOM\\
\bottomrule
\end{tabular}
\end{table*}

\setcounter{dbltopnumber}{4}\setcounter{totalnumber}{6}\renewcommand{\dbltopfraction}{0.96}\renewcommand{\textfraction}{0.03}\renewcommand{\dblfloatpagefraction}{0.9}\setcounter{topnumber}{3}\setcounter{bottomnumber}{2}\renewcommand{\topfraction}{0.9}\renewcommand{\bottomfraction}{0.8}\renewcommand{\floatpagefraction}{0.9}
\FloatBarrier
\section{Additional experiments and results}
\label{app:extended}
\subsection{\rev{Gradient conflict and dedicated writer}}
\label{app:controls}
\paragraph{\rev{Gradient conflict.}}
\rev{We measure the generator gradient of dKV Forcing without rollout dropout at training step 1400, on two batches with fixed noise, over all generator parameters: the direct gradient is the gradient with the writes detached, and the memory gradient is the remainder. Repeated measurements differ by 0.5--1\%. The norm ratio of the memory gradient to the direct gradient is 1.04--1.06, and their cosine is $-0.08$ to $-0.14$, about $-0.22$ after Adam normalization, in which each gradient is divided by $\sqrt{\hat v}+\epsilon$ with the second-moment estimate $\hat v$ of the optimizer. The writer residual receives no direct gradient.}

\paragraph{The dedicated writer.}
The dedicated-writer variant introduces a second 1.3B model initialized from \rev{the same initialization}. With parameters $\psi$, replace Eq.~\eqref{eq:context} by $\KV_i=\Kwriter_\psi(\bar x_i^\ctx,\tau,\KV_{<i})$. The generator receives the \rev{direct gradient}, and $\psi$ receives the memory gradient. \rev{The writer has its own optimizer group and EMA.} \rev{It costs two generator-sized parameter sets and a full-network write forward.}

\paragraph{\rev{Numerical verification.}}
\rev{To check that the dedicated writer separates the two gradients, we compare its gradients with those of the shared model.} At identical initialization, \rev{the generator gradient differs from that of the shared model with detached writes} by 0.435\% \rev{(run-to-run variation 0.434\%)}, the writer gradient differs from \rev{the memory gradient of the shared model} by 0.97\%, and the summed gradient from the full gradient by 0.66\%.

\paragraph{\rev{Results.}}
\rev{The} dedicated writer of Table~\ref{tab:roles} reaches $64.9\pm0.6$ aesthetic and $71.6\pm0.7$ imaging points over three seeds with the highest mean subject consistency of the table ($98.6\pm0.4$). \rev{Aesthetic quality and subject consistency rise from SGF to the writer residual to the dedicated writer, as more parameters are used by the write alone.} \rev{The dedicated writer keeps this gain at 240\,s (Table~\ref{tab:complete-240}). With the K and V projections of SGF trained at the learning rate of the residual, $10^{-4}$, the dynamic degree is $4.0\pm3.2$ at 60\,s and $5.2\pm4.4$ at 240\,s over three seeds.}

\begin{table*}[!t]
\caption{\textbf{Generation and memory KV differ in representation and in future attention readout.} We compare the written K/V with the frame's own generation K/V (same frame, same layer; layer means over 30 layers, 48 videos $\times$ 21 frames, ranges over the checkpoints). The attention mass and the attention output are measured with only frame $i$'s cache entry replaced and every later query held fixed.}
\label{tab:repr}
\centering\small
\begin{tabular}{@{}lccc@{}}
\toprule
& \multicolumn{2}{c}{$\KV^\mem$ vs.\ $\KV^\gen$} & read by later frames\\
\cmidrule(lr){2-3}\cmidrule(lr){4-4}
& K & V & attention output\\
\midrule
relative difference $\|A-B\|/\|B\|$ & 0.35--0.37 & 0.70--0.72 & 0.65--0.68\\
token-wise cosine & 0.93 & 0.73 & 0.76--0.77\\
per-head linear CKA & 0.89--0.91 & 0.72--0.74 & --\\
\midrule
\rev{attention mass, written / generation} & \multicolumn{3}{c}{$1.17$--$1.40\times$}\\
\bottomrule
\end{tabular}
\end{table*}
\begin{table*}[!t]
\caption{\textbf{Replacing written KV changes the autoregressive continuation.} With checkpoint, prompt and sampling randomness fixed, the written K/V of frames 0--2 or of all 21 frames is replaced by their generation K/V at 5\,s (48 prompts per checkpoint). Next-frame change: latent relative $L_2$ to the unmodified rollout, with an independent seed in parentheses as the reference scale. $\Delta$: paired change of the prefix$\to$later DINO consistency (mean $\pm$ s.e.m.).}
\label{tab:intervention}
\centering\small
\begin{tabular}{@{}lccc@{}}
\toprule
checkpoint & next-frame change (other seed) & $\Delta$ consistency, frames 0--2 & $\Delta$ consistency, all frames\\
\midrule
\rev{CF++ initialization} & 0.55 (0.97) & $-0.231\pm0.024$ & $-0.597\pm0.018$\\
SGF & 0.57 (1.08) & $-0.059\pm0.010$ & $-0.183\pm0.017$\\
\rev{dKV Forcing without rollout dropout} & 0.56 (1.07) & $-0.070\pm0.009$ & $-0.166\pm0.014$\\
\bottomrule
\end{tabular}
\end{table*}
\begin{figure*}[!t]
\centering
\includegraphics[width=0.92\textwidth]{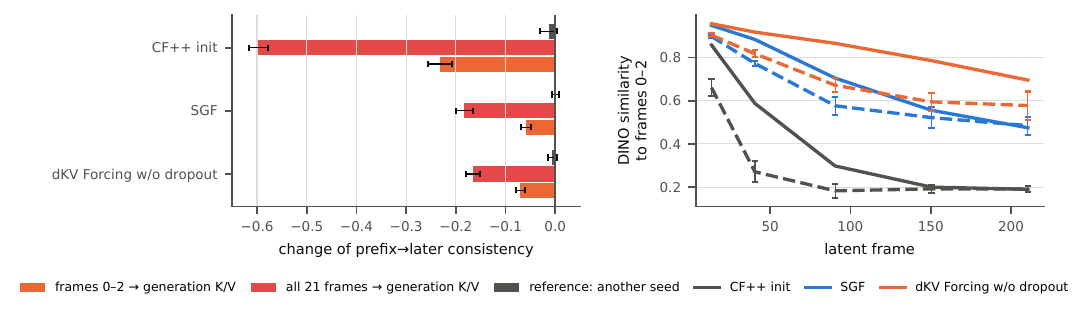}
\caption{\textbf{Replacing the written KV by the generation KV \rev{reduces temporal consistency.}} Left: paired change of the prefix$\to$later DINO consistency at 5\,s when the memory of frames 0--2, or of all 21 frames, is replaced by generation K/V, with an independent seed as the reference scale (mean $\pm$ s.e.m. over 48 prompts). Right: 60\,s rollouts (16 prompts), similarity to the opening by latent-frame window, unchanged (solid) versus memory of frames 0--2 replaced (dashed).}
\label{fig:intervention}
\end{figure*}
\subsection{\rev{Analysis of generation and memory representations}}
\label{app:repr}
\begingroup
\paragraph{Representation differences.}
Table~\ref{tab:repr} compares the written KV with the generation KV of the same frame and layer, by the relative difference $\|A-B\|/\|B\|$, the token-wise cosine and the per-head linear CKA. The differences are already present at the CF++ initialization, before our fine-tuning, and change little after $1500$ steps of training without the residual (the relative difference grows by $0.012\pm0.003$ for K and $0.019\pm0.004$ for V, paired by prompt). A linear probe separates the two representations with $100\%$ held-out accuracy at every layer.

\paragraph{Effects on attention readout.}
Later frames read the two differently. With their queries held fixed, they give the written KV $1.17$--$1.40\times$ the attention mass of the generation KV, up to $2.5\times$ in layer 18, and the attention output changes by $0.65$--$0.68$ in relative terms (Table~\ref{tab:repr}).

\paragraph{Effects on autoregressive continuation.}
\endgroup
The cache intervention of Sec.~\ref{sec:representation} replaces the memory of the first three frames by their generation K/V, with checkpoint, prompt, initial noise, re-noising noise and denoising schedule held fixed (Table~\ref{tab:intervention}, Figure~\ref{fig:intervention}). The consistency drop is present in $44$--$46$ of $48$ prompts for frames 0--2 and in $47$--$48$ of $48$ when all frames are replaced, and it grows with \rev{the number of frames whose memory is replaced}. A second seed used as the reference changes consistency by $0.00\pm0.01$. In 60\,s rollouts \rev{the drop in DINO consistency} persists through latent frames $61$--$120$ ($15$--$30$\,s) in all three checkpoints tested and through frames $121$--$180$ for \rev{dKV Forcing without rollout dropout}. The same three-frame intervention at frames $11$--$13$ \rev{changes the later latents less at every matched lag and gives no measurable drop in consistency}. Switching off only the residual at the write ($\Delta W=0$) \rev{changes the next frame less ($0.14$--$0.26$) and} lowers consistency in the models with the residual when applied to all frames.

\subsection{\rev{Motion--fidelity trade-off}}
\rev{In the frame-wise setting the memory gradient raises aesthetic quality and lowers motion (Table~\ref{tab:complete-fw60}): from CF++ to SGF, aesthetic quality rises from $51.6$ to $57.2$ while dynamic degree falls from $75$ to $39$, and the writer residual raises aesthetic quality further, to $62.1$, at a dynamic degree of $22.6$. Rollout dropout disables the writer residual for an entire rollout with probability $p$, in both the critic and the generator step. We use $p=0.5$, which gives equal probability to rollouts with and without the residual. It recovers the dynamic degree of CF++ ($74.0$) at an aesthetic quality of $60.1$. Chunk-wise, the residual does not lower motion (Table~\ref{tab:complete-cw60}) and is used without dropout.}

\FloatBarrier
\begin{table*}[t]
\captionof{table}{\rev{The reported frame-wise configurations}, VBench-Long standard prompts at 60\,s, step 1500, mean $\pm$ SD over $n$ training seeds (same-seed repeats averaged first).}
\label{tab:complete-fw60}
\centering\small
\begin{tabular}{@{}lccccc@{}}
\toprule
configuration & $n$ & aesthetic & dynamic & imaging & subject\\
\midrule
CF++ (writes detached) & 4 & $51.6\pm2.4$ & $74.9\pm15.8$ & $62.7\pm1.8$ & $94.8\pm1.1$\\
SGF & 5 & $57.2\pm2.6$ & $38.7\pm16.3$ & $68.0\pm1.8$ & $96.7\pm0.8$\\
dedicated writer & 3 & $64.9\pm0.6$ & $26.0\pm18.2$ & $71.6\pm0.7$ & $98.6\pm0.4$\\
\rev{dKV Forcing without rollout dropout} & 4 & $62.1\pm1.4$ & $22.6\pm14.4$ & $69.6\pm0.6$ & $97.9\pm0.6$\\
\rev{dKV Forcing (ours)} & 4 & $60.1\pm1.2$ & $74.0\pm28.7$ & $69.0\pm1.1$ & $96.6\pm1.0$\\
\bottomrule
\end{tabular}

\par\vspace{0.2em}
\captionof{table}{\rev{The reported chunk-wise configurations} (three latent frames per segment, cache $3+15+3$), VBench-Long standard prompts at 60\,s, step 1500, mean $\pm$ SD over $n$ seeds.}
\label{tab:complete-cw60}
\centering\small
\resizebox{\linewidth}{!}{\begin{tabular}{@{}lcccccccc@{}}
\toprule
configuration & $n$ & aesthetic & dynamic & imaging & subject & background & smoothness & flicker\\
\midrule
CF++ (writes detached) & 3 & $57.8\pm1.6$ & $50.6\pm15.8$ & $65.6\pm1.2$ & $96.8\pm0.7$ & $95.7\pm0.7$ & $98.3\pm0.5$ & $98.8\pm0.7$\\
SGF & 3 & $58.1\pm0.1$ & $48.4\pm19.8$ & $67.8\pm1.4$ & $96.9\pm0.5$ & $95.7\pm0.6$ & $98.6\pm0.5$ & $98.8\pm0.3$\\
\rev{dKV Forcing (ours)} & 3 & $59.6\pm0.8$ & $61.9\pm19.7$ & $68.7\pm1.7$ & $96.8\pm0.8$ & $95.8\pm0.1$ & $98.6\pm0.4$ & $99.0\pm0.3$\\
\bottomrule
\end{tabular}}

\par\vspace{0.2em}
\captionof{table}{\rev{The configurations evaluated} at 240\,s (VBench-Long standard prompts, $963$ latent frames), step 1500, mean $\pm$ SD over all available training seeds.}
\label{tab:complete-240}
\centering\small
\resizebox{\linewidth}{!}{\begin{tabular}{@{}lcccccccc@{}}
\toprule
configuration & $n$ & aesthetic & dynamic & imaging & subject & background & smoothness & flicker\\
\midrule
CF++ (writes detached) & 4 & $47.7\pm3.2$ & $77.8\pm14.5$ & $60.8\pm3.0$ & $93.9\pm0.8$ & $94.4\pm0.6$ & $97.8\pm1.3$ & $98.4\pm0.3$\\
SGF & 5 & $51.9\pm2.3$ & $38.8\pm15.0$ & $65.0\pm2.6$ & $95.9\pm0.8$ & $95.5\pm0.6$ & $98.6\pm0.3$ & $98.2\pm0.7$\\
\rev{dedicated writer} & \rev{3} & \rev{$63.4\pm0.7$} & \rev{$25.8\pm19.6$} & \rev{$71.4\pm0.6$} & \rev{$98.2\pm0.4$} & \rev{$96.6\pm0.5$} & \rev{$99.0\pm0.2$} & \rev{$98.9\pm0.3$}\\
\rev{dKV Forcing without rollout dropout} & 4 & $58.9\pm2.2$ & $22.1\pm16.0$ & $67.6\pm1.1$ & $97.3\pm0.8$ & $96.3\pm0.4$ & $99.0\pm0.2$ & $99.1\pm0.3$\\
\rev{dKV Forcing (ours)} & 4 & $57.4\pm0.3$ & $74.3\pm28.0$ & $68.2\pm0.4$ & $96.2\pm0.8$ & $94.9\pm0.6$ & $97.8\pm0.8$ & $98.6\pm0.1$ ($n{=}3$)\\
chunk-wise CF++ & 3 & $52.5\pm4.3$ & $50.4\pm16.1$ & $62.3\pm2.9$ & $95.9\pm0.5$ & $95.3\pm0.4$ & $98.0\pm0.7$ & $98.7\pm0.6$\\
chunk-wise SGF & 3 & $53.2\pm1.0$ & $46.8\pm20.4$ & $65.4\pm2.7$ & $96.4\pm0.5$ & $95.4\pm0.5$ & $98.6\pm0.5$ & $98.7\pm0.3$\\
\rev{chunk-wise dKV Forcing (ours)} & 3 & $55.7\pm2.2$ & $57.1\pm24.3$ & $66.2\pm2.3$ & $96.6\pm0.6$ & $95.4\pm0.1$ & $98.8\pm0.3$ & $98.9\pm0.2$\\
\bottomrule
\end{tabular}}

\end{table*}
\begin{table}[!t]
\caption{\textbf{VBench at 5\,s, the training horizon.} Official Total, Quality and Semantic scores and dynamic degree, mean $\pm$ SD over training seeds $\{1,10,20\}$. Per block, best value per column in bold, second best underlined.}
\label{tab:vbench5s}
\centering\footnotesize\setlength{\tabcolsep}{3pt}
\resizebox{\linewidth}{!}{\begin{tabular}{@{}lcccc@{}}
\toprule
configuration & Total & Quality & Semantic & dynamic\\
\midrule
\multicolumn{5}{@{}l}{\emph{frame-wise}}\\
CF++ & $\mathbf{83.5}\pm0.7$ & $\mathbf{84.8}\pm0.5$ & $78.1\pm1.5$ & $\mathbf{84.0}\pm12.9$\\
SGF & $\underline{83.1}\pm0.4$ & $\underline{84.1}\pm0.3$ & $\underline{79.4}\pm1.1$ & $59.4\pm7.6$\\
\rev{dKV Forcing without rollout dropout} & $\underline{83.1}\pm1.0$ & $83.8\pm1.2$ & $\mathbf{80.4}\pm0.2$ & $36.7\pm19.6$\\
\rev{dKV Forcing (ours)} & $82.9\pm0.1$ & $83.8\pm0.2$ & $79.3\pm0.9$ & $\underline{76.2}\pm28.1$\\
\midrule
\multicolumn{5}{@{}l}{\emph{chunk-wise}}\\
CF++ & $83.8\pm0.1$ & $84.8\pm0.2$ & $\mathbf{79.9}\pm0.5$ & $69.1\pm10.8$\\
SGF & $\underline{83.9}\pm0.2$ & $\underline{85.0}\pm0.2$ & $\underline{79.5}\pm0.4$ & $\underline{71.3}\pm11.6$\\
\rev{dKV Forcing (ours)} & $\mathbf{84.2}\pm0.5$ & $\mathbf{85.4}\pm0.5$ & $79.1\pm0.8$ & $\mathbf{80.7}\pm4.8$\\
\bottomrule
\end{tabular}}
\end{table}
\begin{figure}[!t]
\centering
\includegraphics[width=\linewidth]{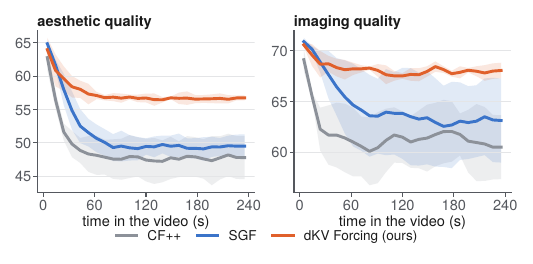}
\caption{\textbf{Quality over time in 240\,s videos.} Per-clip VBench-Long aesthetic and imaging quality, mean $\pm$ SD over training seeds. \rev{The methods start at similar quality, and dKV Forcing degrades the least.}}
\label{fig:decayfull}
\end{figure}
\subsection{\rev{Additional quantitative results}}
\label{app:complete}
\begingroup
\paragraph{Short horizon.}
Table~\ref{tab:vbench5s} lists the VBench scores at 5\,s, the training horizon, where the overall scores of the configurations are similar.

\paragraph{Long horizons.}
Tables~\ref{tab:complete-fw60}--\ref{tab:complete-240} list the reported configurations at the standard training budget: frame-wise at 60\,s, chunk-wise at 60\,s, and every 240\,s evaluation. Conventions are those of Sec.~\ref{sec:exp-setup}: mean $\pm$ SD over training seeds, same-seed repeats averaged first, flicker on the static-filtered subset (missing where no clip passed the filter). The main text reports the three matched seeds, and these tables include every available seed, so the same configuration can have slightly different values in the two. The ranking is unchanged: with all seeds the quality score is $81.6\pm1.6$ for dKV Forcing against $78.4\pm1.0$ for SGF and $78.5\pm0.8$ for CF++ at 60\,s, and $80.9\pm1.8$ against $76.6\pm1.1$ and $77.4\pm1.3$ at 240\,s.

\begin{figure*}[t]
\centering
\captionsetup{skip=2pt,font=footnotesize}
\scalebox{0.95}{
\begingroup
\setlength{\tabcolsep}{0.75pt}\renewcommand{\arraystretch}{1.0}
\sffamily\footnotesize
\begin{tabular}{@{}>{\raggedleft\arraybackslash}m{0.105\textwidth}@{\hspace{4pt}}*{6}{>{\centering\arraybackslash}m{0.12\textwidth}}@{}}
 & 0\,s & 48\,s & 96\,s & 144\,s & 192\,s & 238\,s\\[1pt]
CF++ & \multicolumn{6}{@{\hspace{4pt}}>{\centering\arraybackslash}m{\dimexpr0.72\textwidth+7.5pt\relax}@{}}{\includegraphics[width=\linewidth]{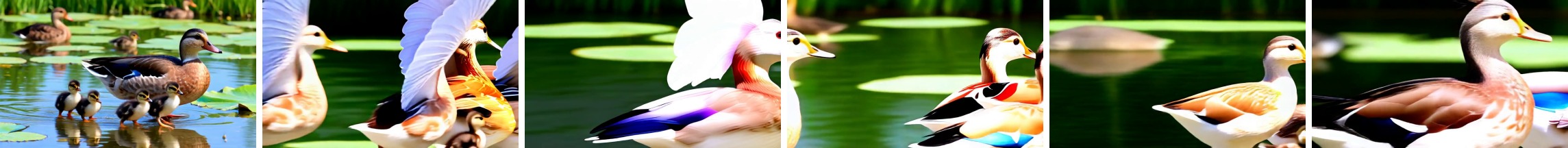}}\\
SGF & \multicolumn{6}{@{\hspace{4pt}}>{\centering\arraybackslash}m{\dimexpr0.72\textwidth+7.5pt\relax}@{}}{\includegraphics[width=\linewidth]{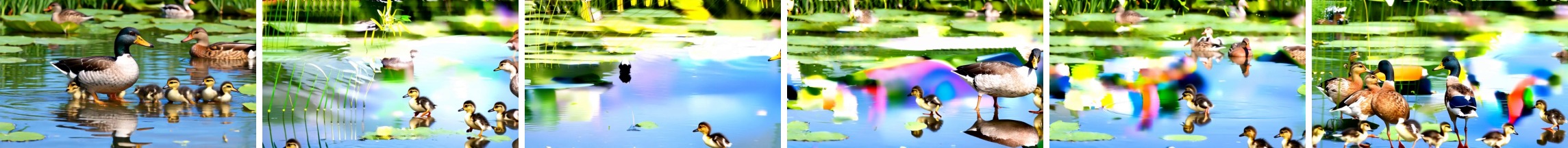}}\\
\rev{dKV Forcing w/o dropout} & \multicolumn{6}{@{\hspace{4pt}}>{\centering\arraybackslash}m{\dimexpr0.72\textwidth+7.5pt\relax}@{}}{\includegraphics[width=\linewidth]{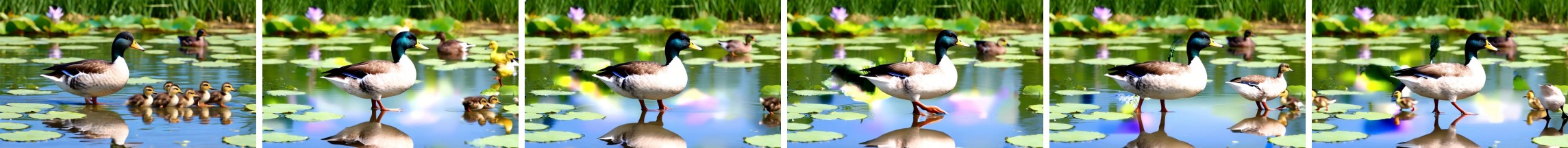}}\\
\textbf{dKV Forcing (ours)} & \multicolumn{6}{@{\hspace{4pt}}>{\centering\arraybackslash}m{\dimexpr0.72\textwidth+7.5pt\relax}@{}}{\includegraphics[width=\linewidth]{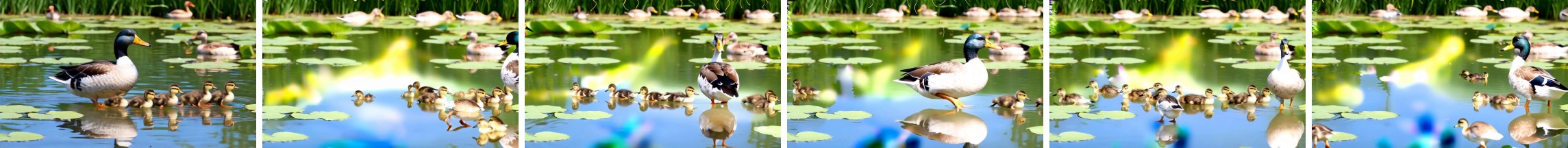}}\\
\end{tabular}
\endgroup
}
\caption{``Mother duck and six ducklings''. Prompt: \emph{A serene watercolor painting of a mother duck leading her six ducklings across a tranquil pond. The mother duck has a gentle expression, her feathers glistening in the sunlight, and she frequently glances back to ensure all her ducklings are safely following in a neat little line. The ducklings follow closely behind, their small heads bobbing up and down as they waddle along. The background features a peaceful pond with lily pads and ducks floating nearby, creating a harmonious and natural scene. A mid-shot from a slightly elevated angle captures the mother duck and her ducklings in motion.}}
\label{fig:qual1}
\vspace{-3pt}
\scalebox{0.95}{
\begingroup
\setlength{\tabcolsep}{0.75pt}\renewcommand{\arraystretch}{1.0}
\sffamily\footnotesize
\begin{tabular}{@{}>{\raggedleft\arraybackslash}m{0.105\textwidth}@{\hspace{4pt}}*{6}{>{\centering\arraybackslash}m{0.12\textwidth}}@{}}
 & 0\,s & 48\,s & 96\,s & 144\,s & 192\,s & 238\,s\\[1pt]
CF++ & \multicolumn{6}{@{\hspace{4pt}}>{\centering\arraybackslash}m{\dimexpr0.72\textwidth+7.5pt\relax}@{}}{\includegraphics[width=\linewidth]{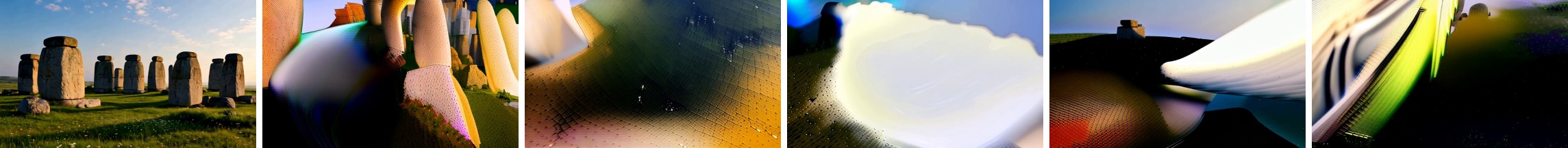}}\\
SGF & \multicolumn{6}{@{\hspace{4pt}}>{\centering\arraybackslash}m{\dimexpr0.72\textwidth+7.5pt\relax}@{}}{\includegraphics[width=\linewidth]{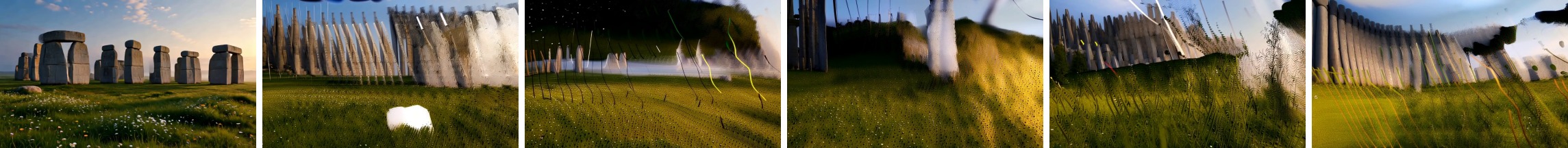}}\\
\rev{dKV Forcing w/o dropout} & \multicolumn{6}{@{\hspace{4pt}}>{\centering\arraybackslash}m{\dimexpr0.72\textwidth+7.5pt\relax}@{}}{\includegraphics[width=\linewidth]{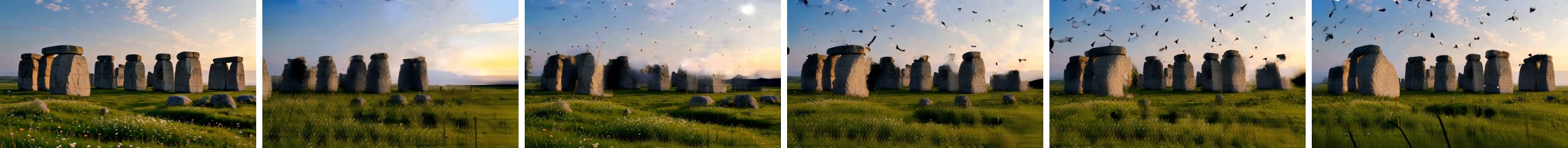}}\\
\textbf{dKV Forcing (ours)} & \multicolumn{6}{@{\hspace{4pt}}>{\centering\arraybackslash}m{\dimexpr0.72\textwidth+7.5pt\relax}@{}}{\includegraphics[width=\linewidth]{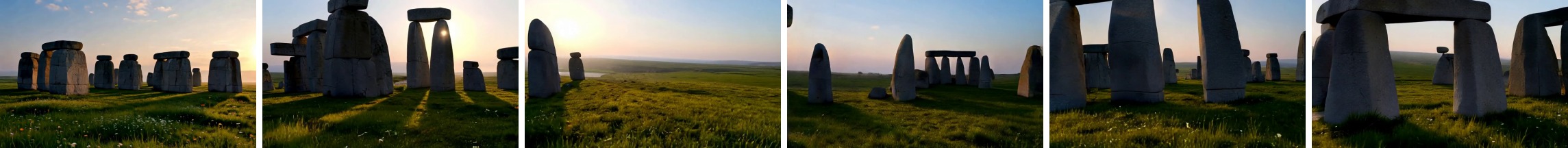}}\\
\end{tabular}
\endgroup
}
\caption{``Tranquil tableau of Stonehenge''. Prompt: \emph{A tranquil tableau of the Stonehenge presented itself as an enigmatic puzzle, each colossal stone meticulously placed against the backdrop of tranquility}}
\label{fig:qual2}
\end{figure*}

\paragraph{Paired comparisons.}
Table~\ref{tab:matched} compares dKV Forcing with CF++ and SGF on videos of matched motion: aesthetic quality, imaging quality, subject consistency and text alignment improve at both horizons ($77$--$113$ pairs against CF++, $44$--$107$ against SGF). Without matching the motion, the paired differences over all prompts are $+2.5$ $[2.0,3.1]$ points of aesthetic quality against SGF at 60\,s and $+5.1$ $[4.4,5.8]$ at 240\,s.

\paragraph{Quality over time.}
Figure~\ref{fig:decayfull} shows the per-clip aesthetic and imaging quality over the 240\,s videos. The methods start at similar quality, and dKV Forcing degrades the least.
\endgroup

\subsection{Additional qualitative comparisons}
\label{app:qualitative}
\begin{figure*}[p]
\centering
\captionsetup{skip=2pt,font=footnotesize}
\scalebox{0.95}{
\begingroup
\setlength{\tabcolsep}{0.75pt}\renewcommand{\arraystretch}{1.0}
\sffamily\footnotesize
\begin{tabular}{@{}>{\raggedleft\arraybackslash}m{0.105\textwidth}@{\hspace{4pt}}*{6}{>{\centering\arraybackslash}m{0.12\textwidth}}@{}}
 & 0\,s & 48\,s & 96\,s & 144\,s & 192\,s & 238\,s\\[1pt]
CF++ & \multicolumn{6}{@{\hspace{4pt}}>{\centering\arraybackslash}m{\dimexpr0.72\textwidth+7.5pt\relax}@{}}{\includegraphics[width=\linewidth]{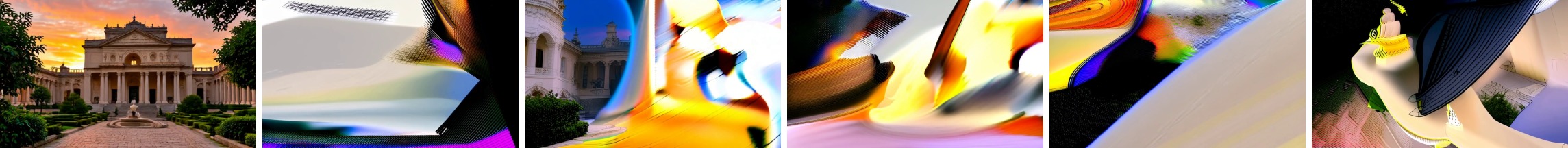}}\\
SGF & \multicolumn{6}{@{\hspace{4pt}}>{\centering\arraybackslash}m{\dimexpr0.72\textwidth+7.5pt\relax}@{}}{\includegraphics[width=\linewidth]{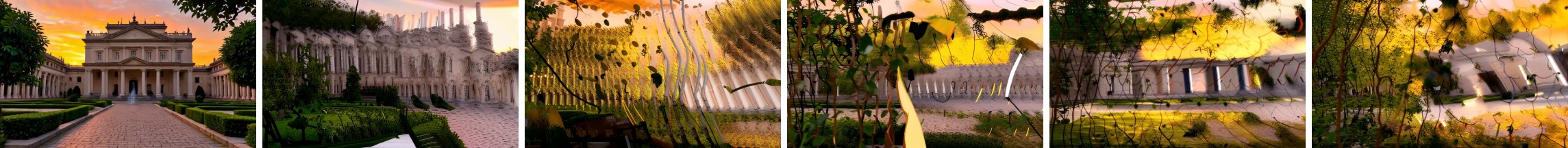}}\\
\rev{dKV Forcing w/o dropout} & \multicolumn{6}{@{\hspace{4pt}}>{\centering\arraybackslash}m{\dimexpr0.72\textwidth+7.5pt\relax}@{}}{\includegraphics[width=\linewidth]{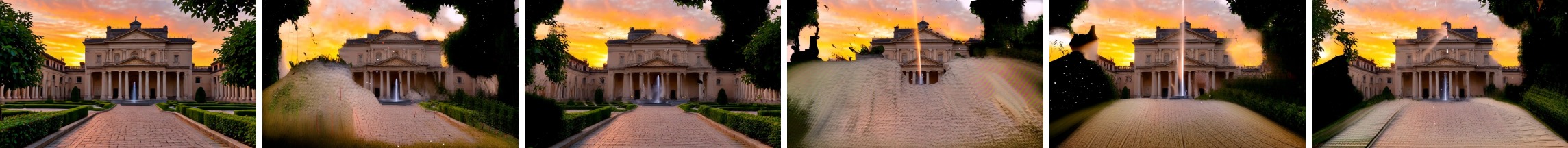}}\\
\textbf{dKV Forcing (ours)} & \multicolumn{6}{@{\hspace{4pt}}>{\centering\arraybackslash}m{\dimexpr0.72\textwidth+7.5pt\relax}@{}}{\includegraphics[width=\linewidth]{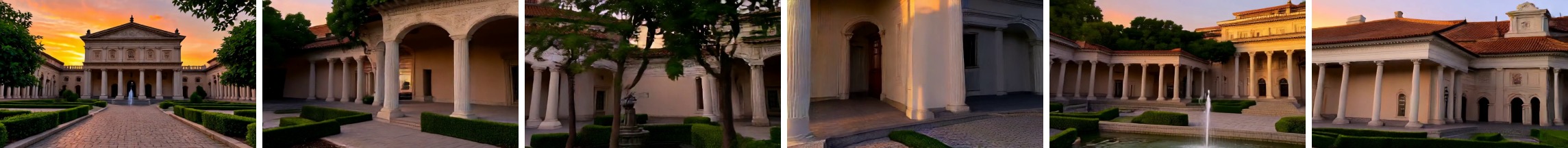}}\\
\end{tabular}
\endgroup
}
\caption{``Tranquil tableau of a palace''. Prompt: \emph{A tranquil tableau of palace}}
\label{fig:qual3}
\vspace{-3pt}
\scalebox{0.95}{
\begingroup
\setlength{\tabcolsep}{0.75pt}\renewcommand{\arraystretch}{1.0}
\sffamily\footnotesize
\begin{tabular}{@{}>{\raggedleft\arraybackslash}m{0.105\textwidth}@{\hspace{4pt}}*{6}{>{\centering\arraybackslash}m{0.12\textwidth}}@{}}
 & 0\,s & 48\,s & 96\,s & 144\,s & 192\,s & 238\,s\\[1pt]
CF++ & \multicolumn{6}{@{\hspace{4pt}}>{\centering\arraybackslash}m{\dimexpr0.72\textwidth+7.5pt\relax}@{}}{\includegraphics[width=\linewidth]{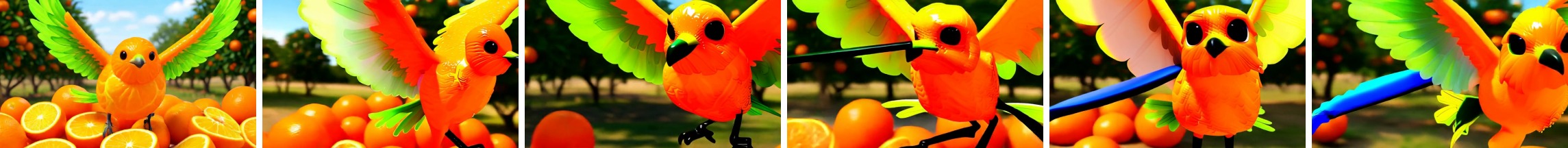}}\\
SGF & \multicolumn{6}{@{\hspace{4pt}}>{\centering\arraybackslash}m{\dimexpr0.72\textwidth+7.5pt\relax}@{}}{\includegraphics[width=\linewidth]{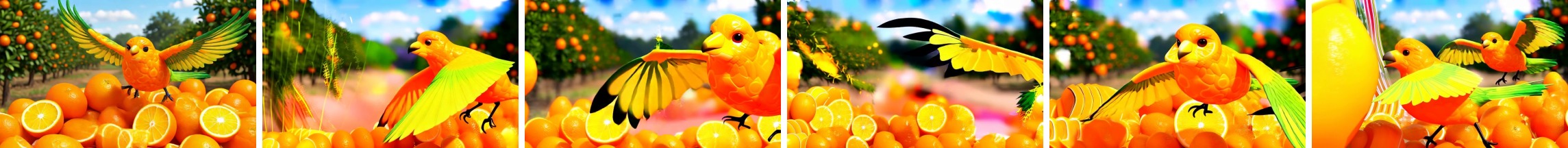}}\\
\rev{dKV Forcing w/o dropout} & \multicolumn{6}{@{\hspace{4pt}}>{\centering\arraybackslash}m{\dimexpr0.72\textwidth+7.5pt\relax}@{}}{\includegraphics[width=\linewidth]{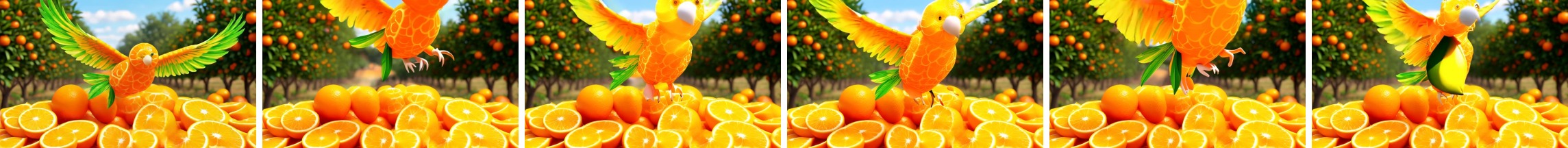}}\\
\textbf{dKV Forcing (ours)} & \multicolumn{6}{@{\hspace{4pt}}>{\centering\arraybackslash}m{\dimexpr0.72\textwidth+7.5pt\relax}@{}}{\includegraphics[width=\linewidth]{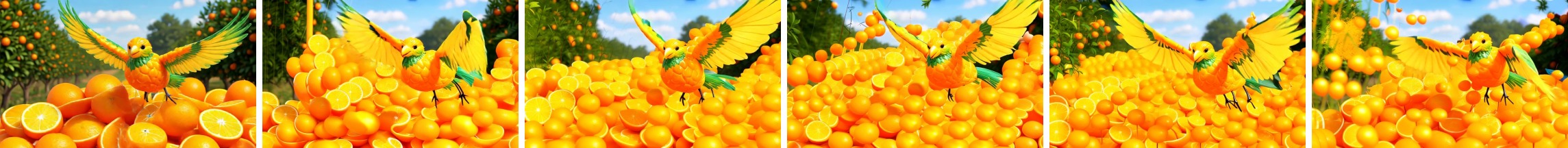}}\\
\end{tabular}
\endgroup
}
\caption{``Bird made of oranges''. Prompt: \emph{A dynamic and vibrant illustration in the style of a modern digital painting, depicting a bird crafted entirely from fresh oranges in vivid hues of orange and green. The bird is in flight, its wings spread wide and agile, as if rushing out of a large pile of freshly sliced oranges. The bird's body is composed of various sizes and shapes of oranges, with some segments showing the inner fruit flesh, adding a sense of freshness and vitality. Its eyes are small and black, giving it a lively and spirited expression. The background is a blurred orchard with rows of orange trees, some leaves fluttering in the wind. The scene captures the bird in mid-flight, with a slight tilt of its head and a spread of its wings, creating a sense of movement and energy. A close-up shot from a low angle, emphasizing the intricate details of the bird's form.}}
\label{fig:qual4}
\vspace{-3pt}
\scalebox{0.95}{
\begingroup
\setlength{\tabcolsep}{0.75pt}\renewcommand{\arraystretch}{1.0}
\sffamily\footnotesize
\begin{tabular}{@{}>{\raggedleft\arraybackslash}m{0.105\textwidth}@{\hspace{4pt}}*{6}{>{\centering\arraybackslash}m{0.12\textwidth}}@{}}
 & 0\,s & 48\,s & 96\,s & 144\,s & 192\,s & 238\,s\\[1pt]
CF++ & \multicolumn{6}{@{\hspace{4pt}}>{\centering\arraybackslash}m{\dimexpr0.72\textwidth+7.5pt\relax}@{}}{\includegraphics[width=\linewidth]{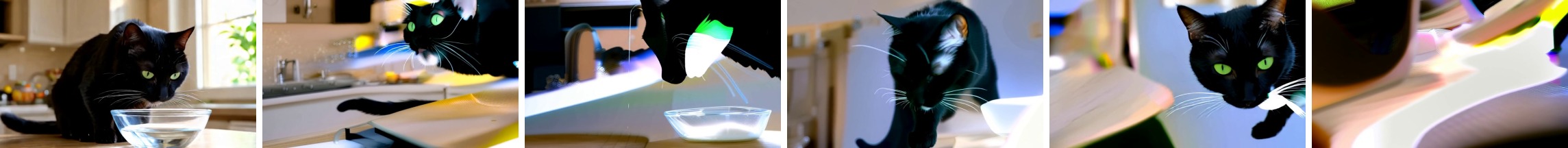}}\\
SGF & \multicolumn{6}{@{\hspace{4pt}}>{\centering\arraybackslash}m{\dimexpr0.72\textwidth+7.5pt\relax}@{}}{\includegraphics[width=\linewidth]{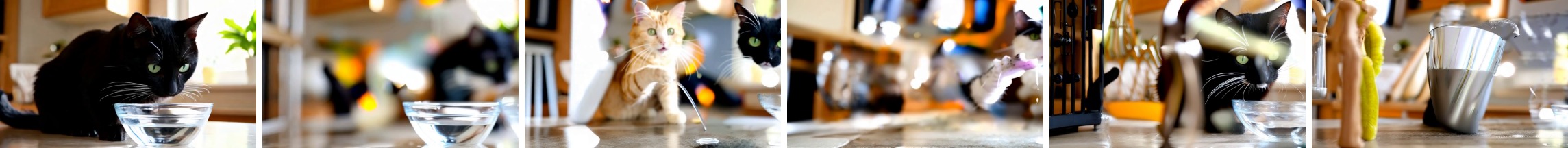}}\\
\rev{dKV Forcing w/o dropout} & \multicolumn{6}{@{\hspace{4pt}}>{\centering\arraybackslash}m{\dimexpr0.72\textwidth+7.5pt\relax}@{}}{\includegraphics[width=\linewidth]{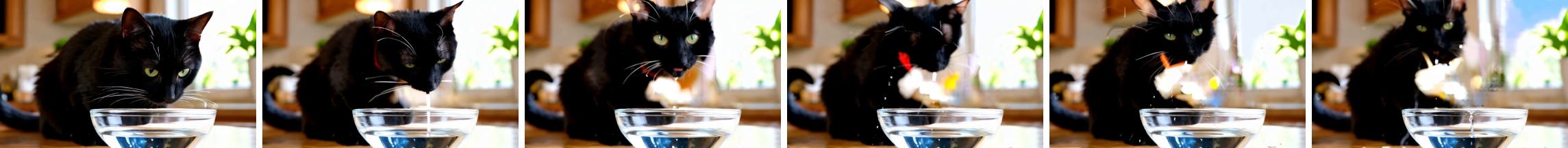}}\\
\textbf{dKV Forcing (ours)} & \multicolumn{6}{@{\hspace{4pt}}>{\centering\arraybackslash}m{\dimexpr0.72\textwidth+7.5pt\relax}@{}}{\includegraphics[width=\linewidth]{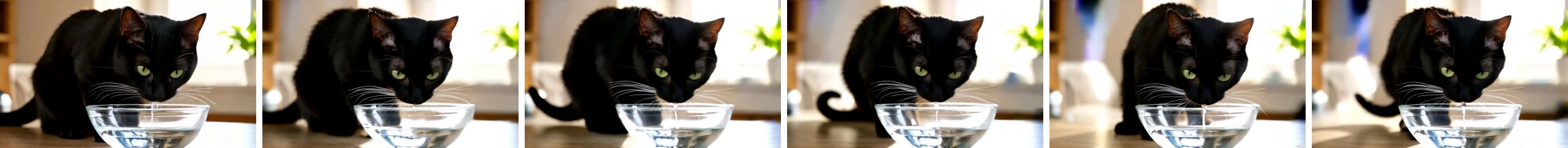}}\\
\end{tabular}
\endgroup
}
\caption{``Cat drinking water''. Prompt: \emph{a cat drinking water}}
\label{fig:qual5}
\end{figure*}

\begin{figure*}[p]
\centering
\captionsetup{skip=2pt,font=footnotesize}
\scalebox{0.95}{
\begingroup
\setlength{\tabcolsep}{0.75pt}\renewcommand{\arraystretch}{1.0}
\sffamily\footnotesize
\begin{tabular}{@{}>{\raggedleft\arraybackslash}m{0.105\textwidth}@{\hspace{4pt}}*{6}{>{\centering\arraybackslash}m{0.12\textwidth}}@{}}
 & 0\,s & 48\,s & 96\,s & 144\,s & 192\,s & 238\,s\\[1pt]
CF++ & \multicolumn{6}{@{\hspace{4pt}}>{\centering\arraybackslash}m{\dimexpr0.72\textwidth+7.5pt\relax}@{}}{\includegraphics[width=\linewidth]{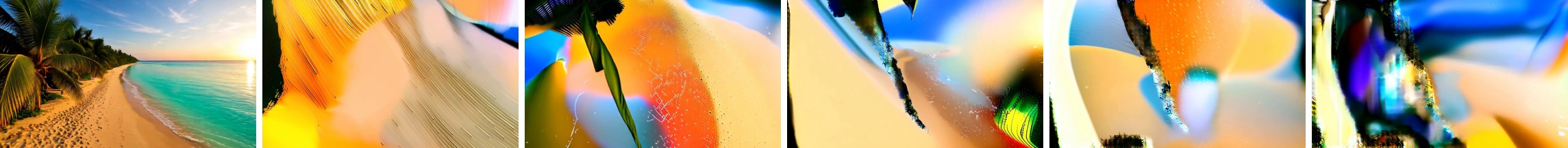}}\\
SGF & \multicolumn{6}{@{\hspace{4pt}}>{\centering\arraybackslash}m{\dimexpr0.72\textwidth+7.5pt\relax}@{}}{\includegraphics[width=\linewidth]{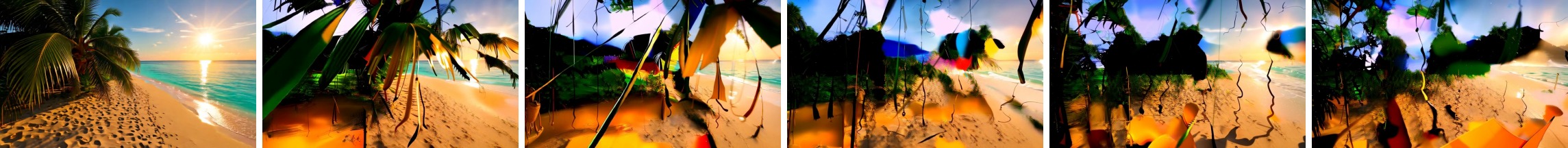}}\\
\rev{dKV Forcing w/o dropout} & \multicolumn{6}{@{\hspace{4pt}}>{\centering\arraybackslash}m{\dimexpr0.72\textwidth+7.5pt\relax}@{}}{\includegraphics[width=\linewidth]{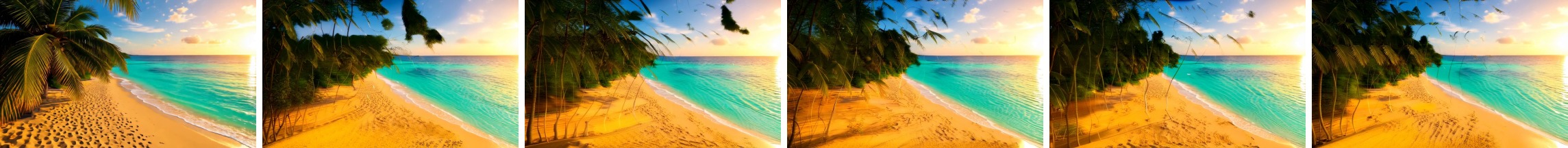}}\\
\textbf{dKV Forcing (ours)} & \multicolumn{6}{@{\hspace{4pt}}>{\centering\arraybackslash}m{\dimexpr0.72\textwidth+7.5pt\relax}@{}}{\includegraphics[width=\linewidth]{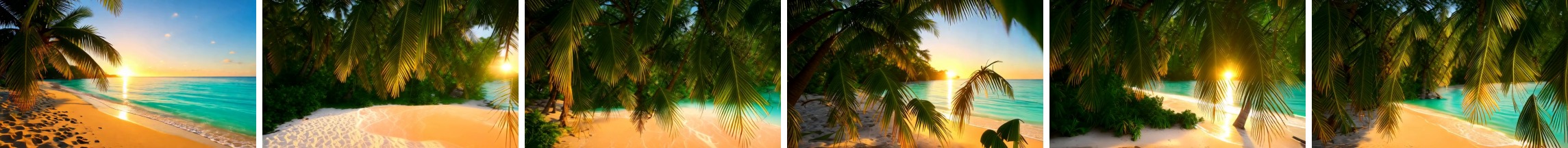}}\\
\end{tabular}
\endgroup
}
\caption{``Tropical beach at sunrise''. Prompt: \emph{A tropical beach at sunrise, with palm trees and crystal-clear water in the foreground}}
\label{fig:qual6}
\vspace{-3pt}
\scalebox{0.95}{
\begingroup
\setlength{\tabcolsep}{0.75pt}\renewcommand{\arraystretch}{1.0}
\sffamily\footnotesize
\begin{tabular}{@{}>{\raggedleft\arraybackslash}m{0.105\textwidth}@{\hspace{4pt}}*{6}{>{\centering\arraybackslash}m{0.12\textwidth}}@{}}
 & 0\,s & 48\,s & 96\,s & 144\,s & 192\,s & 238\,s\\[1pt]
CF++ & \multicolumn{6}{@{\hspace{4pt}}>{\centering\arraybackslash}m{\dimexpr0.72\textwidth+7.5pt\relax}@{}}{\includegraphics[width=\linewidth]{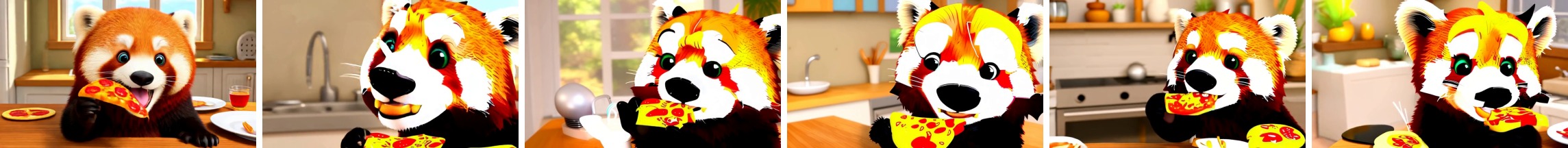}}\\
SGF & \multicolumn{6}{@{\hspace{4pt}}>{\centering\arraybackslash}m{\dimexpr0.72\textwidth+7.5pt\relax}@{}}{\includegraphics[width=\linewidth]{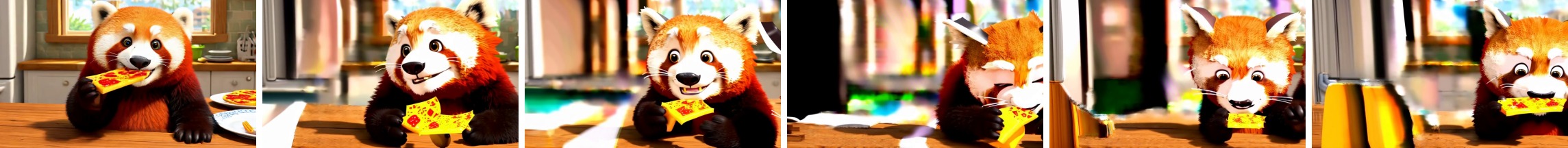}}\\
\rev{dKV Forcing w/o dropout} & \multicolumn{6}{@{\hspace{4pt}}>{\centering\arraybackslash}m{\dimexpr0.72\textwidth+7.5pt\relax}@{}}{\includegraphics[width=\linewidth]{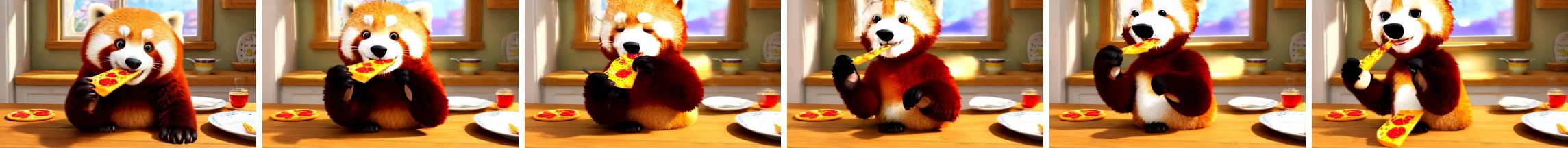}}\\
\textbf{dKV Forcing (ours)} & \multicolumn{6}{@{\hspace{4pt}}>{\centering\arraybackslash}m{\dimexpr0.72\textwidth+7.5pt\relax}@{}}{\includegraphics[width=\linewidth]{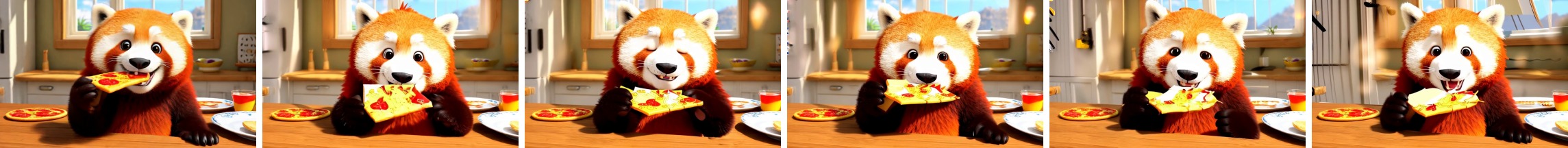}}\\
\end{tabular}
\endgroup
}
\caption{``Red panda eating pizza''. Prompt: \emph{A vibrant and lively illustration in a whimsical cartoon style of a red panda taking a bite of a pepperoni pizza. The red panda has a playful expression, with large, round eyes and a mischievous grin. It has reddish-brown fur and black markings around its eyes and muzzle. The panda is sitting upright on a wooden table, one paw holding the pizza slice while it takes a big bite. The background features a cozy kitchen setting with a few scattered plates and utensils, and a window letting in warm sunlight. A close-up shot from a slightly lower angle, capturing the panda's joyful moment.}}
\label{fig:qual7}
\vspace{-3pt}
\scalebox{0.95}{
\begingroup
\setlength{\tabcolsep}{0.75pt}\renewcommand{\arraystretch}{1.0}
\sffamily\footnotesize
\begin{tabular}{@{}>{\raggedleft\arraybackslash}m{0.105\textwidth}@{\hspace{4pt}}*{6}{>{\centering\arraybackslash}m{0.12\textwidth}}@{}}
 & 0\,s & 48\,s & 96\,s & 144\,s & 192\,s & 238\,s\\[1pt]
CF++ & \multicolumn{6}{@{\hspace{4pt}}>{\centering\arraybackslash}m{\dimexpr0.72\textwidth+7.5pt\relax}@{}}{\includegraphics[width=\linewidth]{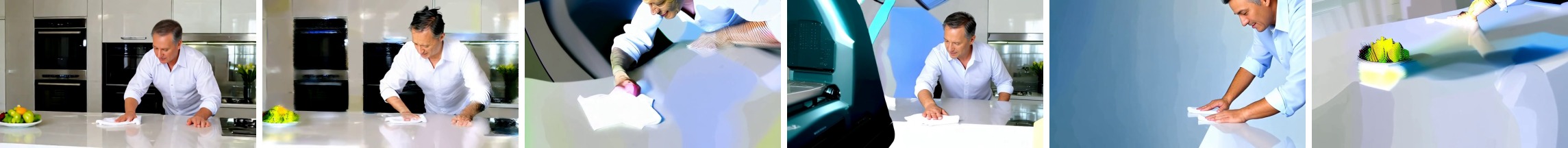}}\\
SGF & \multicolumn{6}{@{\hspace{4pt}}>{\centering\arraybackslash}m{\dimexpr0.72\textwidth+7.5pt\relax}@{}}{\includegraphics[width=\linewidth]{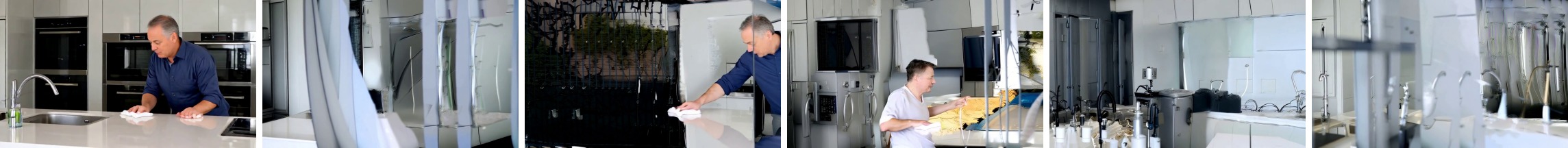}}\\
\rev{dKV Forcing w/o dropout} & \multicolumn{6}{@{\hspace{4pt}}>{\centering\arraybackslash}m{\dimexpr0.72\textwidth+7.5pt\relax}@{}}{\includegraphics[width=\linewidth]{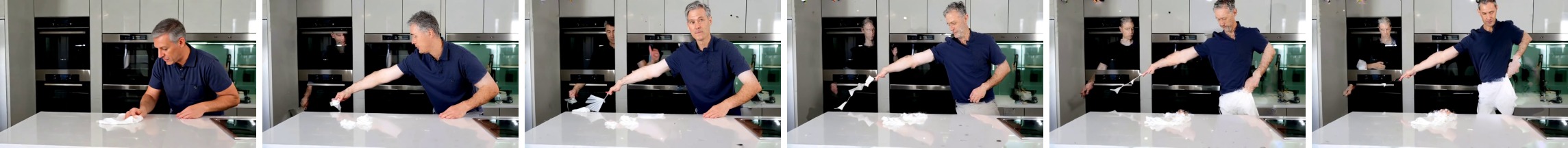}}\\
\textbf{dKV Forcing (ours)} & \multicolumn{6}{@{\hspace{4pt}}>{\centering\arraybackslash}m{\dimexpr0.72\textwidth+7.5pt\relax}@{}}{\includegraphics[width=\linewidth]{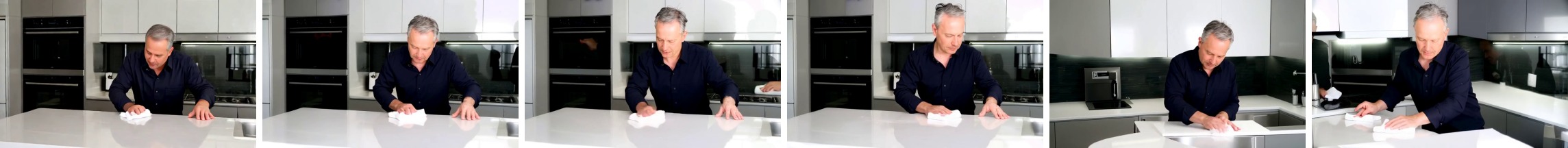}}\\
\end{tabular}
\endgroup
}
\caption{``Man wiping a kitchen counter''. Prompt: \emph{A detailed realist photograph captures a middle-aged man methodically wiping down a kitchen counter with a clean, white cloth. His focused expression conveys determination as he ensures every surface is spotlessly clean. He stands upright, leaning slightly forward, with one hand gripping the edge of the counter and the other holding the cloth. The background features modern kitchen appliances and cabinets, with subtle reflections in the glass surfaces. Shadows cast by the overhead lights add depth to the scene. The photo has a crisp, clear texture. A medium shot from a slightly elevated angle, highlighting the man's dedication and the pristine cleanliness of the kitchen.}}
\label{fig:qual8}
\end{figure*}

\begin{figure*}[p]
\centering
\captionsetup{skip=2pt,font=footnotesize}
\scalebox{0.95}{
\begingroup
\setlength{\tabcolsep}{0.75pt}\renewcommand{\arraystretch}{1.0}
\sffamily\footnotesize
\begin{tabular}{@{}>{\raggedleft\arraybackslash}m{0.105\textwidth}@{\hspace{4pt}}*{6}{>{\centering\arraybackslash}m{0.12\textwidth}}@{}}
 & 0\,s & 12\,s & 24\,s & 36\,s & 48\,s & 59\,s\\[1pt]
CF++ & \multicolumn{6}{@{\hspace{4pt}}>{\centering\arraybackslash}m{\dimexpr0.72\textwidth+7.5pt\relax}@{}}{\includegraphics[width=\linewidth]{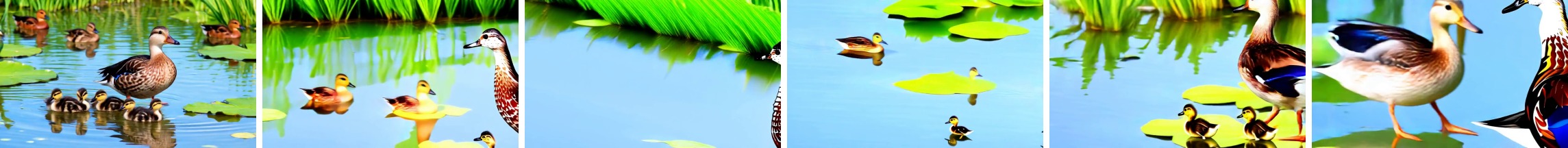}}\\
SGF & \multicolumn{6}{@{\hspace{4pt}}>{\centering\arraybackslash}m{\dimexpr0.72\textwidth+7.5pt\relax}@{}}{\includegraphics[width=\linewidth]{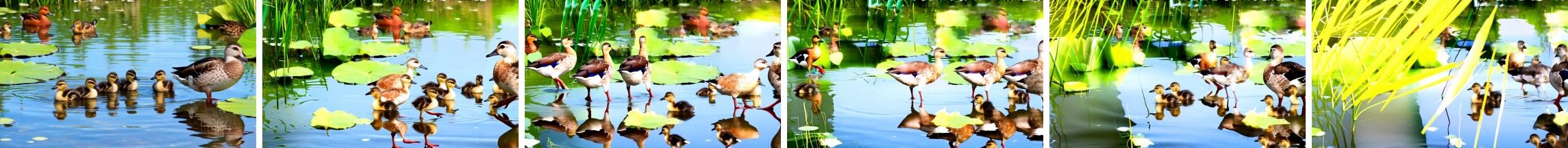}}\\
\rev{dKV Forcing w/o dropout} & \multicolumn{6}{@{\hspace{4pt}}>{\centering\arraybackslash}m{\dimexpr0.72\textwidth+7.5pt\relax}@{}}{\includegraphics[width=\linewidth]{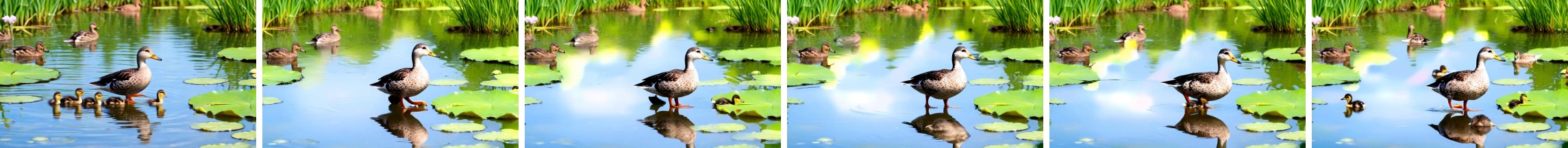}}\\
\textbf{dKV Forcing (ours)} & \multicolumn{6}{@{\hspace{4pt}}>{\centering\arraybackslash}m{\dimexpr0.72\textwidth+7.5pt\relax}@{}}{\includegraphics[width=\linewidth]{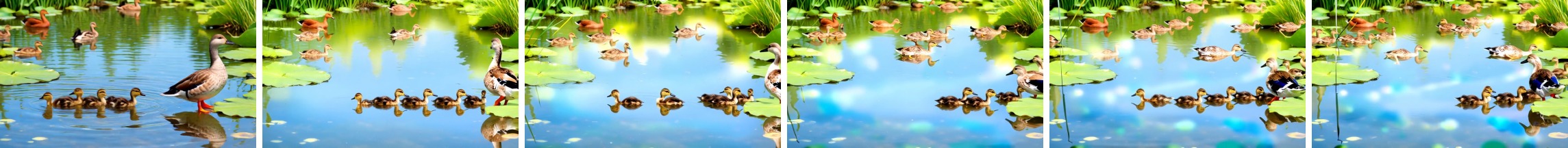}}\\
\end{tabular}
\endgroup
}
\caption{``Mother duck and six ducklings''. Prompt as in Figure~\ref{fig:qual1}.}
\label{fig:qualsixty1}
\vspace{-3pt}
\scalebox{0.95}{
\begingroup
\setlength{\tabcolsep}{0.75pt}\renewcommand{\arraystretch}{1.0}
\sffamily\footnotesize
\begin{tabular}{@{}>{\raggedleft\arraybackslash}m{0.105\textwidth}@{\hspace{4pt}}*{6}{>{\centering\arraybackslash}m{0.12\textwidth}}@{}}
 & 0\,s & 12\,s & 24\,s & 36\,s & 48\,s & 59\,s\\[1pt]
CF++ & \multicolumn{6}{@{\hspace{4pt}}>{\centering\arraybackslash}m{\dimexpr0.72\textwidth+7.5pt\relax}@{}}{\includegraphics[width=\linewidth]{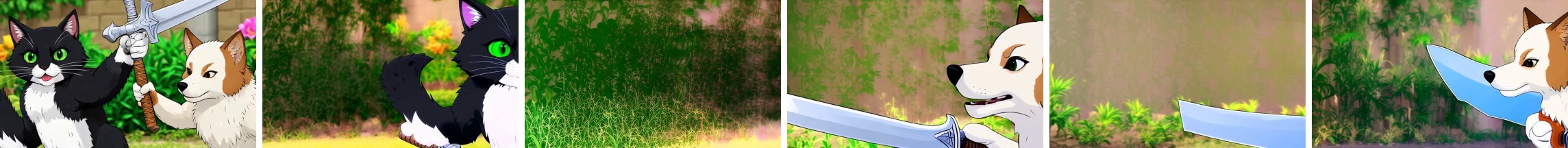}}\\
SGF & \multicolumn{6}{@{\hspace{4pt}}>{\centering\arraybackslash}m{\dimexpr0.72\textwidth+7.5pt\relax}@{}}{\includegraphics[width=\linewidth]{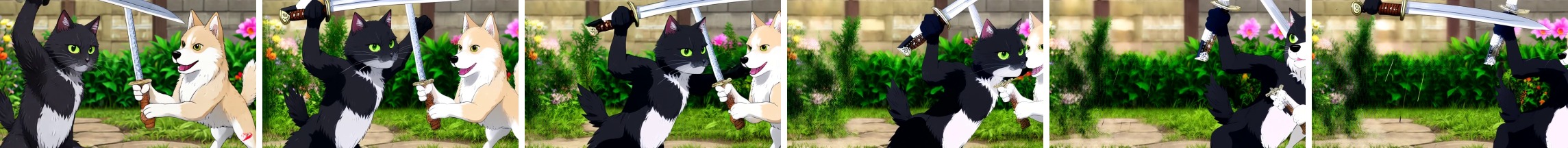}}\\
\rev{dKV Forcing w/o dropout} & \multicolumn{6}{@{\hspace{4pt}}>{\centering\arraybackslash}m{\dimexpr0.72\textwidth+7.5pt\relax}@{}}{\includegraphics[width=\linewidth]{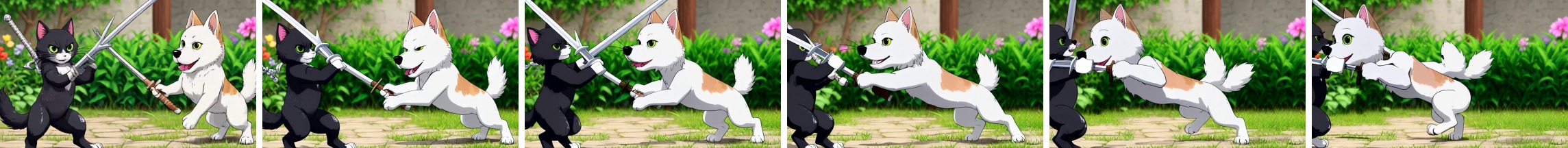}}\\
\textbf{dKV Forcing (ours)} & \multicolumn{6}{@{\hspace{4pt}}>{\centering\arraybackslash}m{\dimexpr0.72\textwidth+7.5pt\relax}@{}}{\includegraphics[width=\linewidth]{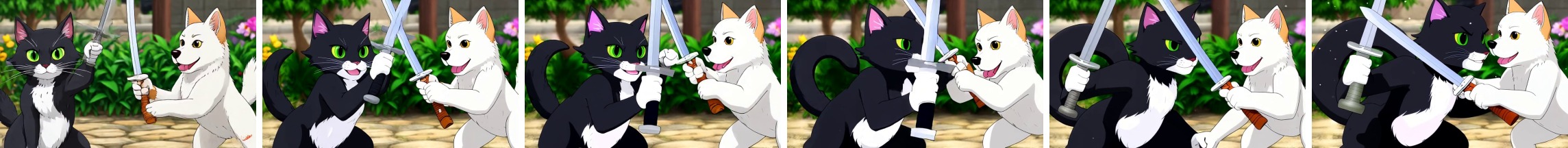}}\\
\end{tabular}
\endgroup
}
\caption{``Cat and dog in a sword fight''. Prompt: \emph{A dynamic illustration in a manga style depicting two cats and dogs engaged in a fierce sword fight. One cat, with sleek black fur and green eyes, holds a silver sword aloft, while the other, a fluffy white dog with brown eyes, lunges forward with a wooden sword. Both animals display intense focus and determination, their bodies tensed and ready for action. The background features a blurred garden setting with hints of green foliage and flowers. The scene is captured from a low-angle perspective, emphasizing the movement and energy of the battle.}}
\label{fig:qualsixty2}
\vspace{-3pt}
\scalebox{0.95}{
\begingroup
\setlength{\tabcolsep}{0.75pt}\renewcommand{\arraystretch}{1.0}
\sffamily\footnotesize
\begin{tabular}{@{}>{\raggedleft\arraybackslash}m{0.105\textwidth}@{\hspace{4pt}}*{6}{>{\centering\arraybackslash}m{0.12\textwidth}}@{}}
 & 0\,s & 12\,s & 24\,s & 36\,s & 48\,s & 59\,s\\[1pt]
CF++ & \multicolumn{6}{@{\hspace{4pt}}>{\centering\arraybackslash}m{\dimexpr0.72\textwidth+7.5pt\relax}@{}}{\includegraphics[width=\linewidth]{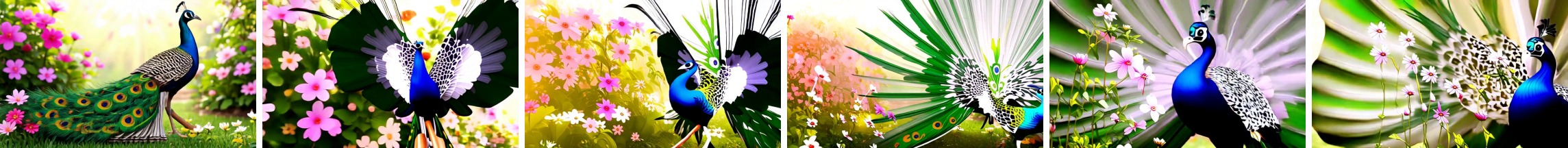}}\\
SGF & \multicolumn{6}{@{\hspace{4pt}}>{\centering\arraybackslash}m{\dimexpr0.72\textwidth+7.5pt\relax}@{}}{\includegraphics[width=\linewidth]{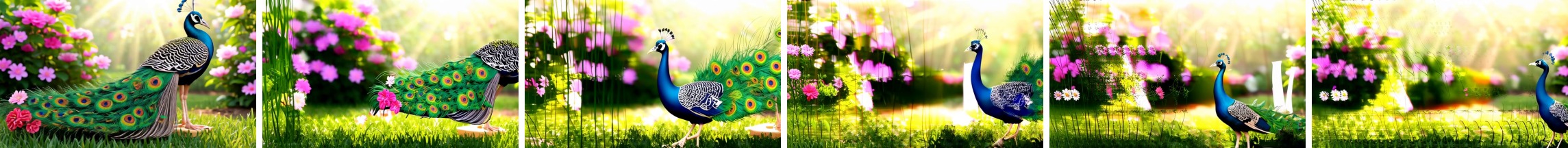}}\\
\rev{dKV Forcing w/o dropout} & \multicolumn{6}{@{\hspace{4pt}}>{\centering\arraybackslash}m{\dimexpr0.72\textwidth+7.5pt\relax}@{}}{\includegraphics[width=\linewidth]{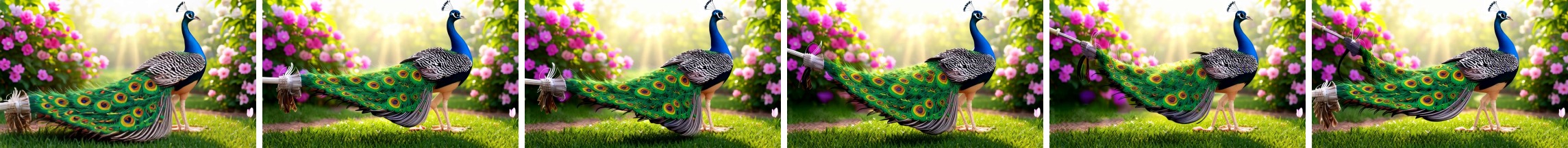}}\\
\textbf{dKV Forcing (ours)} & \multicolumn{6}{@{\hspace{4pt}}>{\centering\arraybackslash}m{\dimexpr0.72\textwidth+7.5pt\relax}@{}}{\includegraphics[width=\linewidth]{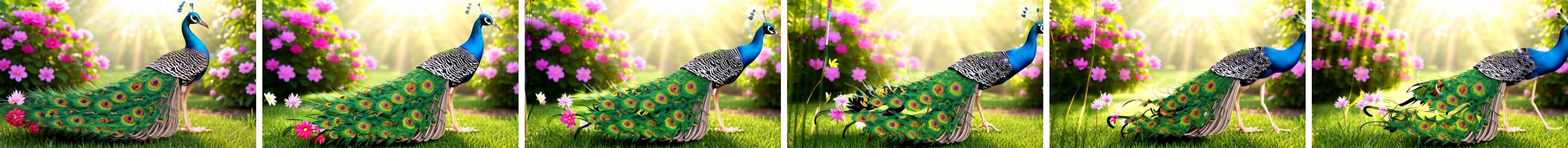}}\\
\end{tabular}
\endgroup
}
\caption{``Broom morphing into a peacock''. Prompt: \emph{A dynamic motion shot of a broom morphing surreally and magically into a peacock. The broom handle shortens and curves into a slender neck, while the bristles fan out into a magnificent tail. Vibrant colors and eye-shaped patterns emerge on the expanding feathers as the camera moves around, capturing every detail. A small head forms at the top, complete with a delicate crest. The transformation completes as the peacock proudly displays its newly formed plumage, standing tall and regal. The background features a soft, pastel garden with blooming flowers and gentle sunlight filtering through, enhancing the ethereal quality of the scene.}}
\label{fig:qualsixty3}
\end{figure*}

\begin{figure*}[p]
\centering
\captionsetup{skip=2pt,font=footnotesize}
\scalebox{0.95}{
\begingroup
\setlength{\tabcolsep}{0.75pt}\renewcommand{\arraystretch}{1.0}
\sffamily\footnotesize
\begin{tabular}{@{}>{\raggedleft\arraybackslash}m{0.105\textwidth}@{\hspace{4pt}}*{6}{>{\centering\arraybackslash}m{0.12\textwidth}}@{}}
 & 0\,s & 12\,s & 24\,s & 36\,s & 48\,s & 59\,s\\[1pt]
CF++ & \multicolumn{6}{@{\hspace{4pt}}>{\centering\arraybackslash}m{\dimexpr0.72\textwidth+7.5pt\relax}@{}}{\includegraphics[width=\linewidth]{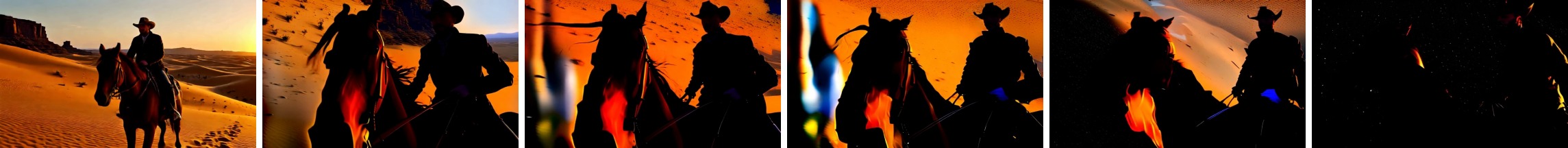}}\\
SGF & \multicolumn{6}{@{\hspace{4pt}}>{\centering\arraybackslash}m{\dimexpr0.72\textwidth+7.5pt\relax}@{}}{\includegraphics[width=\linewidth]{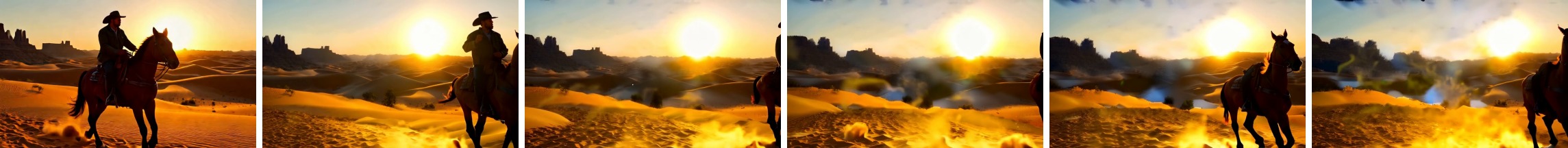}}\\
\rev{dKV Forcing w/o dropout} & \multicolumn{6}{@{\hspace{4pt}}>{\centering\arraybackslash}m{\dimexpr0.72\textwidth+7.5pt\relax}@{}}{\includegraphics[width=\linewidth]{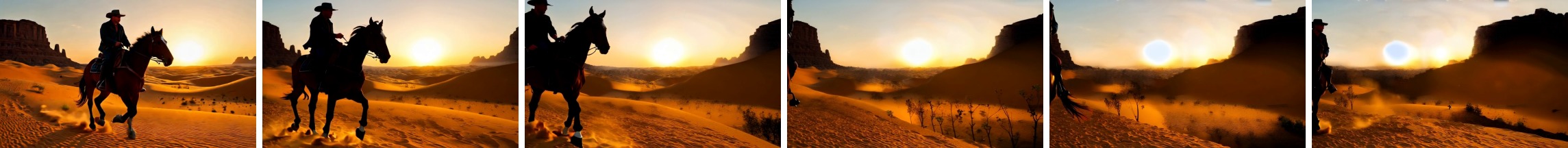}}\\
\textbf{dKV Forcing (ours)} & \multicolumn{6}{@{\hspace{4pt}}>{\centering\arraybackslash}m{\dimexpr0.72\textwidth+7.5pt\relax}@{}}{\includegraphics[width=\linewidth]{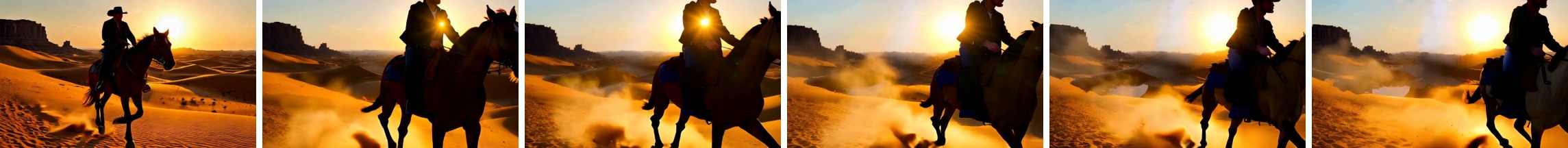}}\\
\end{tabular}
\endgroup
}
\caption{``Rider in the Gobi Desert at sunset''. Prompt: \emph{A cinematic scene from a classic western movie, featuring a rugged man riding a powerful horse through the vast Gobi Desert at sunset. The man, dressed in a dusty cowboy hat and a worn leather jacket, reins tightly on the horse's neck as he gallops across the golden sands. The sun sets dramatically behind them, casting long shadows and warm hues across the landscape. The background is filled with rolling dunes and sparse, rocky outcrops, emphasizing the harsh beauty of the desert. A dynamic wide shot from a low angle, capturing both the man and the expansive desert vista.}}
\label{fig:qualsixty4}
\vspace{-3pt}
\scalebox{0.95}{
\begingroup
\setlength{\tabcolsep}{0.75pt}\renewcommand{\arraystretch}{1.0}
\sffamily\footnotesize
\begin{tabular}{@{}>{\raggedleft\arraybackslash}m{0.105\textwidth}@{\hspace{4pt}}*{6}{>{\centering\arraybackslash}m{0.12\textwidth}}@{}}
 & 0\,s & 12\,s & 24\,s & 36\,s & 48\,s & 59\,s\\[1pt]
CF++ & \multicolumn{6}{@{\hspace{4pt}}>{\centering\arraybackslash}m{\dimexpr0.72\textwidth+7.5pt\relax}@{}}{\includegraphics[width=\linewidth]{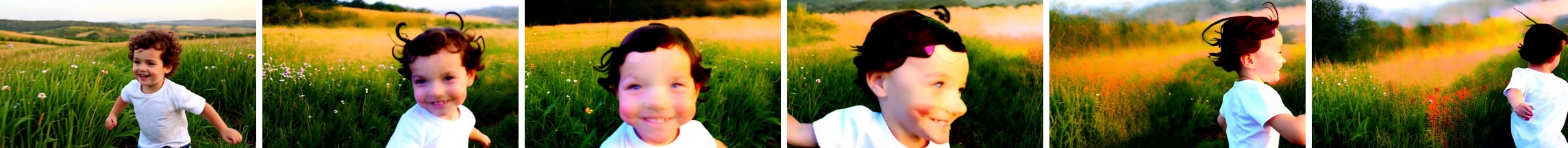}}\\
SGF & \multicolumn{6}{@{\hspace{4pt}}>{\centering\arraybackslash}m{\dimexpr0.72\textwidth+7.5pt\relax}@{}}{\includegraphics[width=\linewidth]{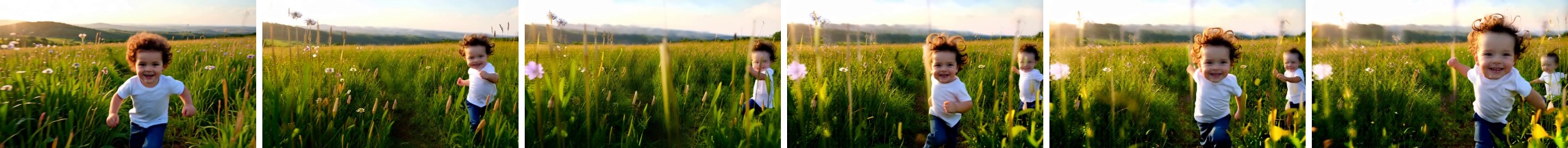}}\\
\rev{dKV Forcing w/o dropout} & \multicolumn{6}{@{\hspace{4pt}}>{\centering\arraybackslash}m{\dimexpr0.72\textwidth+7.5pt\relax}@{}}{\includegraphics[width=\linewidth]{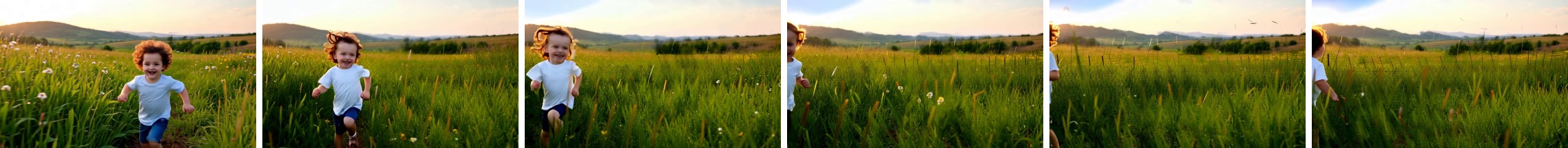}}\\
\textbf{dKV Forcing (ours)} & \multicolumn{6}{@{\hspace{4pt}}>{\centering\arraybackslash}m{\dimexpr0.72\textwidth+7.5pt\relax}@{}}{\includegraphics[width=\linewidth]{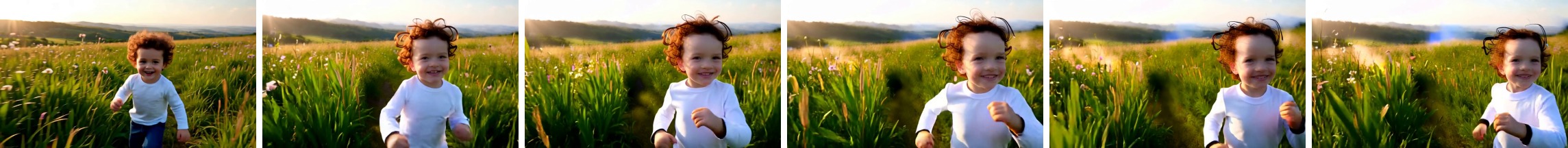}}\\
\end{tabular}
\endgroup
}
\caption{``Child running through tall grass''. Prompt: \emph{A handheld shot following a young child running through a field of tall grass, capturing the spontaneity and playfulness of their movements. The child has curly brown hair and a mischievous smile, arms swinging freely as they sprint across the green expanse. Their small feet kick up bits of grass and dirt, creating a trail behind them. The background features a blurred landscape with rolling hills and scattered wildflowers, bathed in warm sunlight. The photo has a natural, documentary-style quality, emphasizing the dynamic motion and joy of the moment. A dynamic handheld shot from a slightly elevated angle, following the child's energetic run.}}
\label{fig:qualsixty5}
\vspace{-3pt}
\scalebox{0.95}{
\begingroup
\setlength{\tabcolsep}{0.75pt}\renewcommand{\arraystretch}{1.0}
\sffamily\footnotesize
\begin{tabular}{@{}>{\raggedleft\arraybackslash}m{0.105\textwidth}@{\hspace{4pt}}*{6}{>{\centering\arraybackslash}m{0.12\textwidth}}@{}}
 & 0\,s & 12\,s & 24\,s & 36\,s & 48\,s & 59\,s\\[1pt]
CF++ & \multicolumn{6}{@{\hspace{4pt}}>{\centering\arraybackslash}m{\dimexpr0.72\textwidth+7.5pt\relax}@{}}{\includegraphics[width=\linewidth]{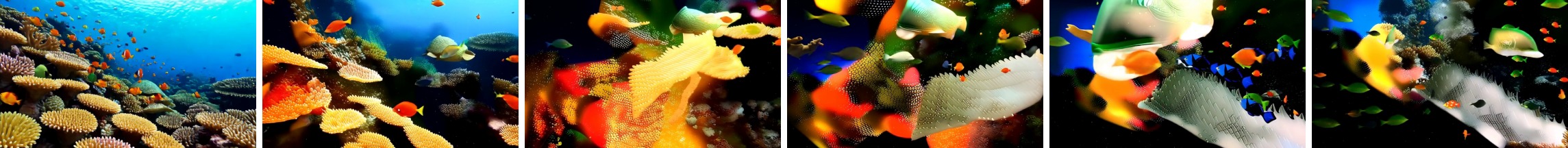}}\\
SGF & \multicolumn{6}{@{\hspace{4pt}}>{\centering\arraybackslash}m{\dimexpr0.72\textwidth+7.5pt\relax}@{}}{\includegraphics[width=\linewidth]{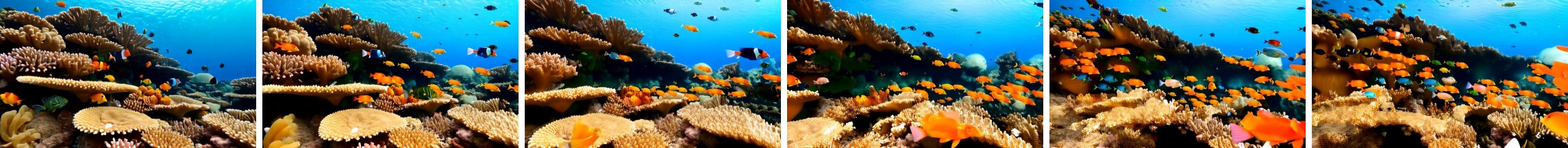}}\\
\rev{dKV Forcing w/o dropout} & \multicolumn{6}{@{\hspace{4pt}}>{\centering\arraybackslash}m{\dimexpr0.72\textwidth+7.5pt\relax}@{}}{\includegraphics[width=\linewidth]{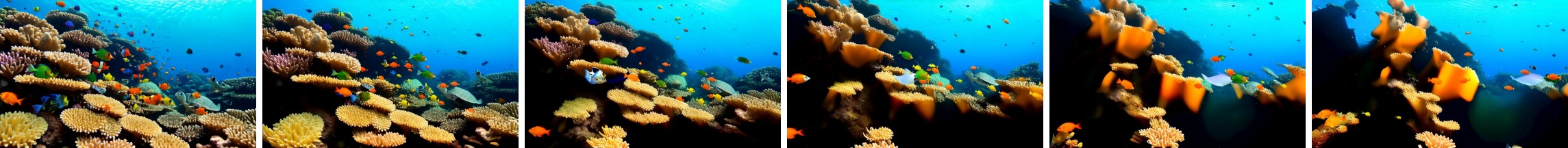}}\\
\textbf{dKV Forcing (ours)} & \multicolumn{6}{@{\hspace{4pt}}>{\centering\arraybackslash}m{\dimexpr0.72\textwidth+7.5pt\relax}@{}}{\includegraphics[width=\linewidth]{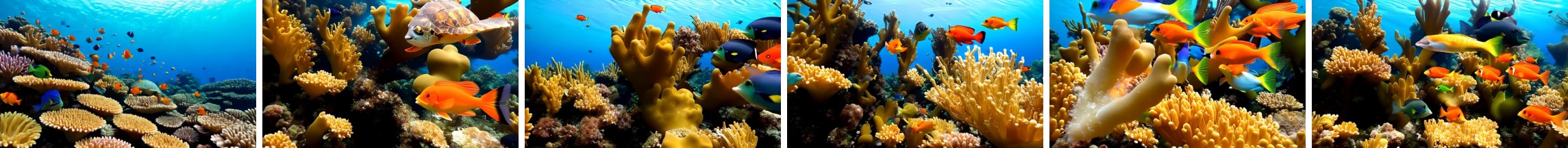}}\\
\end{tabular}
\endgroup
}
\caption{``Coral reef with fish and a turtle''. Prompt: \emph{A vibrant coral reef teeming with colorful fish and marine life, captured in a dynamic underwater scene. The reef is adorned with a variety of coral shapes and sizes, creating a lively and intricate landscape. Schools of fish dart between the corals, their vibrant hues ranging from electric blues and greens to fiery oranges and yellows. A schools of clownfish can be seen swimming among the anemones, while a majestic turtle glides gracefully nearby. The water has a clear, turquoise hue, allowing the viewer to see the rich biodiversity beneath the surface. The camera angle is slightly from below, emphasizing the vertical depth and the bustling activity of the reef.}}
\label{fig:qualsixty6}
\end{figure*}

\begin{figure*}[p]
\centering
\captionsetup{skip=2pt,font=footnotesize}
\scalebox{0.95}{
\begingroup
\setlength{\tabcolsep}{0.75pt}\renewcommand{\arraystretch}{1.0}
\sffamily\footnotesize
\begin{tabular}{@{}>{\raggedleft\arraybackslash}m{0.105\textwidth}@{\hspace{4pt}}*{6}{>{\centering\arraybackslash}m{0.12\textwidth}}@{}}
 & 0\,s & 12\,s & 24\,s & 36\,s & 48\,s & 59\,s\\[1pt]
CF++ & \multicolumn{6}{@{\hspace{4pt}}>{\centering\arraybackslash}m{\dimexpr0.72\textwidth+7.5pt\relax}@{}}{\includegraphics[width=\linewidth]{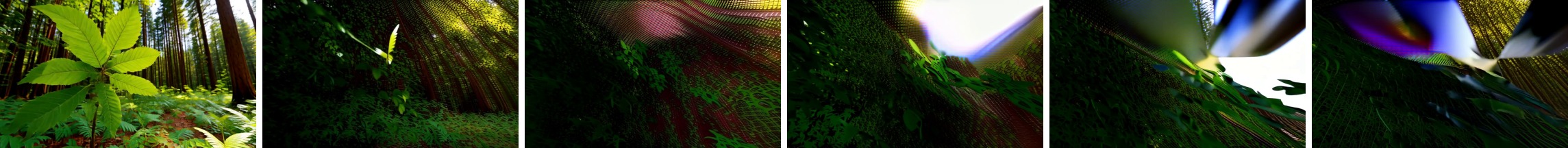}}\\
SGF & \multicolumn{6}{@{\hspace{4pt}}>{\centering\arraybackslash}m{\dimexpr0.72\textwidth+7.5pt\relax}@{}}{\includegraphics[width=\linewidth]{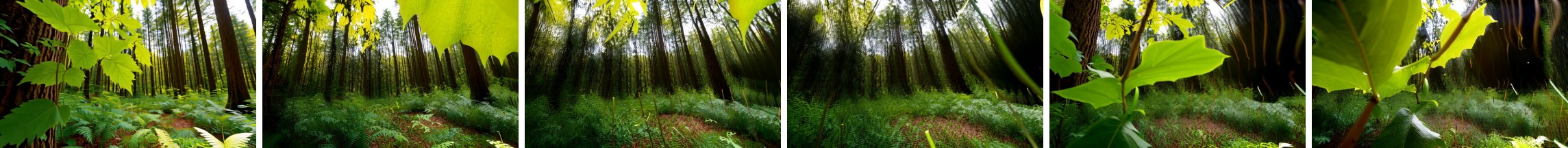}}\\
\rev{dKV Forcing w/o dropout} & \multicolumn{6}{@{\hspace{4pt}}>{\centering\arraybackslash}m{\dimexpr0.72\textwidth+7.5pt\relax}@{}}{\includegraphics[width=\linewidth]{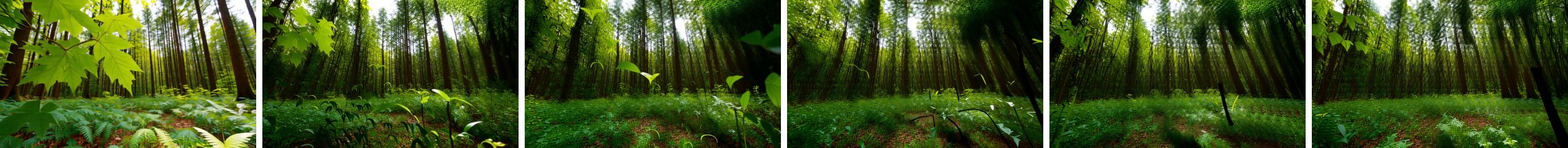}}\\
\textbf{dKV Forcing (ours)} & \multicolumn{6}{@{\hspace{4pt}}>{\centering\arraybackslash}m{\dimexpr0.72\textwidth+7.5pt\relax}@{}}{\includegraphics[width=\linewidth]{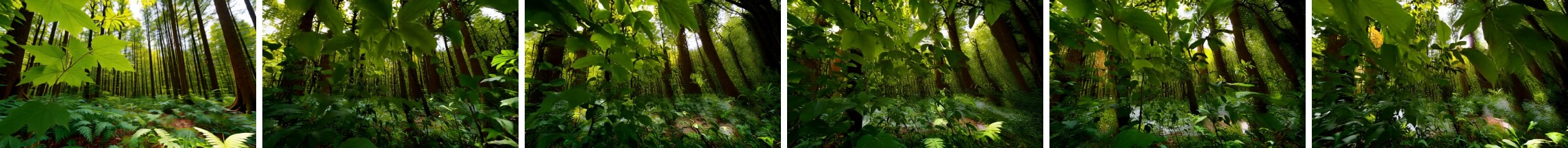}}\\
\end{tabular}
\endgroup
}
\caption{``Zoom-out from a leaf to the forest''. Prompt: \emph{A zoom-out from a single emerald green leaf on a tree to reveal the entire forest, capturing the vastness and diversity of the woodland. The scene starts with a close-up of the leaf, highlighting its intricate vein patterns and vibrant color. As the camera pulls back, it reveals a dense forest with towering trees, their trunks varying in size and shape, and a rich tapestry of foliage in various shades of green, brown, and gold. Ferns and wildflowers dot the forest floor, adding to the natural beauty. The sunlight filters through the canopy, casting dappled shadows on the ground. The forest exudes a serene and tranquil atmosphere, with birds chirping in the background. A wide-angle shot with a gradual pull-back.}}
\label{fig:qualsixty7}
\vspace{-3pt}
\scalebox{0.95}{
\begingroup
\setlength{\tabcolsep}{0.75pt}\renewcommand{\arraystretch}{1.0}
\sffamily\footnotesize
\begin{tabular}{@{}>{\raggedleft\arraybackslash}m{0.105\textwidth}@{\hspace{4pt}}*{6}{>{\centering\arraybackslash}m{0.12\textwidth}}@{}}
 & 0\,s & 12\,s & 24\,s & 36\,s & 48\,s & 59\,s\\[1pt]
CF++ & \multicolumn{6}{@{\hspace{4pt}}>{\centering\arraybackslash}m{\dimexpr0.72\textwidth+7.5pt\relax}@{}}{\includegraphics[width=\linewidth]{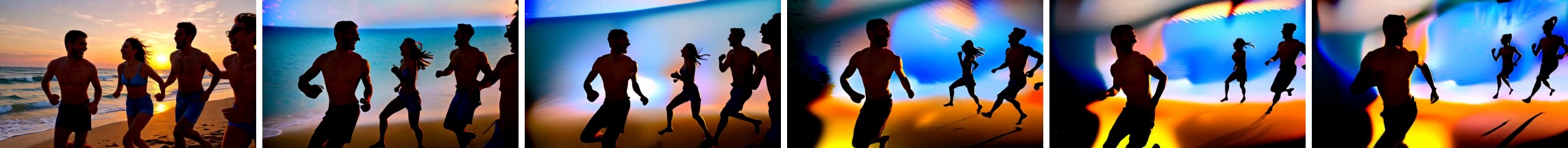}}\\
SGF & \multicolumn{6}{@{\hspace{4pt}}>{\centering\arraybackslash}m{\dimexpr0.72\textwidth+7.5pt\relax}@{}}{\includegraphics[width=\linewidth]{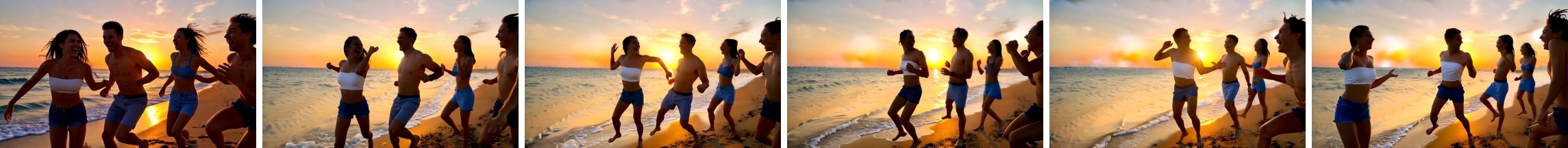}}\\
\rev{dKV Forcing w/o dropout} & \multicolumn{6}{@{\hspace{4pt}}>{\centering\arraybackslash}m{\dimexpr0.72\textwidth+7.5pt\relax}@{}}{\includegraphics[width=\linewidth]{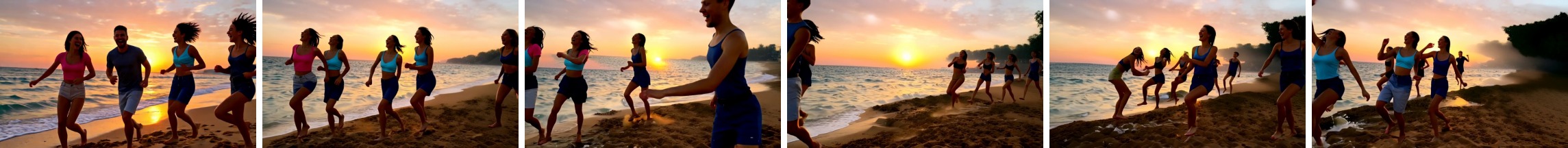}}\\
\textbf{dKV Forcing (ours)} & \multicolumn{6}{@{\hspace{4pt}}>{\centering\arraybackslash}m{\dimexpr0.72\textwidth+7.5pt\relax}@{}}{\includegraphics[width=\linewidth]{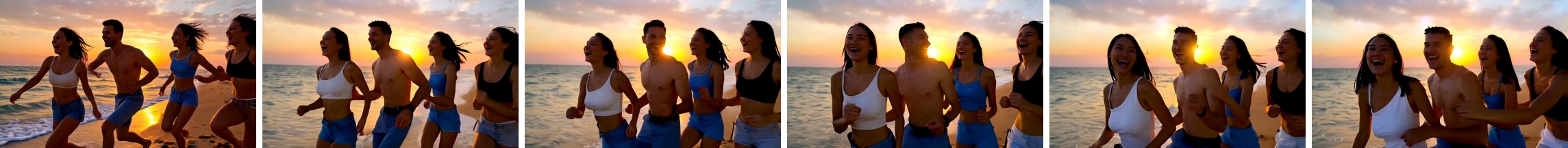}}\\
\end{tabular}
\endgroup
}
\caption{``Friends on the beach at sunset''. Prompt: \emph{A handheld shot capturing a group of friends laughing and playing on the beach at sunset. The friends, with joyful expressions, run and play among the waves, their laughter echoing across the golden sand. The camera follows them closely, emphasizing their lively movements and the joyous atmosphere. The background features a breathtaking sunset, with orange and pink hues blending into the horizon, and the gentle sea breeze blowing through. The sand is warm and the water sparkles under the setting sun. A dynamic and energetic scene, with the friends' silhouettes gradually becoming softer against the fading light.}}
\label{fig:qualsixty8}
\vspace{-3pt}
\scalebox{0.95}{
\begingroup
\setlength{\tabcolsep}{0.75pt}\renewcommand{\arraystretch}{1.0}
\sffamily\footnotesize
\begin{tabular}{@{}>{\raggedleft\arraybackslash}m{0.105\textwidth}@{\hspace{4pt}}*{6}{>{\centering\arraybackslash}m{0.12\textwidth}}@{}}
 & 0\,s & 12\,s & 24\,s & 36\,s & 48\,s & 59\,s\\[1pt]
CF++ & \multicolumn{6}{@{\hspace{4pt}}>{\centering\arraybackslash}m{\dimexpr0.72\textwidth+7.5pt\relax}@{}}{\includegraphics[width=\linewidth]{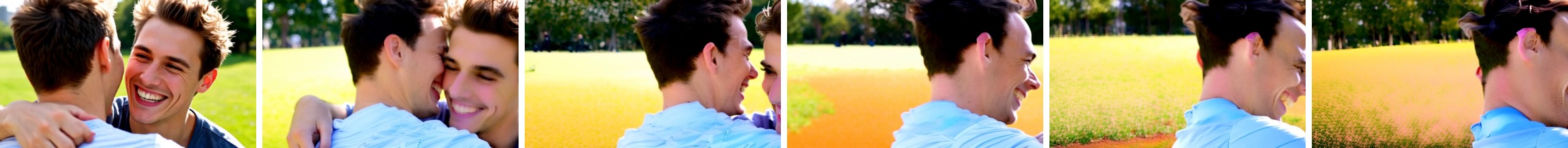}}\\
SGF & \multicolumn{6}{@{\hspace{4pt}}>{\centering\arraybackslash}m{\dimexpr0.72\textwidth+7.5pt\relax}@{}}{\includegraphics[width=\linewidth]{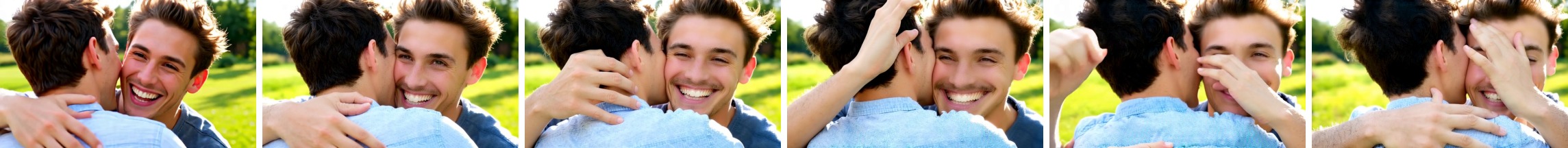}}\\
\rev{dKV Forcing w/o dropout} & \multicolumn{6}{@{\hspace{4pt}}>{\centering\arraybackslash}m{\dimexpr0.72\textwidth+7.5pt\relax}@{}}{\includegraphics[width=\linewidth]{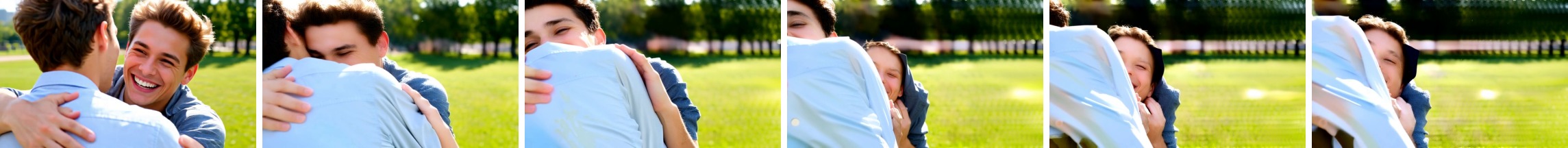}}\\
\textbf{dKV Forcing (ours)} & \multicolumn{6}{@{\hspace{4pt}}>{\centering\arraybackslash}m{\dimexpr0.72\textwidth+7.5pt\relax}@{}}{\includegraphics[width=\linewidth]{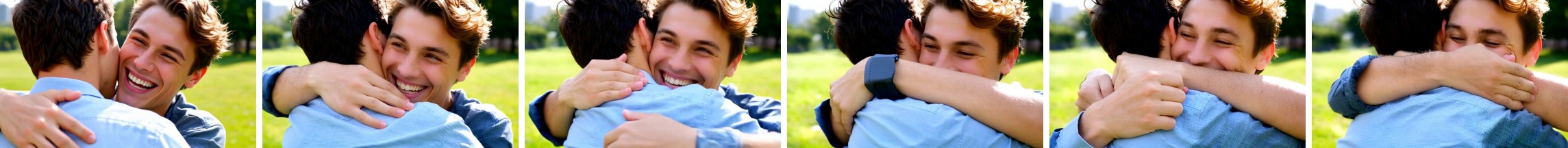}}\\
\end{tabular}
\endgroup
}
\caption{``Young man hugging a friend''. Prompt: \emph{A vibrant and lively photograph capturing a moment of genuine joy, depicting a young man with a radiant smile as he hugs a dear friend. His face is filled with happiness, with his eyes sparkling and cheeks flushed. He wears a casual shirt and jeans, and his hair is tousled, adding to the natural and relaxed feeling. His friend, equally joyful, returns the embrace with equal warmth. The background is a blurred setting, hinting at a park or a sunny outdoor space, with patches of green grass and trees in the distance. The photo has a warm and candid feel, with soft lighting and natural shadows. A close-up shot from a slightly angled perspective, emphasizing their heartfelt connection.}}
\label{fig:qualsixty9}
\end{figure*}

\begin{figure*}[p]
\centering
\captionsetup{skip=2pt,font=footnotesize}
\scalebox{0.95}{
\begingroup
\setlength{\tabcolsep}{0.75pt}\renewcommand{\arraystretch}{1.0}
\sffamily\footnotesize
\begin{tabular}{@{}>{\raggedleft\arraybackslash}m{0.105\textwidth}@{\hspace{4pt}}*{6}{>{\centering\arraybackslash}m{0.12\textwidth}}@{}}
 & 0\,s & 12\,s & 24\,s & 36\,s & 48\,s & 59\,s\\[1pt]
CF++ & \multicolumn{6}{@{\hspace{4pt}}>{\centering\arraybackslash}m{\dimexpr0.72\textwidth+7.5pt\relax}@{}}{\includegraphics[width=\linewidth]{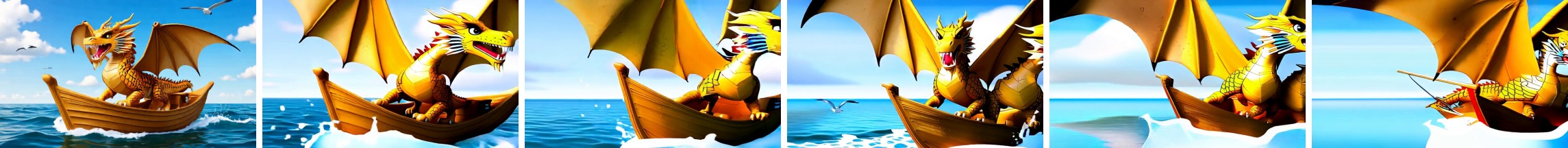}}\\
SGF & \multicolumn{6}{@{\hspace{4pt}}>{\centering\arraybackslash}m{\dimexpr0.72\textwidth+7.5pt\relax}@{}}{\includegraphics[width=\linewidth]{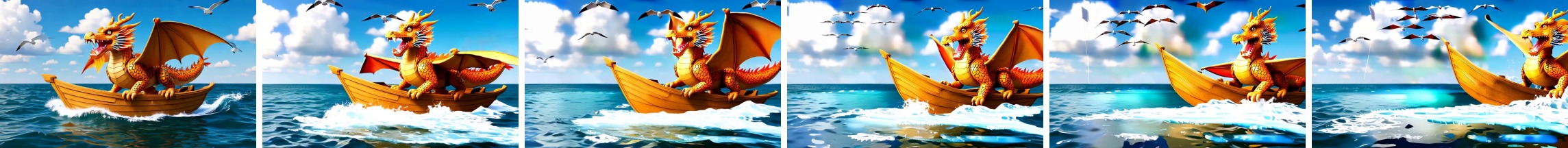}}\\
\rev{dKV Forcing w/o dropout} & \multicolumn{6}{@{\hspace{4pt}}>{\centering\arraybackslash}m{\dimexpr0.72\textwidth+7.5pt\relax}@{}}{\includegraphics[width=\linewidth]{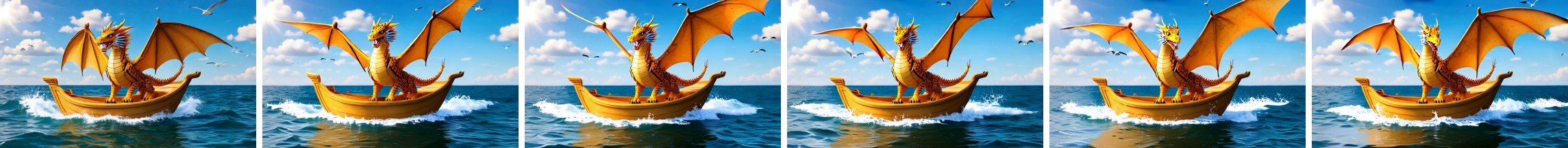}}\\
\textbf{dKV Forcing (ours)} & \multicolumn{6}{@{\hspace{4pt}}>{\centering\arraybackslash}m{\dimexpr0.72\textwidth+7.5pt\relax}@{}}{\includegraphics[width=\linewidth]{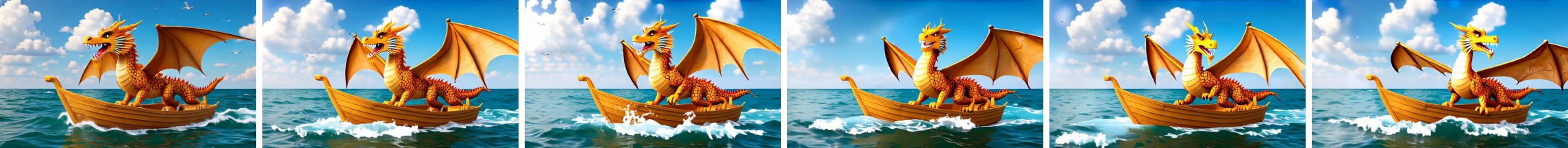}}\\
\end{tabular}
\endgroup
}
\caption{``Origami dragon riding a boat''. Prompt: \emph{A realistic style paper origami dragon riding a boat through waves, with intricate folds and textures. The dragon has a fierce expression, its eyes glowing with intensity, and its scales shimmering in the sunlight. It is perched on the edge of the boat, wings partially spread, ready to take flight. The boat bobs up and down with the waves, creating a dynamic motion. The water is choppy, with ripples and splashes around the boat, adding to the sense of movement. The background features a clear blue sky with fluffy clouds, and a few seagulls flying overhead. A mid-shot capturing the dragon's powerful stance and the boat's motion.}}
\label{fig:qualsixty10}
\vspace{-3pt}
\scalebox{0.95}{
\begingroup
\setlength{\tabcolsep}{0.75pt}\renewcommand{\arraystretch}{1.0}
\sffamily\footnotesize
\begin{tabular}{@{}>{\raggedleft\arraybackslash}m{0.105\textwidth}@{\hspace{4pt}}*{6}{>{\centering\arraybackslash}m{0.12\textwidth}}@{}}
 & 0\,s & 12\,s & 24\,s & 36\,s & 48\,s & 59\,s\\[1pt]
CF++ & \multicolumn{6}{@{\hspace{4pt}}>{\centering\arraybackslash}m{\dimexpr0.72\textwidth+7.5pt\relax}@{}}{\includegraphics[width=\linewidth]{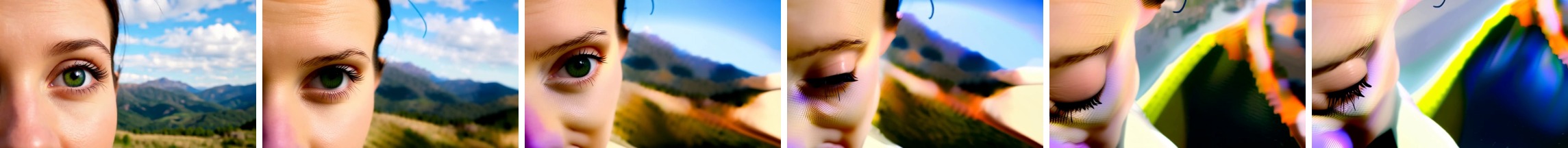}}\\
SGF & \multicolumn{6}{@{\hspace{4pt}}>{\centering\arraybackslash}m{\dimexpr0.72\textwidth+7.5pt\relax}@{}}{\includegraphics[width=\linewidth]{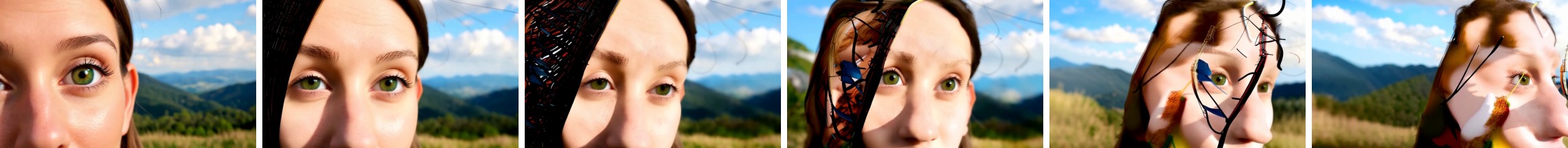}}\\
\rev{dKV Forcing w/o dropout} & \multicolumn{6}{@{\hspace{4pt}}>{\centering\arraybackslash}m{\dimexpr0.72\textwidth+7.5pt\relax}@{}}{\includegraphics[width=\linewidth]{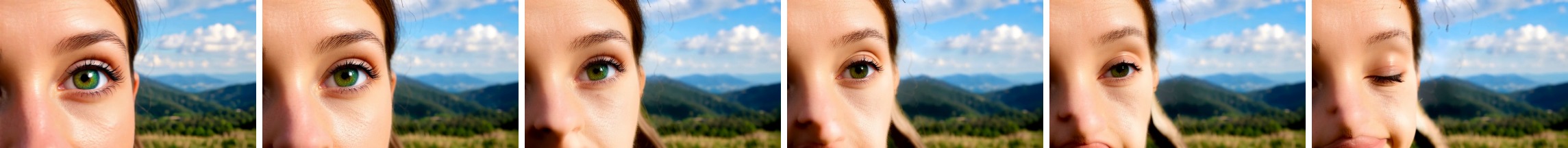}}\\
\textbf{dKV Forcing (ours)} & \multicolumn{6}{@{\hspace{4pt}}>{\centering\arraybackslash}m{\dimexpr0.72\textwidth+7.5pt\relax}@{}}{\includegraphics[width=\linewidth]{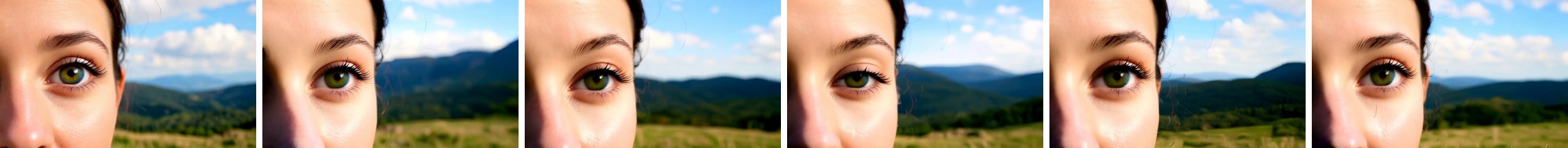}}\\
\end{tabular}
\endgroup
}
\caption{``Eye with the earth as its iris''. Prompt: \emph{An extreme close-up shot of a woman's eye, where her iris appears to be a vivid representation of the earth, with rich greens, blues, and browns blending together. Her large, expressive eyes capture a sense of wonder and depth, with the pupils slightly narrowed, hinting at a thoughtful gaze. The background is a blurred, natural landscape with distant mountains and rolling hills, giving the image a serene and contemplative feel. The photo has a soft, realistic rendering style, emphasizing the intricate details of her eye.}}
\label{fig:qualsixty11}
\end{figure*}

\raggedbottom
\makeatletter\setlength{\@fptop}{0pt}\makeatother   
\renewcommand{\dblfloatpagefraction}{0.5}   
\rev{Figure~\ref{fig:qual1} and Figures~\ref{fig:qual2}--\ref{fig:qual8}} show frames of the same prompt generated by four frame-wise configurations of Table~\ref{tab:complete-240} (CF++ with detached writes, SGF, \rev{dKV Forcing without rollout dropout, and dKV Forcing}) at $0$, $48$, $96$, $144$, $192$ and $238$\,s. The eight 240\,s generations are four from the 128 MovieGen prompts (the 240\,s prompt set of SGF~\citep{zhuang2026sgf}) and four from the VBench-Long standard prompts (Figures~\ref{fig:qual2}, \ref{fig:qual3}, \ref{fig:qual5} and~\ref{fig:qual6}; $963$ latent frames; EMA weights, sampling seed $0$, \rev{first 4 + window 16 + current 1}). Figures~\ref{fig:qualsixty1}--\ref{fig:qualsixty11} show the same four configurations at $0$, $12$, $24$, $36$, $48$ and $59$\,s on eleven 60\,s generations ($241$ latent frames, same sampling) from the same prompt set. Figure~\ref{fig:teaser} shows dKV Forcing alone, and the same four generations are compared with the baselines in Figure~\ref{fig:qual60} (bubbles), Figure~\ref{fig:qualsixty2} (sword fight), Figure~\ref{fig:qual5} (cat) and Figure~\ref{fig:qual8} (kitchen).

In these examples the detached-write model loses the scene within the first minute and degenerates into color fields or a different scene, SGF keeps a plausible scene but drifts to other subjects or layouts, and dKV Forcing keeps the subject and the layout through the full minute or the full four minutes while the subject keeps moving\rev{, in contrast to dKV Forcing without rollout dropout, whose videos stay nearly static}.


\end{document}